\documentclass{article}

\usepackage{arxiv}

\usepackage[utf8]{inputenc} 
\usepackage[T1]{fontenc}    
\usepackage{hyperref}       
\usepackage{url}            
\usepackage{booktabs}       
\usepackage{amsfonts}       
\usepackage{nicefrac}       
\usepackage{microtype}      
\usepackage{xcolor}         
\usepackage{graphicx}       
\usepackage{adjustbox}      
\usepackage{subcaption}     
\usepackage{float}
\usepackage{tcolorbox}
\usepackage[linesnumbered,ruled,vlined]{algorithm2e}
\usepackage{amsthm}

\usepackage{amsmath}
\usepackage{amssymb}
\usepackage{mathtools}
\usepackage{amsthm}

\usepackage{pifont}
\usepackage[table,xcdraw]{xcolor}
\usepackage{multirow}
\usepackage{wrapfig}

\title{Rethinking Graph Structure Learning in the Era of LLMs}

\author{ 
        Zhihan Zhang \\
	Beijing Institude of Technology\\
	\texttt{3220241443@bit.edu.cn} \\
    \And
	Xunkai Li \\
	Beijing Institude of Technology\\
	\texttt{cs.xunkai.li@gmail.com} \\
    \And
    Yilong Zuo \\
	Beijing Institude of Technology\\
	\texttt{1120231863@bit.edu.cn} \\
    \And
    Henan Sun \\
	Beijing Institude of Technology\\
	\texttt{magneto0617@foxmail.com} \\
    \And 
    Zhenjun Li \\
    \textbf{Bing Zhou} \\
	Shenzhen Institute of Technology \\
    \texttt{lizhenjun@szsit.edu.cn} \\
    \texttt{zhoubing@szcp.edu.cn} \\
    \And
	Rong-Hua Li \\
    \textbf{Guoren Wang} \\
	Beijing Institude of Technology \\
    \texttt{lironghuabit@126.com} \\
    \texttt{wanggrbit@126.com} \\
}

\renewcommand{\shorttitle}{\textit{arXiv} Template}

\hypersetup{
pdftitle={A template for the arxiv style},
pdfsubject={q-bio.NC, q-bio.QM},
pdfauthor={David S.~Hippocampus, Elias D.~Striatum},
pdfkeywords={First keyword, Second keyword, More},
}

\begin{document}

\maketitle

\vspace{-5pt}
\begin{abstract}
\vspace{-5pt}
    Text-attributed graphs (TAGs) have become a key form of graph-structured data in modern data management and analytics, combining structural relationships with rich textual semantics for diverse applications. However, the effectiveness of analytical models, particularly graph neural networks (GNNs), is highly sensitive to data quality. Our empirical analysis shows that both conventional and LLM-enhanced GNNs degrade notably under textual, structural, and label imperfections, underscoring TAG quality as a key bottleneck for reliable analytics.
    Existing studies have explored data-level optimization for TAGs, but most focus on specific degradation types and target a single aspect like structure or label, lacking a systematic and comprehensive perspective on data quality improvement.
    To address this gap, we propose \textbf{LAGA} (Large Language and Graph Agent), a unified multi-agent framework for comprehensive TAG quality optimization. LAGA formulates graph quality control as a data-centric process, integrating detection, planning, action, and evaluation agents into an automated loop. It holistically enhances textual, structural, and label aspects through coordinated multi-modal optimization.
    Extensive experiments on 5 datasets and 16 baselines across 9 scenarios demonstrate the effectiveness, robustness and scalability of LAGA, confirming the importance of data-centric quality optimization for reliable TAG analytics. 
\end{abstract}

\section{Introduction}
\label{section:introduction}
\vspace{-5pt}

    \begin{wrapfigure}{l}{0.5\textwidth}
        \centering
        \vspace{-10pt}
        \includegraphics[width=\linewidth]{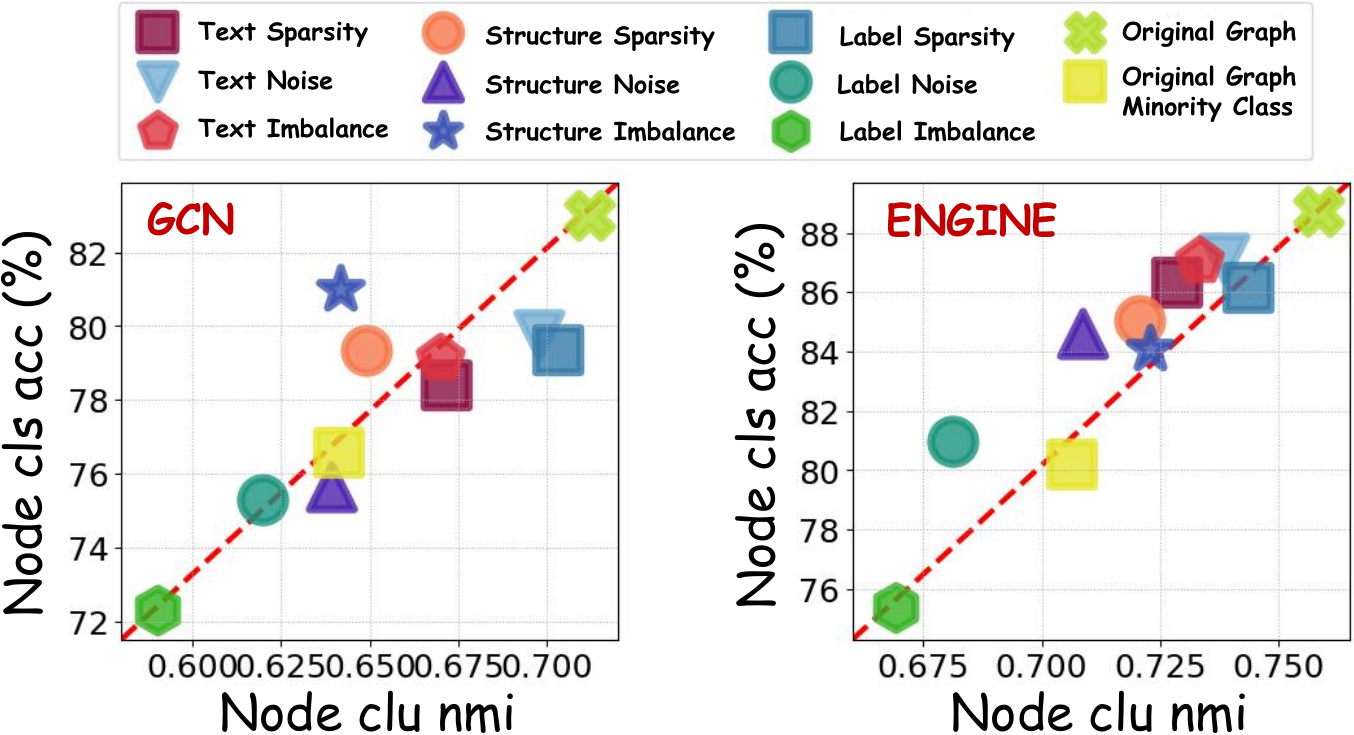} 
        \vspace{-13pt}
        \caption{Performance comparison of GCN and ENGINE across 9 TAG quality scenarios on Cora.}
        \vspace{-5pt}
        \label{figure:emperical_study2}
    \end{wrapfigure}

    Text-attributed graphs (TAGs) provide a natural way to couple graph structure with node-associated text, enabling learning systems to jointly exploit relational signals and rich semantic content for tasks such as classification, retrieval, and reasoning~\cite{rethinkingtag,tsgfm,li2024graph}. Recently, LLM-augmented graph neural networks (GNNs) have emerged as a strong paradigm for TAGs, using LLMs to enrich node semantics while leveraging message passing to capture structural dependencies. However, the performance of LLM-augmented GNNs on TAGs is highly sensitive to data quality. Real-world TAGs commonly suffer from imperfections across three modalities—text, structure, and labels—each of which may exhibit sparsity, noise, or imbalance. We summarize these issues in a 3-by-3 taxonomy with nine representative degradation scenarios, and our empirical results (Fig.~\ref{figure:emperical_study2}) show that such defects cause significant performance degradation at both node and graph levels for traditional as well as LLM-augmented GNNs. (More details are in Appendix~\ref{appendix:emperical_study}). These observations motivate the need for systematic quality improvement of TAGs to ensure robust and generalizable learning on graph-structured data.
    
    Existing methods have started addressing TAG quality issues, such as noise reduction, structure refinement, and label balancing~\cite{tsgfm,noisygl,gslsurvey,iglbench}. However, they have several limitations:
    \ding{172} \textbf{Limited coverage.} TAGs experience a wide range of defects across text, structure, and labels, summarized into nine scenarios. Yet most existing methods~\cite{llm4rgnn, llata, crgnn, rawlsgcn, ultratag-s} focus on one or two isolated issues, lacking a unified, comprehensive optimization approach.
    \ding{173} \textbf{Text modality overlooked.} Many methods~\cite{cogsl, topoauc, rncgln} assume purely structural graphs and fail to address text-specific issues like sparsity, noise, and imbalance, limiting their effectiveness on TAGs.
    \ding{174} \textbf{Poor generalizability.} Several methods rely on task-specific modules or designs that are tightly coupled with particular backbones, reducing scalability and reuse, especially in dynamic or system-level applications.
    These limitations hinder the deployment of TAGs in real-world analytics, where high-quality, semantically rich graphs are essential for reliable querying, reasoning, and downstream modeling. A unified, data-centric framework is needed to optimize TAG quality across all scenarios.
    
   To address these challenges, we propose \textbf{LAGA} (Large Language and Graph Agent), a unified, automated LLM-based multi-agent framework for enhancing TAG quality. Unlike prior approaches focused on stronger GNN architectures, LAGA emphasizes \textbf{data-centric optimization}, treating graph quality control as a primary task. The framework features a multi-agent system with four roles: \textit{Detection}, \textit{Planning}, \textit{Action}, and \textit{Evaluation}. These agents work in a closed loop to detect multi-modal quality issues, generate adaptive repair plans with LLM reasoning, executes tailored optimization strategies, and iteratively assess results. The \textbf{Action agent}, central to the framework, uses a dual-encoder design (semantic + structure encoders) to optimize text, structure, and label objectives, enabling comprehensive TAG quality enhancement. This approach not only improves LAGA’s reliability but also provides a holistic method for building high-quality TAGs, essential for graph analytics and downstream reasoning in data-driven applications.
    
    The design of LAGA is guided by three key insights that shape its architecture and optimization strategy:
    \ding{182} \textbf{Why a multi-agent architecture?} TAG quality defects are diverse (across text, structure, and labels) and require \emph{different} competencies at different stages: (i) accurately localizing and characterizing anomalies, (ii) deciding \emph{what} to fix and \emph{in what order} under budget/side-effect constraints, (iii) executing concrete repair operations, and (iv) validating whether changes truly improve quality without introducing new issues. It is difficult for a single monolithic model to reliably cover this end-to-end pipeline. LAGA therefore decomposes the workflow into four specialized agents—\textit{Detection}, \textit{Planning}, \textit{Action}, and \textit{Evaluation}—which enables targeted expertise, modular updates, and a fully automated closed-loop optimization process.
    \ding{183} \textbf{Why LLM-driven planning?} TAG defects are often ambiguous, context-dependent, and intertwined across modalities, making fixed rules or heuristic templates brittle. LLM-based reasoning allows LAGA to interpret detection signals, weigh trade-offs, and generate adaptive, cross-modality repair plans that coordinate multiple operators.
    \ding{184} \textbf{Why cross-modality joint learning?} Defects in one modality can mask or amplify defects in another, so optimizing text/structure/labels in isolation can lead to inconsistent updates. LAGA uses a dual-encoder design with modality-specific objectives to integrate complementary signals, prioritize more reliable information, and improve TAG quality in a holistic and robust manner.

    \textbf{Our Contributions.} 
    (1) \textit{\underline{New perspective.}} We establish a unified taxonomy of TAG quality issues across three modalities and three defect types, providing a holistic view of graph quality challenges and framing them as a \textit{data-centric problem} relevant to data management.  
    (2) \textit{\underline{New method.}} We propose \textbf{LAGA}, a LLM-based multi-agent graph quality optimization approach with four collaborative agents. By comprehensively enhancing TAG quality, LAGA directly improves reliability across various GNN backbones and downstream tasks. 
    (3) \textit{\underline{SOTA performance.}} \textbf{LAGA} achieves state-of-the-art results across all 9 degradation scenarios, multiple tasks, and 5 datasets, demonstrating not only model effectiveness but also the importance of high-quality TAGs for reliable graph data management and analytics across diverse data-centric scenarios.

\section{Preliminaries}

    \subsection{Notations Formulation}
    \textbf{Node-wise Text-Attributed Graph}. We consider a TAG defined as $\mathcal{G}=(\mathcal{V}, \mathcal{E}, \mathcal{T})$ with $n$ nodes and $m$ edges, where its structure is represented by a symmetric adjacency matrix $\mathbf{A}$. Each node $v_i \in \mathcal{V}$ is associated with a text description $\mathrm{t}_i \in \mathcal{T}$. A language model encodes each $\mathrm{t}_i$ into a feature vector $x_i$, forming the node feature matrix $\mathbf{X} \in \mathbb{R}^{n \times f}$. Furthermore, $\mathbf{Y} \in \mathbb{R}^{n \times d}$ denotes the one-hot label matrix, where $c \in \mathcal{C}$ represents a specific class.

    \subsection{Quality Optimization Objective}  
    We categorize the quality issues on TAGs into nine types, covering three modalities (text, structure and label) and three types of problems (sparsity, noise and imbalance). The detailed problem definition is in the Appendix~\ref{appendix:scene_defination}. Given the above challenges, our goal is to improve the overall quality of TAGs by jointly addressing issues in text, structure, and label modalities. Formally, for an input TAG  $\mathcal{G} = (\mathcal{V}, \mathcal{E}, \mathcal{T}, \mathbf{Y})$, we aim to construct an optimal graph  $\mathcal{G}^{opt} = (\mathcal{V}^{opt}, \mathcal{E}^{opt}, \mathcal{T}^{opt}, \mathbf{Y}^{opt})$. 
    The objective is that node representations learned on $\mathcal{G}^{opt}$ yield consistently better performance across diverse GNN/LLM-GNN backbones $f_{\theta}$ and downstream tasks $\mathcal{Z}$: 
    \begin{equation}
        \forall f_{\theta}, \; \forall z \in \mathcal{Z}, \quad \mathcal{M}(f_{\theta}(\mathcal{G}^{opt})) \; \geq \; \mathcal{M}(f_{\theta}(\mathcal{G})),
    \vspace{-1pt}
    \end{equation}
    where $\mathcal{M}(\cdot)$ denotes the evaluation metric (e.g., Accuracy, Normalized MI).  
    In other words, the optimization aims to holistically refine $(\mathcal{V}, \mathcal{E}, \mathcal{T}, \mathbf{Y})$ $\rightarrow (\mathcal{V}', \mathcal{E}', \mathcal{T}', \mathbf{Y}')$, ensuring robust and reliable graph analytics.

    \subsection{Related Work}
    GNNs are widely used for analyzing graph-structured data, but real-world graphs often suffer from degradations in text, structure, and labels, leading to unreliable analyses. Recent work has focused on enhancing GNN robustness by tackling noise, sparsity, and imbalance across these modalities.
    
    \textbf{Text-related issues.}
    In TAGs, degraded text quality—such as sparsity, noise, or incompleteness—weakens semantic representations. Existing methods improve robustness via semantic augmentation, denoising, and retrieval-based enrichment, as seen in UltraTAG-S~\cite{ultratag-s}, CTD-MLM~\cite{ctd-mlm}, and PoDA~\cite{poda}. 
    \textbf{Structure-related issues.}
    Structural degradation, like noisy edges or topological imbalance, hinders representation learning. Methods such as similarity modeling or probabilistic optimization refine connectivity~\cite{idgl,prognn,sublime}, while LLM-based approaches like LLM4RGNN~\cite{llm4rgnn} and LLaTA~\cite{llata} enable semantic-aware refinement. Others, like Tail-GNN~\cite{tail-gnn} and GraphPatcher~\cite{graphpatcher}, address imbalance through reweighting or edge generation.
    \textbf{Label-related issues.}
    Label sparsity, noise, and imbalance challenge robust graph analysis. Solutions like semi-supervised propagation, pseudo-label correction, and class-balanced optimization in GraphHop~\cite{graphhop}, PI-GNN~\cite{pi-gnn}, and GraphSHA~\cite{graphsha} enhance learning stability under unreliable supervision.

\section{Methods}

\subsection{Overview of LAGA}
\begin{figure*}[htbp]
        \centering
        \includegraphics[width=\textwidth]{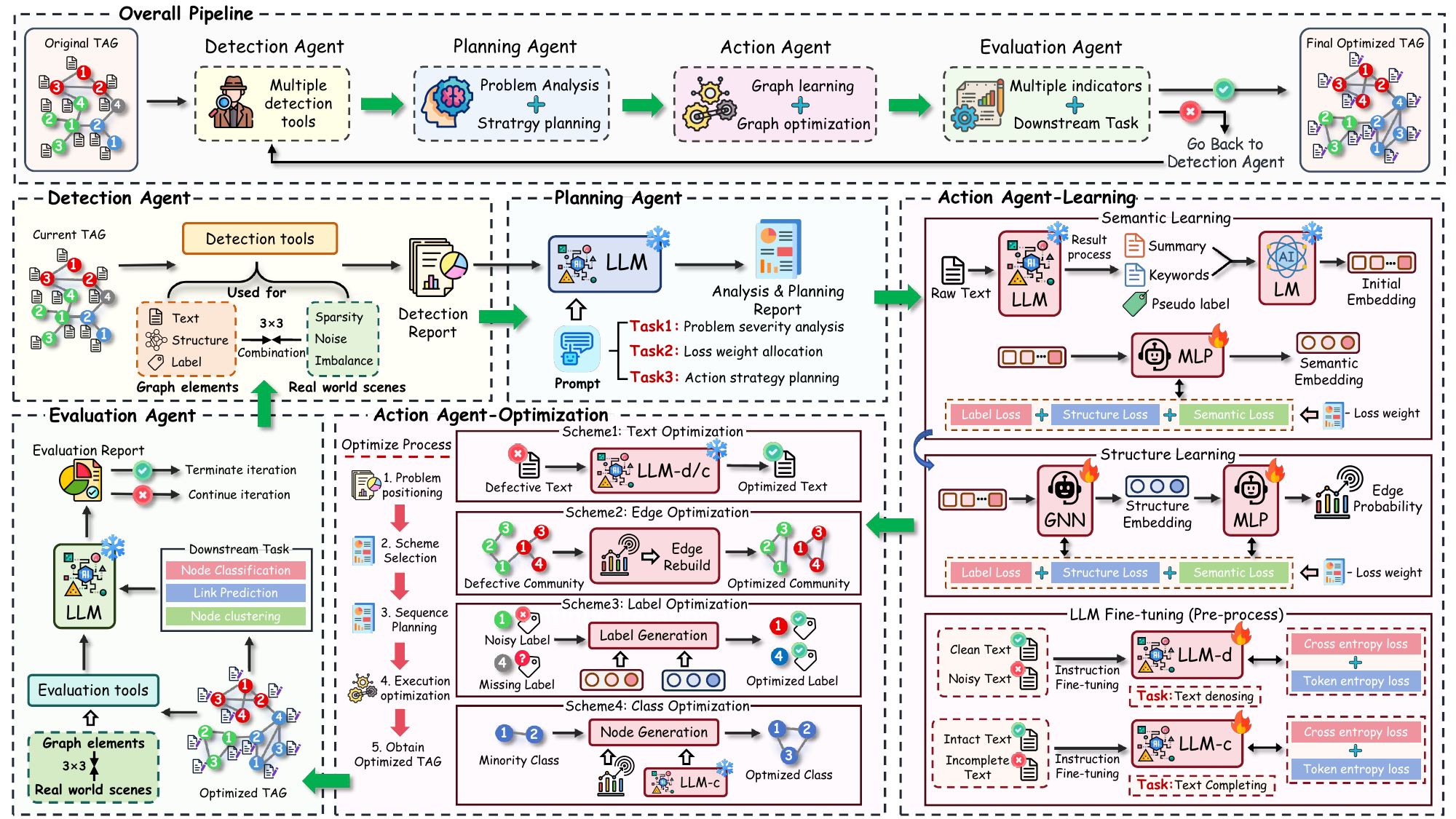}
        \caption{
            The framework of LAGA, including the overall workflow of LAGA and the internal module details of each agent.
        }
        \label{figure:framework}
    \end{figure*}

    We propose \textbf{LAGA}, a LLM-based multi-agent framework for iterative quality optimization of TAGs. As shown in Fig.~\ref{figure:framework}, LAGA includes four agents in a closed-loop pipeline: \textit{Detection}, \textit{Planning}, \textit{Action}, and \textit{Evaluation}. This end-to-end, data-driven framework integrates graph learning with issue diagnosis, adaptive optimization, and quality assessment, enabling continuous improvement of TAG quality across diverse real-world scenarios.

\subsection{Detection Agent}
    The Detection Agent systematically identifies and localizes data quality issues in TAGs by evaluating text, structure, and label modalities. Specifically targeting sparsity, noise, and imbalance, it generates a comprehensive diagnostic report to effectively guide subsequent analysis and automated planning.
    
    \textbf{Text-level Detection.}  
    For each node text \( t_i \), sparsity is measured by text length: a node is flagged sparse if \( |t_i| < \tau_s^{\text{text}} \). Noise is quantified by the error rate: $N^{\text{text}}_i = \frac{\#\text{text-errors}(t_i)}{|t_i|}$. A node is noisy if \( N^{\text{text}}_i > \tau_n^{\text{text}} \). And imbalance is captured using average TF-IDF~\cite{tfidf}:
    \begin{equation}
        I^{\text{text}}_i = \frac{1}{|t_i|} \sum_{w \in t_i} \frac{\#(w,t_i)}{|t_i|} \log \frac{|T|}{1 + |\{t_j \in T : w \in t_j\}|}, \; \text{text is uninformative if} \; I^{\text{text}}_i < \tau_{imb}^{\text{text}}.
    \end{equation}
    
    \textbf{Structure-level Detection.}  
    The agent partitions the graph into communities \( \{C_k\} \) using the Louvain method~\cite{louvain}. Sparsity is assessed by average node degree and edge density:
    \begin{equation}
        \bar{\mathbf{d}}(C_k) = \frac{1}{|C_k|}\sum_{v\in C_k} \mathbf{d}(v), \quad \rho(C_k) = \frac{2|\mathcal{E}(C_k)|}{|C_k|(|C_k|-1)}.
    \end{equation}
    Noise is captured by structural entropy~\cite{se} and Jaccard similarity:
    \begin{equation}
        SE(C_k) = -\sum_{v\in C_k} \frac{\mathbf{d}(v)}{\sum_{u\in C_k} \mathbf{d}(u)} \log \frac{\mathbf{d}(v)}{\sum_{u\in C_k} \mathbf{d}(u)}, \;
        \bar{J}(C_k) = \frac{1}{|\mathcal{E}(C_k)|}\sum_{(u,v)\in \mathcal{E}(C_k)} \frac{|\mathcal{N}(u)\cap \mathcal{N}(v)|}{|\mathcal{N}(u)\cup \mathcal{N}(v)|}.
    \end{equation}
    Structural imbalance is detected by analyzing the coefficient of variation of the degree distribution \( P(\mathbf{d}) \).
    
    \textbf{Label-level Detection.}  
    Sparsity is reflected by unlabeled nodes: \( S^{\text{label}}_i = \mathbb{I}[y_i = \varnothing] \). Noise is detected using neighborhood majority voting and adaptive \( k \)-means clustering, with the final prediction selected based on confidence scores:
    \begin{equation}
        \hat{y}_i = \arg\max_{j\in\{1,2\}} cs_i^{(j)} \quad \text{s.t.} \quad \max_{j\in\{1,2\}} cs_i^{(j)} \geq \tau_c,
    \end{equation}
    where $cs$ represents confidence scores. Imbalance is identified by examining the empirical label distribution \( p_c \).
    
    Finally, the Detection Agent generates a report: $\;\mathcal{R}_{\text{det}} = (\mathcal{M}_{\text{global}}, \mathcal{M}_{\text{local}}),\;$
    where \( \mathcal{M}_{\text{global}} \) provides an overview of the graph’s quality, and \( \mathcal{M}_{\text{local}} \) localizes specific issues. This report guides the Planning Agent in strategy design and informs the Action Agent for targeted optimization.

\subsection{Planning Agent}
    The Planning Agent transcends fixed rule-based heuristics by dynamically formulating optimization trajectories based on detection reports. It utilizes LLMs to infer issue severity and generate a tailored plan (e.g., priority ranking, loss re-weighting), acting as an intelligent optimizer to maximize quality gain under varying degradation patterns, see Appendix~\ref{appendix:dynamic_plan} for a case study vs. static strategies.
    
    \textbf{Severity \& Priority Analysis.}  
    The Planning Agent uses global statistics (\(\mathcal{M}_{global}\)) to evaluate each of the nine quality issues, assigning severity levels from \(\mathcal{S}_{ser}=\{\texttt{negligible:}0, \texttt{mild:}1, \texttt{moderate:}2, \texttt{severe:}3\}\). The severity levels are used with heuristic rule-weights \(\mathbf{r}\) to generate an optimization priority order, producing a report \(\mathcal{R}_{ana}=(\mathbf{s}, \boldsymbol{\pi})\) for further optimization.
    
    \textbf{Reliability-Aware Loss Weighting.}  
    For stable training, the Planning Agent configures loss weights based on aggregated severities \(\bar{s}_{\text{text}}\), \(\bar{s}_{\text{struct}}\), and \(\bar{s}_{\text{label}}\). The LLM uses these to output weights \((\alpha, \beta, \gamma)\) that sum to 1, ensuring severity-aware loss weighting for graph learning:
    \begin{equation}
    (\alpha, \beta, \gamma) = \Omega_{\text{LLM}}\left(\tilde{\boldsymbol{\omega}}, \, \bar{\mathbf{s}} \,|\, \mathcal{P}_{\mathbf{r}}\right), \quad \alpha + \beta + \gamma = 1.
    \end{equation}
    
    \textbf{Optimization Strategy Planning.}  
    The Planning Agent uses a cost-aware strategy, selecting a sequence of actions from an action library \(\mathcal{A}\). The LLM optimizes for quality gain while considering cost, using the formula:
    \begin{equation}
    \mathcal{P}^\star = \arg\max_{\mathcal{P} \subseteq \mathcal{A}} \left[ U(\mathcal{P}) - \lambda \, CT(\mathcal{P}) \right] \quad \text{s.t.} \ \text{PREC}(\mathcal{P}, \boldsymbol{\pi}),
    \end{equation}
    where $U(\mathcal{P})$ is the expected quality gain, $CT(\mathcal{P})$ is the cost, $\lambda \geq 0$ controls the balance and $\text{PREC}(\mathcal{P},\boldsymbol{\pi})$ enforces the priority order.
    
    Finally, the Planning Agent outputs a planning report:
    $\mathcal{R}_{plan} = (\mathcal{R}_{ana}, (\alpha, \beta, \gamma), \mathcal{P}^\star),\;$
    which guides targeted optimization by the Action Agent and tracks progress through the Evaluation Agent.

\subsection{Action Agent}
\label{section:action_agent}

    The Action Agent is the central executor in LAGA, transforming analysis and planning into operations that improve TAG quality. It unifies \emph{Graph Learning}, which builds robust embeddings, with \emph{Graph Optimization}, which repairs and enhances data elements. This integration ensures seamless model training and data refinement, improving downstream analytics.
    
    \noindent \ding{117} \textbf{Graph Learning.}  
    Graph Learning builds robust semantic and structural representations, integrating multi-modal signals (text, structure, and labels). The algorithm is provided in Appendix~\ref{appendix:alg_gl}.
    
    \textbf{Semantic Learning.}  
    Each node's text representation \( \mathrm{t}_i \) is enriched by LLM-generated summaries, keywords, and pseudo-labels. The initial embedding \( h_i^{init} \) is processed by an LM encoder to produce a semantic embedding \( h_i^{sem} \), optimized by a multi-objective loss:
    \begin{equation}
        \mathcal{L}_{total}^{sem} = \alpha\,\mathcal{L}_{sema} 
        + \beta\,\mathcal{L}_{stu} 
        + \gamma\,\mathcal{L}_{label}, \;
        \mathcal{L}_{sema} = \|h^{sem}-y^{pse}\|^2, 
    \end{equation}
    \begin{equation}
        \mathcal{L}_{stu}(z_i) 
        = - \log \frac{\sum_{j \in \mathcal{N}(i)} e^{\mathrm{sim}(z_i,z_j)/\tau}}
        {\sum\limits_{j \in \mathcal{N}(i)} \!\!\!e^{\mathrm{sim}(z_i,z_j)/\tau} + \sum\limits_{n \in \mathcal{N}^-(i)} \!\!\!e^{\mathrm{sim}(z_i,z_n)/\tau}}, \;
        \mathcal{L}_{label} = \mathrm{CE}(h^{sem},y),
    \end{equation}
    where \( \mathcal{N}(i) \) and \( \mathcal{N}^-(i) \) represent neighboring and non-neighboring nodes, respectively. $\mathrm{CE}(\cdot)$ is the cross entropy and $\mathrm{sim}(z_i,z_j)$ represents the cosine similarity.
    
    \textbf{Structure Learning.}  
    A GCN learns structural embeddings \( h_i^{stu} \), and a link predictor \( \hat{a}_{ij} \) estimates edge probabilities:
    $\hat{a}_{ij} = \sigma(\mathrm{MLP}([h^{stu}_i \|\ h^{stu}_j])).\;$
    The total structure learning loss is:
    \begin{equation}
        \mathcal{L}_{total}^{stu} = \alpha\,\mathcal{L}_{sema} 
        + \beta\,\mathcal{L}_{stu}
        + \gamma\,\mathcal{L}_{label}, \;
        \mathcal{L}_{stu} = \|\mathbf{A} - \hat{\mathbf{A}}\|^2, \;
        \mathcal{L}_{label} = \mathrm{CE}(h^{stu},y),
    \end{equation}
    \begin{equation}
        \mathcal{L}_{sema} = \|h^{stu}-h^{sem}\|^2 + 
        \sum_{(i,j)} \big( \sigma(\mathrm{sim}(h^{sem}_i,h^{sem}_j)) - \hat{a}_{ij} \big)^2.
    \end{equation}
    Aligning $h^{stu}$ and $h^{sem}$ captures both topological patterns and semantic information, ensuring robust structure learning through a multi-loss design that integrates semantic alignment, structural reconstruction, and label supervision.
    
    \textbf{LLM Fine-tuning.}  
    To improve textual quality, the LLM is fine-tuned for denoising and completion using a loss that combines cross-entropy and token-level entropy:
    \begin{equation}
        \mathcal{L}_{LLM} = \mathrm{CE}(t^{out}, t^{ref}) + \lambda H(t^{out}), \;
        H(t^{out}) = - \frac{1}{|t^{out}|} \sum_{k=1}^{|t^{out}|} \sum_{w \in \mathcal{W}} p_\theta(w|t^{out}_{<k}) \log p_\theta(w|t^{out}_{<k}),
    \end{equation}
    where \( H(t^{out}) \) is the token-level entropy, encouraging more informative and diverse token generation.
    
    \noindent \ding{117} \textbf{Graph Optimization.}  
    Graph Optimization applies quality-improving operations based on the detection and planning reports. It selects actions from the action library \( \mathcal{A} \) and applies them to repair and enhance the graph. The algorithm is provided in Appendix~\ref{appendix:alg_go} and the optimization process is modeled as:
    \begin{equation}
        \mathcal{OPT}: (\mathcal{G} \,|\, \mathcal{M}_{local}, \mathcal{P}^\star, \mathcal{A}) \longmapsto \mathcal{G}' = \{\mathcal{V}', \mathcal{E}', \mathcal{T}', \mathbf{Y}'\}.
    \end{equation}
    
    \textbf{Text Optimization (\(\mathcal{A}^{\text{t}}\))}.  
    For text issues, the fine-tuned LLM is used for denoising or completion:
    \begin{equation}
        t_i^{opt} = \mathcal{F}_\text{LLM}(t_i, Con(i) \,|\, \mathcal{P}_{\text{denoise}} \text{ or } \mathcal{P}_{\text{complete}}),
    \end{equation}
    where $Con(i)$ denotes optional context (e.g., neighbor texts), and $\mathcal{P}_{task}$ is the prompt template.
    
    \textbf{Structure Optimization (\(\mathcal{A}^{\text{s}}\))}.  
    For structural sparsity or noise, edges are pruned or added based on predicted probabilities \( \hat{a}_{ij} \):
    \begin{equation}
        \hat{A}^{opt}_{ij} = 
        \begin{cases}
        0, & (i,j)\in E(C_k) \ \text{and}\ \hat{a}_{ij}<\tau_{edge},\\
        1, & (i,j)\notin E(C_k),\ j=\arg\max_{u}\hat{a}_{iu}, \;\text{s.t.}\ \deg(i)<k_{edge}\ \text{and}\ \hat{a}_{iu}>\tau_{edge}, \\
        a_{ij}, & otherwise. 
        \end{cases}
    \end{equation}

    \textbf{Label Optimization (\(\mathcal{A}^{\text{l}}\))}. 
    For noisy or missing labels, the Action Agent uses classification Prediction via  $h^{stu}_i$ and neighborhood voting. First, we define a prediction confidence score:
    \begin{equation}
        c_i = \exp\Big( \lambda \log p_{m}(h_i) + (1-\lambda)\log\big(p_{m}(h_i)-p_{sm}(h_i)\big) \Big),
    \end{equation}
    where $p_{m}(h_i)$ is the highest predicted class probability and $p_{sm}(h_i)$ is the second highest probability. If \( c_i > \tau_{\text{lape}} \), we assign \( \hat{y}_i = \arg\max_y p(h_i^{stu}) \); otherwise, node label is voted by its neighbors:
    \begin{equation}
        \hat{y}_i = \arg\max_{c \in \mathcal{Y}} \sum_{j \in \mathcal{N}(i)} \hat{a}_{ij} \mathbb{I}[y_j = c].
    \end{equation}
    \textbf{Node Generation (\(\mathcal{A}^{\text{c}}\))}.  
    To alleviate label imbalance, synthetic nodes are generated for minority classes:
    \begin{equation}
        \mathrm{t}_{new} = \mathcal{F}_{\text{LLM}}(\varnothing, Con_c \,|\, \mathcal{P}_{\text{complete}}), \; h^{init}_{new} = LM(\mathrm{t}_{new}).
    \end{equation}
    Edges are added based on predicted probabilities $\hat{a}_{ij}$. For a class $c$ with count $|V_c|$ below a threshold: $\tau_{gen} = r_{gen}\frac{1}{d}\sum_{c \in \mathcal{C}}(|V_c|)$, we generate $n_c=\tau_{gen}-|V_c|$ nodes.

\subsection{Evaluation Agent}
    The Evaluation Agent acts as the \textit{quality controller} of LAGA, evaluating the optimized TAG and determining whether further optimization is needed. It ensures that the closed-loop optimization converges to a reliable graph.
    
    \textbf{Evaluation Process.}  
    The Evaluation Agent combines three sources of evidence: (i) results from evaluation tools (same as detection) $\mathcal{M}'_{\text{global}}$, (ii) downstream task performance $\mathcal{M}_{\text{down}}$, and (iii) the previous evaluation report $\mathcal{R}^{\text{prev}}_{\text{eval}}$. These inputs are provided to the LLM, which outputs a quality score $q \in [0,10]$ and a binary decision $\delta \in \{\texttt{True}, \texttt{False}\}$, indicating whether more optimization is required:
    \vspace{-3pt}
    \begin{equation}
        (q, \delta) = \Psi_{\text{LLM}} \left( \mathcal{M}'_{\text{global}}, \mathcal{M}_{\text{down}}, \mathcal{R}^{\text{prev}}_{\text{eval}} \right).
    \end{equation}
    The final evaluation report is:
    $\mathcal{R}_{\text{eval}} = (\mathcal{M}'_{\text{global}}, \mathcal{M}_{\text{down}}, q, \delta).$
    
    \textbf{Stopping Criterion.}  
    For the first iteration, optimization stops if \( q > \tau_{\text{impf}} \) and \( \delta = \texttt{False} \); otherwise, it continues. For subsequent iterations, the process stops if:
    $(q^t - q^{t-1} < \tau_{\text{imp}}) \ \text{and} \ (\delta = \texttt{False}),\;$
    otherwise another optimization round is triggered.
    By combining evaluation tool signals, downstream performance, and LLM-as-a-Judge, the Evaluation Agent establishes a closed feedback loop, ensuring adaptive quality assessment and governing the optimization trajectory.

    \begin{table*}[t]
    \centering
    \caption{Node classification accuracy (\%) comparison across datasets with varying perturbation ratios and scenarios, each scenario contains two baselines. The highest results are highlighted in \colorbox[HTML]{DADADA}{\textbf{bold}}, while ``$\star$'' indicates that the baseline includes LLM-related technologies.}
    \resizebox{\textwidth}{!}{
    \begin{tabular}{c|c|c c c c c c c c c c c c}
    \specialrule{1.5pt}{1.5pt}{1.5pt}
    \multicolumn{2}{c|}{\textbf{Dataset}} 
    & \multicolumn{3}{c}{\textbf{Cora}} 
    & \multicolumn{3}{c}{\textbf{Citeseer}} 
    & \multicolumn{3}{c}{\textbf{WikiCS}} 
    & \multicolumn{3}{c}{\textbf{Photo}} \\ 
    \cmidrule(lr){1-2} \cmidrule(lr){3-5} \cmidrule(lr){6-8} \cmidrule(lr){9-11} \cmidrule(lr){12-14}
    \multicolumn{2}{c|}{\textbf{Perturbation Ratio}} & \textbf{0.2} & \textbf{0.4} & \textbf{0.8} & \textbf{0.2} & \textbf{0.4} & \textbf{0.8}
    & \textbf{0.2} & \textbf{0.4} & \textbf{0.8} & \textbf{0.2} & \textbf{0.4} & \textbf{0.8} \\ 
    \midrule
    
    \multirow{3}{*}{\parbox{1.5cm}{\centering \textbf{Text\\Sparsity}}} 
    & \cellcolor[HTML]{FFF0D5} $\text{LLM-TG}^{\star}$  
    & $83.86_{\scriptstyle \pm 0.27}$ & $79.33_{\scriptstyle \pm 0.24}$ & $73.20_{\scriptstyle \pm 0.21}$ & $73.43_{\scriptstyle \pm 0.15}$ & $69.70_{\scriptstyle \pm 0.21}$ & $61.50_{\scriptstyle \pm 0.19}$ & $83.02_{\scriptstyle \pm 0.08}$ & $80.97_{\scriptstyle \pm 0.15}$ & $77.51_{\scriptstyle \pm 0.11}$ & $82.65_{\scriptstyle \pm 0.31}$ & $80.90_{\scriptstyle \pm 0.37}$ & $76.88_{\scriptstyle \pm 0.51}$ \\ 
    & \cellcolor[HTML]{FFF0D5} $\text{UltraTAG-S}^\star$ & 
    $\underline{87.08_{\scriptstyle \pm 0.72}}$ & $\underline{83.21_{\scriptstyle \pm 0.77}}$ & $\underline{76.45_{\scriptstyle \pm 1.16}}$ & $\underline{77.13_{\scriptstyle \pm 0.23}}$ & $\underline{73.34_{\scriptstyle \pm 0.24}}$ & $\underline{64.72_{\scriptstyle \pm 0.19}}$ & $\underline{83.72_{\scriptstyle \pm 0.14}}$ & $\underline{81.95_{\scriptstyle \pm 0.35}}$ & $\underline{78.57_{\scriptstyle \pm 0.41}}$ & $\underline{84.69_{\scriptstyle \pm 0.10}}$ & $\underline{83.72_{\scriptstyle \pm 0.09}}$ & $\underline{79.61_{\scriptstyle \pm 0.11}}$ \\  
    & \cellcolor[HTML]{FFF0D5} LAGA (Ours) & 
    \cellcolor[HTML]{DADADA}$\textbf{88.34}_{\scriptstyle \pm \textbf{0.63}}$ & 
    \cellcolor[HTML]{DADADA}$\textbf{86.92}_{\scriptstyle \pm \textbf{0.81}}$ & 
    \cellcolor[HTML]{DADADA}$\textbf{84.15}_{\scriptstyle \pm \textbf{0.88}}$ & 
    \cellcolor[HTML]{DADADA}$\textbf{81.46}_{\scriptstyle \pm \textbf{0.20}}$ & 
    \cellcolor[HTML]{DADADA}$\textbf{79.73}_{\scriptstyle \pm \textbf{0.24}}$ & 
    \cellcolor[HTML]{DADADA}$\textbf{76.99}_{\scriptstyle \pm \textbf{0.24}}$ & 
    \cellcolor[HTML]{DADADA}$\textbf{85.18}_{\scriptstyle \pm \textbf{0.26}}$ & 
    \cellcolor[HTML]{DADADA}$\textbf{83.19}_{\scriptstyle \pm \textbf{0.31}}$ & 
    \cellcolor[HTML]{DADADA}$\textbf{80.02}_{\scriptstyle \pm \textbf{0.32}}$ & 
    \cellcolor[HTML]{DADADA}$\textbf{86.75}_{\scriptstyle \pm \textbf{0.14}}$ & 
    \cellcolor[HTML]{DADADA}$\textbf{85.38}_{\scriptstyle \pm \textbf{0.15}}$ &
    \cellcolor[HTML]{DADADA}$\textbf{82.70}_{\scriptstyle \pm \textbf{0.18}}$ \\  
    \midrule

    \multirow{3}{*}{\parbox{1.5cm}{\centering 
    \textbf{Text\\Noise}}} 
    & \cellcolor[HTML]{FFF7CD} $\text{PoDA}^{\star}$ 
    & $83.12_{\scriptstyle \pm 0.41}$ & 
    $82.87_{\scriptstyle \pm 0.42}$ & 
    $\underline{81.03_{\scriptstyle \pm 0.42}}$ & $74.16_{\scriptstyle \pm 0.21}$ & 
    $72.35_{\scriptstyle \pm 0.22}$ & 
    $68.27_{\scriptstyle \pm 0.26}$ & 
    $82.64_{\scriptstyle \pm 0.15}$ & 
    $81.97_{\scriptstyle \pm 0.16}$ & 
    $80.67_{\scriptstyle \pm 0.16}$ & 
    $84.61_{\scriptstyle \pm 0.63}$ & 
    $\underline{83.10_{\scriptstyle \pm 0.66}}$ & $80.13_{\scriptstyle \pm 0.71}$ \\
    & \cellcolor[HTML]{FFF7CD} $\text{CTD-MLM}^\star$ & 
    $\underline{84.39_{\scriptstyle \pm 0.24}}$ & $\underline{83.27_{\scriptstyle \pm 0.22}}$ & $80.15_{\scriptstyle \pm 0.28}$ & $\underline{74.88_{\scriptstyle \pm 0.13}}$ & $\underline{73.19_{\scriptstyle \pm 0.19}}$ & $\underline{71.03_{\scriptstyle \pm 0.19}}$ & $\underline{82.71_{\scriptstyle \pm 0.36}}$ & $\underline{82.03_{\scriptstyle \pm 0.38}}$ & $\underline{80.89_{\scriptstyle \pm 0.47}}$ & $\underline{84.75_{\scriptstyle \pm 0.13}}$ & $82.84_{\scriptstyle \pm 0.13}$ & $\underline{81.44_{\scriptstyle \pm 0.19}}$ \\  
    & \cellcolor[HTML]{FFF7CD} LAGA (Ours) & 
    \cellcolor[HTML]{DADADA}$\textbf{88.74}_{\scriptstyle \pm \textbf{0.41}}$ & 
    \cellcolor[HTML]{DADADA}$\textbf{88.27}_{\scriptstyle \pm \textbf{0.46}}$ & 
    \cellcolor[HTML]{DADADA}$\textbf{87.96}_{\scriptstyle \pm \textbf{0.46}}$ & 
    \cellcolor[HTML]{DADADA}$\textbf{83.12}_{\scriptstyle \pm \textbf{0.16}}$ & 
    \cellcolor[HTML]{DADADA}$\textbf{81.34}_{\scriptstyle \pm \textbf{0.17}}$ & 
    \cellcolor[HTML]{DADADA}$\textbf{79.19}_{\scriptstyle \pm \textbf{0.20}}$ & 
    \cellcolor[HTML]{DADADA}$\textbf{84.71}_{\scriptstyle \pm \textbf{0.22}}$ & 
    \cellcolor[HTML]{DADADA}$\textbf{83.08}_{\scriptstyle \pm \textbf{0.24}}$ & 
    \cellcolor[HTML]{DADADA}$\textbf{82.44}_{\scriptstyle \pm \textbf{0.25}}$ & 
    \cellcolor[HTML]{DADADA}$\textbf{87.03}_{\scriptstyle \pm \textbf{0.04}}$ & 
    \cellcolor[HTML]{DADADA}$\textbf{86.02}_{\scriptstyle \pm \textbf{0.06}}$ & 
    \cellcolor[HTML]{DADADA}$\textbf{84.87}_{\scriptstyle \pm \textbf{0.06}}$ \\
    \midrule
    
    \multirow{3}{*}{\parbox{1.5cm}{\centering \textbf{Text\\Imbalance}}} 
    & \cellcolor[HTML]{FFF1B8} $\text{LLM-TG}^{\star}$ & 
    $84.06_{\scriptstyle \pm 0.34}$ & 
    $80.42_{\scriptstyle \pm 0.36}$ & 
    $78.09_{\scriptstyle \pm 0.40}$ & 
    $74.25_{\scriptstyle \pm 0.25}$ & 
    $73.19_{\scriptstyle \pm 0.27}$ & 
    $70.22_{\scriptstyle \pm 0.28}$ & 
    $83.05_{\scriptstyle \pm 0.15}$ & 
    $80.89_{\scriptstyle \pm 0.17}$ & 
    $78.44_{\scriptstyle \pm 0.20}$ & 
    $82.65_{\scriptstyle \pm 0.68}$ & 
    $81.08_{\scriptstyle \pm 0.69}$ & 
    $79.38_{\scriptstyle \pm 0.72}$ \\  
    & \cellcolor[HTML]{FFF1B8} $\text{UltraTAG-S}^{\star}$ 
    & $\underline{88.34_{\scriptstyle \pm 0.73}}$ & $\underline{86.72_{\scriptstyle \pm 0.79}}$ & $\underline{82.96_{\scriptstyle \pm 0.82}}$ & $\underline{77.52_{\scriptstyle \pm 0.28}}$ & $\underline{76.27_{\scriptstyle \pm 0.34}}$ & $\underline{73.66_{\scriptstyle \pm 0.35}}$ & $\underline{83.97_{\scriptstyle \pm 0.41}}$ & $\underline{82.30_{\scriptstyle \pm 0.43}}$  & $\underline{80.43_{\scriptstyle \pm 0.43}}$ & $\underline{83.59_{\scriptstyle \pm 0.22}}$ & $\underline{83.01_{\scriptstyle \pm 0.24}}$ & $\underline{82.34_{\scriptstyle \pm 0.27}}$ \\  
    & \cellcolor[HTML]{FFF1B8} LAGA (Ours) & 
    \cellcolor[HTML]{DADADA}$\textbf{90.88}_{\scriptstyle \pm \textbf{0.66}}$ & 
    \cellcolor[HTML]{DADADA}$\textbf{88.26}_{\scriptstyle \pm \textbf{0.84}}$ & 
    \cellcolor[HTML]{DADADA}$\textbf{87.15}_{\scriptstyle \pm \textbf{0.84}}$ & 
    \cellcolor[HTML]{DADADA}$\textbf{83.05}_{\scriptstyle \pm \textbf{0.25}}$ & 
    \cellcolor[HTML]{DADADA}$\textbf{81.51}_{\scriptstyle \pm \textbf{0.27}}$ & 
    \cellcolor[HTML]{DADADA}$\textbf{79.36}_{\scriptstyle \pm \textbf{0.31}}$ & 
    \cellcolor[HTML]{DADADA}$\textbf{85.21}_{\scriptstyle \pm \textbf{0.23}}$ & 
    \cellcolor[HTML]{DADADA}$\textbf{84.39}_{\scriptstyle \pm \textbf{0.29}}$ & 
    \cellcolor[HTML]{DADADA}$\textbf{82.57}_{\scriptstyle \pm \textbf{0.35}}$ & 
    \cellcolor[HTML]{DADADA}$\textbf{87.22}_{\scriptstyle \pm \textbf{0.18}}$ & 
    \cellcolor[HTML]{DADADA}$\textbf{86.41}_{\scriptstyle \pm \textbf{0.21}}$ & 
    \cellcolor[HTML]{DADADA}$\textbf{85.27}_{\scriptstyle \pm \textbf{0.28}}$ \\  
    \midrule

    \multirow{3}{*}{\parbox{1.5cm}{\centering \textbf{Structure\\Sparsity}}}
    & \cellcolor[HTML]{F4FFB8} SUBLIME & 
    $81.39_{\scriptstyle \pm 0.34}$ & 
    $78.20_{\scriptstyle \pm 0.36}$ & 
    $73.45_{\scriptstyle \pm 0.39}$ & 
    $\underline{73.12_{\scriptstyle \pm 0.45}}$ & $70.69_{\scriptstyle \pm 0.45}$ & 
    $66.13_{\scriptstyle \pm 0.48}$ & 
    $81.29_{\scriptstyle \pm 0.34}$ & 
    $77.58_{\scriptstyle \pm 0.35}$ & 
    $70.65_{\scriptstyle \pm 0.40}$ & OOM & OOM & OOM \\  
    & \cellcolor[HTML]{F4FFB8} SEGSL & 
    $\underline{84.27_{\scriptstyle \pm 0.58}}$ & $\underline{82.63_{\scriptstyle \pm 0.60}}$ & $\underline{76.99_{\scriptstyle \pm 0.66}}$ & $73.08_{\scriptstyle \pm 0.23}$ & 
    $\underline{71.83_{\scriptstyle \pm 0.23}}$ & $\underline{68.54_{\scriptstyle \pm 0.26}}$ & $\underline{81.86_{\scriptstyle \pm 0.29}}$ & $\underline{81.86_{\scriptstyle \pm 0.29}}$ & $\underline{72.04_{\scriptstyle \pm 0.35}}$ & $\underline{82.99_{\scriptstyle \pm 0.30}}$ & $\underline{80.35_{\scriptstyle \pm 0.31}}$ & $\underline{74.49_{\scriptstyle \pm 0.35}}$ \\  
    & \cellcolor[HTML]{F4FFB8} LAGA (Ours) & 
    \cellcolor[HTML]{DADADA}$\textbf{88.48}_{\scriptstyle \pm \textbf{0.42}}$ & 
    \cellcolor[HTML]{DADADA}$\textbf{87.67}_{\scriptstyle \pm \textbf{0.44}}$ & 
    \cellcolor[HTML]{DADADA}$\textbf{84.59}_{\scriptstyle \pm \textbf{0.46}}$ & 
    \cellcolor[HTML]{DADADA}$\textbf{81.87}_{\scriptstyle \pm \textbf{0.11}}$ & 
    \cellcolor[HTML]{DADADA}$\textbf{80.72}_{\scriptstyle \pm \textbf{0.12}}$ & 
    \cellcolor[HTML]{DADADA}$\textbf{77.19}_{\scriptstyle \pm \textbf{0.14}}$ & 
    \cellcolor[HTML]{DADADA}$\textbf{84.69}_{\scriptstyle \pm \textbf{0.15}}$ & 
    \cellcolor[HTML]{DADADA}$\textbf{82.31}_{\scriptstyle \pm \textbf{0.14}}$ & 
    \cellcolor[HTML]{DADADA}$\textbf{76.88}_{\scriptstyle \pm \textbf{0.18}}$ & 
    \cellcolor[HTML]{DADADA}$\textbf{86.19}_{\scriptstyle \pm \textbf{0.09}}$ & 
    \cellcolor[HTML]{DADADA}$\textbf{83.63}_{\scriptstyle \pm \textbf{0.11}}$ & 
    \cellcolor[HTML]{DADADA}$\textbf{79.65}_{\scriptstyle \pm \textbf{0.15}}$ \\ 
    \midrule
    
    \multirow{3}{*}{\parbox{1.5cm}{\centering \textbf{Structure\\Noise}}}
    & \cellcolor[HTML]{E6FDD1} DHGR & 
    $80.76_{\scriptstyle \pm 0.44}$ & 
    $76.27_{\scriptstyle \pm 0.44}$ & 
    $70.95_{\scriptstyle \pm 0.46}$ & 
    $73.69_{\scriptstyle \pm 0.16}$ & 
    $68.52_{\scriptstyle \pm 0.18}$ & 
    $63.40_{\scriptstyle \pm 0.21}$ & 
    $76.25_{\scriptstyle \pm 0.22}$ & 
    $71.33_{\scriptstyle \pm 0.25}$ & $64.52_{\scriptstyle \pm 0.29}$ & $\underline{80.43_{\scriptstyle \pm 0.27}}$ & $\underline{76.29_{\scriptstyle \pm 0.31}}$ & $\underline{70.83_{\scriptstyle \pm 0.40}}$ \\
    & \cellcolor[HTML]{E6FDD1} $\text{LLM4RGNN}^\star$ & 
    $\underline{83.41_{\scriptstyle \pm 0.78}}$ & $\underline{80.80_{\scriptstyle \pm 0.84}}$ & $\underline{76.17_{\scriptstyle \pm 0.91}}$ & $\underline{74.12_{\scriptstyle \pm 0.58}}$ & $\underline{72.39_{\scriptstyle \pm 0.61}}$ & $\underline{70.87_{\scriptstyle \pm 0.67}}$ & $\underline{79.43_{\scriptstyle \pm 0.67}}$ & $\underline{75.16_{\scriptstyle \pm 0.38}}$ & $\underline{71.52_{\scriptstyle \pm 0.43}}$ & OOT & OOT & OOT \\
    & \cellcolor[HTML]{E6FDD1} LAGA (Ours) & 
    \cellcolor[HTML]{DADADA}$\textbf{86.56}_{\scriptstyle \pm \textbf{0.57}}$ & 
    \cellcolor[HTML]{DADADA}$\textbf{84.21}_{\scriptstyle \pm \textbf{0.60}}$ & 
    \cellcolor[HTML]{DADADA}$\textbf{80.10}_{\scriptstyle \pm \textbf{0.64}}$ & 
    \cellcolor[HTML]{DADADA}$\textbf{78.45}_{\scriptstyle \pm \textbf{0.35}}$ & 
    \cellcolor[HTML]{DADADA}$\textbf{76.33}_{\scriptstyle \pm \textbf{0.36}}$ & 
    \cellcolor[HTML]{DADADA}$\textbf{71.70}_{\scriptstyle \pm \textbf{0.39}}$ & 
    \cellcolor[HTML]{DADADA}$\textbf{82.67}_{\scriptstyle \pm \textbf{0.19}}$ & 
    \cellcolor[HTML]{DADADA}$\textbf{77.36}_{\scriptstyle \pm \textbf{0.19}}$ & 
    \cellcolor[HTML]{DADADA}$\textbf{73.99}_{\scriptstyle \pm \textbf{0.20}}$ & 
    \cellcolor[HTML]{DADADA}$\textbf{83.93}_{\scriptstyle \pm \textbf{0.23}}$ & 
    \cellcolor[HTML]{DADADA}$\textbf{80.84}_{\scriptstyle \pm \textbf{0.27}}$ & 
    \cellcolor[HTML]{DADADA}$\textbf{76.57}_{\scriptstyle \pm \textbf{0.30}}$ \\
    
    \midrule

    \multirow{3}{*}{\parbox{1.5cm}{\centering \textbf{Structure\\Imbalance}}}                   
    & \cellcolor[HTML]{D9F7BE} RawlsGCN & 
    $81.72_{\scriptstyle \pm 0.24}$ & 
    $80.05_{\scriptstyle \pm 0.17}$ & 
    $\underline{79.66_{\scriptstyle \pm 0.18}}$ & $76.34_{\scriptstyle \pm 0.38}$ & 
    $74.01_{\scriptstyle \pm 0.44}$ & 
    $72.72_{\scriptstyle \pm 0.73}$ & 
    $\underline{83.79_{\scriptstyle \pm 0.27}}$ & $\underline{82.81_{\scriptstyle \pm 0.13}}$ & $82.12_{\scriptstyle \pm 0.13}$ & 
    $84.53_{\scriptstyle \pm 0.15}$ & 
    $\underline{84.18_{\scriptstyle \pm 1.22}}$ & $\underline{82.99_{\scriptstyle \pm 1.14}}$ \\
    & \cellcolor[HTML]{D9F7BE} GraphPatcher & $\underline{84.24_{\scriptstyle \pm 0.55}}$ & $\underline{82.78_{\scriptstyle \pm 0.43}}$ & $79.27_{\scriptstyle \pm 0.27}$ & $\underline{77.53_{\scriptstyle \pm 0.31}}$ & $\underline{76.92_{\scriptstyle \pm 0.44}}$ & $\underline{77.16_{\scriptstyle \pm 1.14}}$ & $83.58_{\scriptstyle \pm 0.45}$ & 
    $82.64_{\scriptstyle \pm 0.11}$ & 
    $\underline{82.14_{\scriptstyle \pm 0.15}}$ & $\underline{85.20_{\scriptstyle \pm 0.34}}$ & $83.28_{\scriptstyle \pm 0.12}$ & 
    $82.57_{\scriptstyle \pm 0.24}$ \\
    & \cellcolor[HTML]{D9F7BE} LAGA (Ours) & 
    \cellcolor[HTML]{DADADA}$\textbf{87.92}_{\scriptstyle \pm \textbf{0.51}}$ & 
    \cellcolor[HTML]{DADADA}$\textbf{87.63}_{\scriptstyle \pm \textbf{0.55}}$ & 
    \cellcolor[HTML]{DADADA}$\textbf{86.95}_{\scriptstyle \pm \textbf{0.58}}$ & 
    \cellcolor[HTML]{DADADA}$\textbf{82.72}_{\scriptstyle \pm \textbf{0.16}}$ & 
    \cellcolor[HTML]{DADADA}$\textbf{81.62}_{\scriptstyle \pm \textbf{0.17}}$ & 
    \cellcolor[HTML]{DADADA}$\textbf{78.11}_{\scriptstyle \pm \textbf{0.20}}$ & 
    \cellcolor[HTML]{DADADA}$\textbf{85.29}_{\scriptstyle \pm \textbf{0.18}}$ & 
    \cellcolor[HTML]{DADADA}$\textbf{83.42}_{\scriptstyle \pm \textbf{0.23}}$ & 
    \cellcolor[HTML]{DADADA}$\textbf{82.50}_{\scriptstyle \pm \textbf{0.27}}$ & 
    \cellcolor[HTML]{DADADA}$\textbf{86.17}_{\scriptstyle \pm \textbf{0.36}}$ & 
    \cellcolor[HTML]{DADADA}$\textbf{84.76}_{\scriptstyle \pm \textbf{0.39}}$ & 
    \cellcolor[HTML]{DADADA}$\textbf{83.59}_{\scriptstyle \pm \textbf{0.44}}$ \\
    \midrule
    
    \multirow{3}{*}{\parbox{1.5cm}{\centering \textbf{Label\\Sparsity}}}   
    & \cellcolor[HTML]{E6F7FF} GraFN & 
    $\underline{82.73_{\scriptstyle \pm 0.43}}$ & $79.51_{\scriptstyle \pm 0.41}$ & 
    $74.20_{\scriptstyle \pm 0.46}$ & 
    $\underline{75.04_{\scriptstyle \pm 0.27}}$ & $\underline{73.48_{\scriptstyle \pm 0.29}}$ & $69.74_{\scriptstyle \pm 0.30}$ & 
    $\underline{83.74_{\scriptstyle \pm 0.24}}$ & $\underline{82.91_{\scriptstyle \pm 0.25}}$ & $\underline{82.41_{\scriptstyle \pm 0.28}}$ & $\underline{84.36_{\scriptstyle \pm 0.19}}$ & $\underline{82.25_{\scriptstyle \pm 0.23}}$ & $78.46_{\scriptstyle \pm 0.26}$ \\
    & \cellcolor[HTML]{E6F7FF} GraphHop & 
    $82.30_{\scriptstyle \pm 0.49}$ & $\underline{80.12_{\scriptstyle \pm 0.51}}$ & $\underline{76.31_{\scriptstyle \pm 0.55}}$ & $74.60_{\scriptstyle \pm 0.25}$ & 
    $72.78_{\scriptstyle \pm 0.26}$ & $\underline{70.11_{\scriptstyle \pm 0.28}}$ & $83.60_{\scriptstyle \pm 0.15}$ & 
    $81.40_{\scriptstyle \pm 0.17}$ & 
    $80.36_{\scriptstyle \pm 0.18}$ & 
    $83.91_{\scriptstyle \pm 0.23}$ & 
    $82.23_{\scriptstyle \pm 0.23}$ & $\underline{79.70_{\scriptstyle \pm 0.24}}$  \\
    & \cellcolor[HTML]{E6F7FF} LAGA (Ours) & 
    \cellcolor[HTML]{DADADA}$\textbf{89.48}_{\scriptstyle \pm \textbf{0.25}}$ & 
    \cellcolor[HTML]{DADADA}$\textbf{88.19}_{\scriptstyle \pm \textbf{0.26}}$ & 
    \cellcolor[HTML]{DADADA}$\textbf{86.90}_{\scriptstyle \pm \textbf{0.24}}$ & 
    \cellcolor[HTML]{DADADA}$\textbf{82.75}_{\scriptstyle \pm \textbf{0.13}}$ & 
    \cellcolor[HTML]{DADADA}$\textbf{83.22}_{\scriptstyle \pm \textbf{0.14}}$ & 
    \cellcolor[HTML]{DADADA}$\textbf{82.48}_{\scriptstyle \pm \textbf{0.16}}$ & 
    \cellcolor[HTML]{DADADA}$\textbf{86.12}_{\scriptstyle \pm \textbf{0.20}}$ & 
    \cellcolor[HTML]{DADADA}$\textbf{85.67}_{\scriptstyle \pm \textbf{0.21}}$ & 
    \cellcolor[HTML]{DADADA}$\textbf{84.30}_{\scriptstyle \pm \textbf{0.23}}$ & 
    \cellcolor[HTML]{DADADA}$\textbf{87.11}_{\scriptstyle \pm \textbf{0.05}}$ & 
    \cellcolor[HTML]{DADADA}$\textbf{85.21}_{\scriptstyle \pm \textbf{0.06}}$ & 
    \cellcolor[HTML]{DADADA}$\textbf{81.29}_{\scriptstyle \pm \textbf{0.10}}$ \\
    \midrule

    \multirow{3}{*}{\parbox{1.5cm}{\centering 
    \textbf{Label\\Noise}}}   
    & \cellcolor[HTML]{F0F5FF} PI-GNN & 
    $77.39_{\scriptstyle \pm 1.24}$ & 
    $74.10_{\scriptstyle \pm 1.31}$ & 
    $51.62_{\scriptstyle \pm 1.73}$ & 
    $74.54_{\scriptstyle \pm 0.66}$ & 
    $68.37_{\scriptstyle \pm 0.64}$ & 
    $48.53_{\scriptstyle \pm 0.70}$ & 
    $\underline{82.13_{\scriptstyle \pm 0.54}}$ & $\underline{80.67_{\scriptstyle \pm 0.59}}$ & $\underline{72.59_{\scriptstyle \pm 0.61}}$ & $\underline{80.34_{\scriptstyle \pm 1.05}}$ & $\underline{73.51_{\scriptstyle \pm 1.36}}$ & $\underline{51.08_{\scriptstyle \pm 1.92}}$ \\ 
    & \cellcolor[HTML]{F0F5FF} NRGNN & $\underline{80.73_{\scriptstyle \pm 0.86}}$ & $\underline{76.23_{\scriptstyle \pm 0.88}}$ & $\underline{58.96_{\scriptstyle \pm 0.91}}$ & $\underline{75.82_{\scriptstyle \pm 0.43}}$ & $\underline{69.11_{\scriptstyle \pm 0.44}}$ & $\underline{51.39_{\scriptstyle \pm 0.46}}$ & $81.57_{\scriptstyle \pm 0.61}$ & 
    $79.43_{\scriptstyle \pm 0.63}$ & 
    $70.03_{\scriptstyle \pm 0.63}$ & 
    $78.31_{\scriptstyle \pm 1.02}$ & 
    $72.18_{\scriptstyle \pm 1.10}$ & 
    $46.39_{\scriptstyle \pm 1.33}$ \\
    & \cellcolor[HTML]{F0F5FF} LAGA (Ours) & 
    \cellcolor[HTML]{DADADA}$\textbf{88.56}_{\scriptstyle \pm \textbf{0.75}}$ & 
    \cellcolor[HTML]{DADADA}$\textbf{87.63}_{\scriptstyle \pm \textbf{0.79}}$ & 
    \cellcolor[HTML]{DADADA}$\textbf{83.50}_{\scriptstyle \pm \textbf{0.86}}$ & 
    \cellcolor[HTML]{DADADA}$\textbf{79.08}_{\scriptstyle \pm \textbf{0.69}}$ & 
    \cellcolor[HTML]{DADADA}$\textbf{75.27}_{\scriptstyle \pm \textbf{0.71}}$ & 
    \cellcolor[HTML]{DADADA}$\textbf{70.94}_{\scriptstyle \pm \textbf{0.74}}$ & 
    \cellcolor[HTML]{DADADA}$\textbf{84.11}_{\scriptstyle \pm \textbf{0.81}}$ & 
    \cellcolor[HTML]{DADADA}$\textbf{82.69}_{\scriptstyle \pm \textbf{0.83}}$ & 
    \cellcolor[HTML]{DADADA}$\textbf{76.14}_{\scriptstyle \pm \textbf{0.87}}$ & 
    \cellcolor[HTML]{DADADA}$\textbf{81.66}_{\scriptstyle \pm \textbf{0.91}}$ & 
    \cellcolor[HTML]{DADADA}$\textbf{76.53}_{\scriptstyle \pm \textbf{1.01}}$ & 
    \cellcolor[HTML]{DADADA}$\textbf{58.69}_{\scriptstyle \pm \textbf{1.08}}$ \\
    \midrule

    \multirow{3}{*}{\parbox{1.5cm}{\centering \textbf{Label\\Imbalance}}}   
    & \cellcolor[HTML]{D6E5FF} LTE4G & 
    $81.25_{\scriptstyle \pm 0.43}$ & 
    $79.92_{\scriptstyle \pm 0.45}$ & 
    $74.48_{\scriptstyle \pm 0.48}$ & 
    $72.57_{\scriptstyle \pm 0.64}$ & 
    $67.20_{\scriptstyle \pm 0.92}$ & 
    $60.88_{\scriptstyle \pm 1.03}$ & 
    $76.27_{\scriptstyle \pm 0.37}$ & 
    $74.03_{\scriptstyle \pm 0.41}$ & 
    $71.54_{\scriptstyle \pm 0.48}$ & 
    $78.85_{\scriptstyle \pm 0.54}$ & 
    $79.15_{\scriptstyle \pm 0.38}$ & 
    $74.29_{\scriptstyle \pm 0.49}$ \\
    & \cellcolor[HTML]{D6E5FF} TOPOAUC & 
    $\underline{84.03_{\scriptstyle \pm 0.08}}$ & $\underline{81.34_{\scriptstyle \pm 0.12}}$ & $\underline{76.52_{\scriptstyle \pm 0.13}}$ & $\underline{75.03_{\scriptstyle \pm 0.21}}$ & $\underline{72.59_{\scriptstyle \pm 0.22}}$ & $\underline{68.17_{\scriptstyle \pm 0.26}}$ & $\underline{77.39_{\scriptstyle \pm 0.36}}$ & $\underline{76.14_{\scriptstyle \pm 0.37}}$ & 
    $\underline{74.58_{\scriptstyle \pm 0.40}}$ & 
    $\underline{80.21_{\scriptstyle \pm 1.87}}$ & $\underline{80.79_{\scriptstyle \pm 2.02}}$ & $\underline{76.45_{\scriptstyle \pm 2.23}}$ \\
    & \cellcolor[HTML]{D6E5FF} LAGA (Ours) & 
    \cellcolor[HTML]{DADADA}$\textbf{87.04}_{\scriptstyle \pm \textbf{0.57}}$ & 
    \cellcolor[HTML]{DADADA}$\textbf{84.61}_{\scriptstyle \pm \textbf{0.59}}$ & 
    \cellcolor[HTML]{DADADA}$\textbf{79.06}_{\scriptstyle \pm \textbf{0.60}}$ & 
    \cellcolor[HTML]{DADADA}$\textbf{78.03}_{\scriptstyle \pm \textbf{0.34}}$ & 
    \cellcolor[HTML]{DADADA}$\textbf{74.72}_{\scriptstyle \pm \textbf{0.34}}$ & 
    \cellcolor[HTML]{DADADA}$\textbf{70.33}_{\scriptstyle \pm \textbf{0.37}}$ & 
    \cellcolor[HTML]{DADADA}$\textbf{80.16}_{\scriptstyle \pm \textbf{0.46}}$ & 
    \cellcolor[HTML]{DADADA}$\textbf{79.31}_{\scriptstyle \pm \textbf{0.44}}$ & 
    \cellcolor[HTML]{DADADA}$\textbf{76.55}_{\scriptstyle \pm \textbf{0.49}}$ & 
    \cellcolor[HTML]{DADADA}$\textbf{82.91}_{\scriptstyle \pm \textbf{0.82}}$ & 
    \cellcolor[HTML]{DADADA}$\textbf{81.59}_{\scriptstyle \pm \textbf{1.01}}$ & 
    \cellcolor[HTML]{DADADA}$\textbf{77.25}_{\scriptstyle \pm \textbf{1.12}}$ \\
    
    \specialrule{1.3pt}{2.0pt}{1.0pt}
    \end{tabular}}
    \vspace{-13pt}

    \label{table:node_cls}
    \end{table*}
    \vspace{-5pt}

\section{Experiments}
    In this section, we conduct a wide range of experiments and aim to answer the following questions: 
    \textbf{Q1: Effectiveness.}
    Compared with state-of-the-art baselines, can LAGA consistently achieve superior performance across diverse quality degradation scenarios?
    \textbf{Q2: Ablation.}
    If LAGA demonstrates effectiveness, what contributes to its outstanding performance? 
    \textbf{Q3: Robustness.}
    How robust is LAGA across different backbone models, under varying hyperparameter settings, and in combined degradation scenarios
    \textbf{Q4: interpretability.}
    Can LAGA offer strong interpretability in its quality optimization process?
    \textbf{Q5: Scalability.} 
    Can LAGA scale efficiently to large-scale TAGs while maintaining competitive performance?

    In the experimental section, we selected 5 mainstream TAG datasets and 16 well-performing baselines. We run all our experiments on an 80G A100 GPU, using Gemma-3-27b as the base LLM and Sentence-BERT as the base LM encoder, with LLaMA3-8B used for fine-tuning. Detailed hyperparameter settings, dataset descriptions, and baseline introductions can be found in the Appendix~\ref{appendix:hyperpara_setting}-\ref{appendix:baseline}.
    In addition, we consider 9 quality issue scenarios in each experiment. The specific scenario settings are provided in the Appendix~\ref{appendix:scene_set}.

\vspace{-3pt}
\subsection{Effectiveness (Q1)}
    \vspace{-3pt}
    \textbf{Node Classification.} 
    To address \textbf{Q1}, the results in Table~\ref{table:node_cls} show that LAGA outperforms all baselines across the three quality dimensions. In the \emph{text dimension}, LAGA improves UltraTAG-S by 3.71\% on Cora with text sparsity and CTD-MLM by 2.31\% on WikiCS with text noise. In the \emph{structure dimension}, it surpasses SE-GSL by 3.73\% on Citeseer with structure sparsity and LLM4RGNN by 1.64\% on WikiCS with structure noise. In the \emph{label dimension}, LAGA achieves 6.11\% higher accuracy than NRGNN on WikiCS with label noise and 2.33\% higher than TOPOAUC on Photo with label imbalance. These results demonstrate that LAGA delivers state-of-the-art performance across various scenarios, confirming its effectiveness.
    
    \textbf{Other Tasks and Backbones.}  
    To comprehensively address \textbf{Q1}, we extended our evaluation to other downstream tasks across various GNN backbones to further validate LAGA's effectiveness. These experiments provide additional insights into LAGA's effectiveness across diverse downstream tasks and GNN backbones. Due to space constraints, detailed results and experimental analysis are included in the Appendix~\ref{appendix:effectiveness}.

\subsection{Ablation Study (\textbf{Q2})}
\vspace{-3pt}
    To answer \textbf{Q2}, we conduct an ablation study on the Citeseer and Photo datasets to evaluate the contribution of core LAGA components, as detailed in Table~\ref{table:ablation}. We replace the LLM-driven planner with a fixed pipeline in \textbf{w/o Dynamic Planning}, and use equal loss weights in \textbf{w/o Dynamic Weights}. We also remove specific objectives in graph learning (\textbf{w/o Label/Semantic/Structure Loss}) and disable the closed-loop feedback (\textbf{w/o Evaluation Agent}) to verify their individual contributions.

    \begin{table*}[htbp]
    \centering
    \caption{Ablation study under different quality degradation scenarios with perturbation ratio = 0.4. For quality problem types, ``Spa'' denotes \emph{Sparsity}, ``Noi'' denotes \emph{Noise}, and ``Imb'' denotes \emph{Imbalance}.}
    \resizebox{\textwidth}{!}{
    \begin{tabular}{c|c|c c c c c c c c c c}
    \specialrule{1.5pt}{1.5pt}{1.5pt}
    \multicolumn{2}{c|}{\textbf{Scenario}} &
    \textbf{Original} &
    \textbf{Text-Spa} &
    \textbf{Text-Noi} &
    \textbf{Text-Imb} &
    \textbf{Structure-Spa} &
    \textbf{Structure-Noi} &
    \textbf{Structure-Imb} &
    \textbf{Label-Spa} &
    \textbf{Label-Noi} &
    \textbf{Label-Imb} \\
    \midrule

    \multirow{7}{*}{\textbf{Citeseer}} 
    & w/o Dynamic Planning
    & $80.42_{\scriptstyle \pm 0.33}$ 
    & $78.14_{\scriptstyle \pm 0.28}$ 
    & $79.05_{\scriptstyle \pm 0.26}$ 
    & $79.65_{\scriptstyle \pm 0.35}$ 
    & $77.64_{\scriptstyle \pm 0.24}$ 
    & $76.24_{\scriptstyle \pm 0.41}$ 
    & $79.09_{\scriptstyle \pm 0.35}$ 
    & $81.03_{\scriptstyle \pm 0.28}$ 
    & $71.92_{\scriptstyle \pm 0.51}$ 
    & $73.13_{\scriptstyle \pm 0.44}$ \\
    & w/o Dynamic Weights
    & $79.38_{\scriptstyle \pm 0.37}$ 
    & $77.51_{\scriptstyle \pm 0.21}$ 
    & $78.35_{\scriptstyle \pm 0.29}$ 
    & $78.44_{\scriptstyle \pm 0.40}$ 
    & $76.49_{\scriptstyle \pm 0.33}$ 
    & $73.67_{\scriptstyle \pm 0.45}$ 
    & $78.35_{\scriptstyle \pm 0.29}$ 
    & $80.14_{\scriptstyle \pm 0.30}$ 
    & $70.91_{\scriptstyle \pm 0.71}$ 
    & $72.65_{\scriptstyle \pm 0.43}$ \\
    & w/o Label Loss 
    & $75.21_{\scriptstyle \pm 0.54}$ 
    & $73.25_{\scriptstyle \pm 0.37}$ 
    & $71.40_{\scriptstyle \pm 0.31}$ 
    & $73.83_{\scriptstyle \pm 0.41}$ 
    & $70.59_{\scriptstyle \pm 0.26}$ 
    & $67.30_{\scriptstyle \pm 0.32}$ 
    & $73.15_{\scriptstyle \pm 0.28}$ 
    & $72.68_{\scriptstyle \pm 0.22}$ 
    & $64.65_{\scriptstyle \pm 0.83}$ 
    & $65.43_{\scriptstyle \pm 0.52}$ \\
    & w/o Semantic Loss 
    & $80.77_{\scriptstyle \pm 0.41}$ 
    & $78.68_{\scriptstyle \pm 0.26}$ 
    & $79.34_{\scriptstyle \pm 0.20}$ 
    & $79.91_{\scriptstyle \pm 0.31}$ 
    & $78.44_{\scriptstyle \pm 0.21}$ 
    & $74.23_{\scriptstyle \pm 0.39}$ 
    & $79.65_{\scriptstyle \pm 0.25}$ 
    & $82.71_{\scriptstyle \pm 0.33}$ 
    & $73.13_{\scriptstyle \pm 0.76}$ 
    & $72.89_{\scriptstyle \pm 0.52}$ \\
    & w/o Structure Loss 
    & $81.17_{\scriptstyle \pm 0.45}$ 
    & $77.93_{\scriptstyle \pm 0.25}$ 
    & $78.69_{\scriptstyle \pm 0.18}$ 
    & $79.83_{\scriptstyle \pm 0.27}$ 
    & $79.31_{\scriptstyle \pm 0.18}$ 
    & $75.74_{\scriptstyle \pm 0.40}$ 
    & $80.04_{\scriptstyle \pm 0.22}$ 
    & $81.05_{\scriptstyle \pm 0.16}$ 
    & $71.78_{\scriptstyle \pm 0.72}$ 
    & $71.15_{\scriptstyle \pm 0.35}$ \\
    & w/o Evaluation Agent 
    & $81.76_{\scriptstyle \pm 0.48}$ 
    & $78.12_{\scriptstyle \pm 0.27}$ 
    & $80.13_{\scriptstyle \pm 0.23}$ 
    & $80.77_{\scriptstyle \pm 0.28}$ 
    & $79.15_{\scriptstyle \pm 0.32}$ 
    & $72.28_{\scriptstyle \pm 0.48}$ 
    & $79.86_{\scriptstyle \pm 0.41}$ 
    & $80.07_{\scriptstyle \pm 0.28}$ 
    & $70.33_{\scriptstyle \pm 1.03}$ 
    & $74.26_{\scriptstyle \pm 0.66}$ \\
    \cmidrule(lr){2-12} 
    
    & LAGA (complete)
    & \cellcolor[HTML]{DADADA}$\textbf{83.54}_{\scriptstyle \pm \textbf{0.19}}$ 
    & \cellcolor[HTML]{DADADA}$\textbf{79.73}_{\scriptstyle \pm \textbf{0.24}}$ 
    & \cellcolor[HTML]{DADADA}$\textbf{81.34}_{\scriptstyle \pm \textbf{0.17}}$ 
    & \cellcolor[HTML]{DADADA}$\textbf{81.51}_{\scriptstyle \pm \textbf{0.27}}$ 
    & \cellcolor[HTML]{DADADA}$\textbf{80.72}_{\scriptstyle \pm \textbf{0.12}}$ 
    & \cellcolor[HTML]{DADADA}$\textbf{76.33}_{\scriptstyle \pm \textbf{0.36}}$ 
    & \cellcolor[HTML]{DADADA}$\textbf{81.62}_{\scriptstyle \pm \textbf{0.17}}$ 
    & \cellcolor[HTML]{DADADA}$\textbf{83.22}_{\scriptstyle \pm \textbf{0.14}}$ 
    & \cellcolor[HTML]{DADADA}$\textbf{75.27}_{\scriptstyle \pm \textbf{0.71}}$ 
    & \cellcolor[HTML]{DADADA}$\textbf{74.72}_{\scriptstyle \pm \textbf{0.34}}$ \\
    \midrule

    \multirow{7}{*}{\textbf{Photo}} 
    & w/o Dynamic Planning 
    & $85.92_{\scriptstyle \pm 0.32}$ 
    & $81.14_{\scriptstyle \pm 0.41}$ 
    & $82.64_{\scriptstyle \pm 0.19}$ 
    & $81.31_{\scriptstyle \pm 0.30}$ 
    & $82.04_{\scriptstyle \pm 0.25}$ 
    & $79.37_{\scriptstyle \pm 0.33}$ 
    & $81.55_{\scriptstyle \pm 0.41}$ 
    & $83.79_{\scriptstyle \pm 0.17}$ 
    & $73.02_{\scriptstyle \pm 0.72}$ 
    & $79.20_{\scriptstyle \pm 0.70}$ \\
    & w/o Dynamic Weights 
    & $84.61_{\scriptstyle \pm 0.40}$ 
    & $80.33_{\scriptstyle \pm 0.51}$ 
    & $81.62_{\scriptstyle \pm 0.36}$ 
    & $81.39_{\scriptstyle \pm 0.33}$ 
    & $81.15_{\scriptstyle \pm 0.27}$ 
    & $78.92_{\scriptstyle \pm 0.34}$ 
    & $80.71_{\scriptstyle \pm 0.51}$ 
    & $83.09_{\scriptstyle \pm 0.22}$ 
    & $72.15_{\scriptstyle \pm 0.71}$ 
    & $78.50_{\scriptstyle \pm 0.69}$ \\
    & w/o Label Loss 
    & $82.62_{\scriptstyle \pm 0.25}$ 
    & $79.26_{\scriptstyle \pm 0.21}$ 
    & $79.80_{\scriptstyle \pm 0.11}$  
    & $78.96_{\scriptstyle \pm 0.30}$ 
    & $76.54_{\scriptstyle \pm 0.18}$ 
    & $72.66_{\scriptstyle \pm 0.38}$ 
    & $76.83_{\scriptstyle \pm 0.42}$ 
    & $79.76_{\scriptstyle \pm 0.26}$ 
    & $70.34_{\scriptstyle \pm 1.16}$ 
    & $72.65_{\scriptstyle \pm 1.18}$ \\
    & w/o Semantic Loss 
    & $85.88_{\scriptstyle \pm 0.14}$ 
    & $83.66_{\scriptstyle \pm 0.17}$ 
    & $85.21_{\scriptstyle \pm 0.09}$ 
    & $84.96_{\scriptstyle \pm 0.25}$ 
    & $82.15_{\scriptstyle \pm 0.13}$ 
    & $78.69_{\scriptstyle \pm 0.31}$ 
    & $81.74_{\scriptstyle \pm 0.41}$ 
    & $82.53_{\scriptstyle \pm 0.12}$ 
    & $70.42_{\scriptstyle \pm 0.88}$ 
    & $77.64_{\scriptstyle \pm 0.85}$ \\
    & w/o Structure Loss 
    & $85.42_{\scriptstyle \pm 0.18}$ 
    & $83.32_{\scriptstyle \pm 0.21}$ 
    & $84.87_{\scriptstyle \pm 0.11}$ 
    & $84.51_{\scriptstyle \pm 0.30}$ 
    & $82.03_{\scriptstyle \pm 0.18}$ 
    & $78.72_{\scriptstyle \pm 0.34}$ 
    & $81.52_{\scriptstyle \pm 0.46}$ 
    & $82.13_{\scriptstyle \pm 0.22}$ 
    & $70.28_{\scriptstyle \pm 1.09}$ 
    & $75.81_{\scriptstyle \pm 1.12}$ \\
    & w/o Evaluation Agent 
    & $86.21_{\scriptstyle \pm 0.28}$ 
    & $83.84_{\scriptstyle \pm 0.26}$ 
    & $85.15_{\scriptstyle \pm 0.15}$ 
    & $85.36_{\scriptstyle \pm 0.28}$ 
    & $82.79_{\scriptstyle \pm 0.26}$ 
    & $79.33_{\scriptstyle \pm 0.48}$ 
    & $82.41_{\scriptstyle \pm 0.51}$ 
    & $84.32_{\scriptstyle \pm 0.31}$ 
    & $73.61_{\scriptstyle \pm 1.26}$ 
    & $80.18_{\scriptstyle \pm 1.18}$ \\
    \cmidrule(lr){2-12}
    
    & LAGA (complete)
    & \cellcolor[HTML]{DADADA}$\textbf{88.27}_{\scriptstyle \pm \textbf{0.14}}$ 
    & \cellcolor[HTML]{DADADA}$\textbf{85.38}_{\scriptstyle \pm \textbf{0.18}}$ 
    & \cellcolor[HTML]{DADADA}$\textbf{86.02}_{\scriptstyle \pm \textbf{0.06}}$
    & \cellcolor[HTML]{DADADA}$\textbf{86.41}_{\scriptstyle \pm \textbf{0.28}}$ 
    & \cellcolor[HTML]{DADADA}$\textbf{83.63}_{\scriptstyle \pm \textbf{0.15}}$ 
    & \cellcolor[HTML]{DADADA}$\textbf{80.84}_{\scriptstyle \pm \textbf{0.30}}$ 
    & \cellcolor[HTML]{DADADA}$\textbf{84.76}_{\scriptstyle \pm \textbf{0.44}}$ 
    & \cellcolor[HTML]{DADADA}$\textbf{85.21}_{\scriptstyle \pm \textbf{0.10}}$ 
    & \cellcolor[HTML]{DADADA}$\textbf{76.53}_{\scriptstyle \pm \textbf{1.08}}$ 
    & \cellcolor[HTML]{DADADA}$\textbf{81.59}_{\scriptstyle \pm \textbf{1.12}}$ \\

    \specialrule{1.3pt}{2.0pt}{1.0pt}
    \end{tabular}}
    \vspace{-3pt}
    
    \label{table:ablation}
    \end{table*}

    Results show that all components contribute to LAGA's success. Removing \textbf{label loss} causes the sharpest drop, confirming its primary supervisory role. Crucially, \textbf{w/o Dynamic Planning} and \textbf{w/o Dynamic Weights} consistently underperform the full model, especially in high-noise scenarios (e.g., Structure-Noi), validating the necessity of adaptive strategy and reliability-aware weighting over static baselines. \textbf{Semantic} and \textbf{structure losses} prove vital for modality alignment, with significant drops observed in their corresponding defect scenarios. Furthermore, disabling the \textbf{evaluation agent} breaks the closed-loop feedback, preventing iterative quality gains (see Appendix~\ref{appendix:convergence} for convergence analysis), confirming the value of continuous refinement. See Appendix~\ref{appendix:llm_backbone} for LLM backbone comparisons.

    \begin{figure*}[t]
        \centering
        \includegraphics[width=\textwidth]{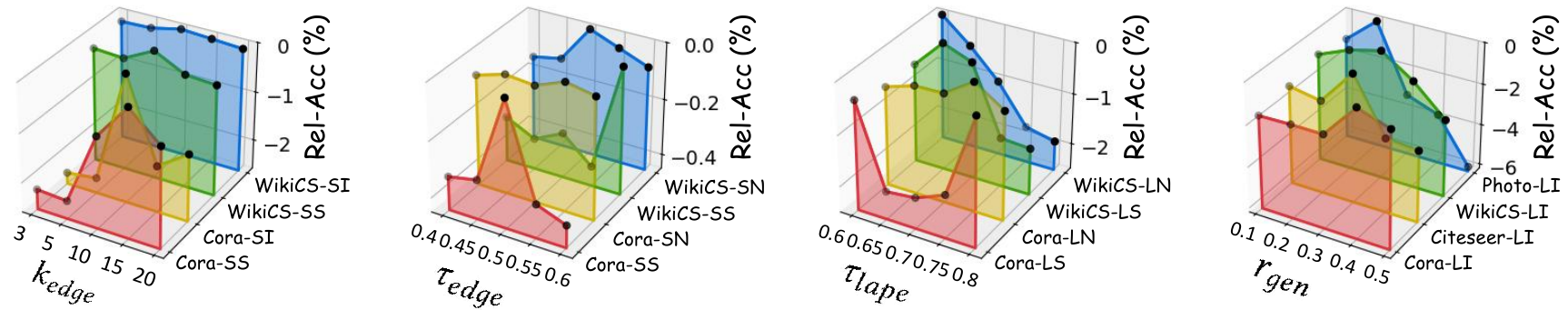}
        \caption{The sensitivity of LAGA to different hyperparameters $(k_{edge}, \tau_{edge}, \tau_{lape}, r_{gen})$. ``Dataset-Scene'' denotes different datasets and experimental scenarios, while ``Rel-Acc'' denotes the relative accuracy normalized by the best-performing baseline.}
        \label{figure:hyperparameter}
    \end{figure*}

\subsection{Robustness (Q3)}
    To answer \textbf{Q3}, we conduct a thorough analysis of the LAGA’s robustness from the following aspects:

    \textbf{Hyperparameters.}
    To assess the robustness of LAGA, we evaluate its sensitivity to key hyperparameters, varying $k_{\text{edge}}$, $\tau_{\text{edge}}$, $\tau_{\text{lape}}$, and $r_{\text{gen}}$ (Figure~\ref{figure:hyperparameter}). LAGA remains stable across a wide range of values, with only minor fluctuations in relative accuracy.
    The parameter $k_{\text{edge}}$ controls edge addition during structure optimization. A moderate range (10–15) provides the best balance, avoiding insufficient sparsity alleviation or noisy connections. $\tau_{\text{edge}}$ determines edge retention; $\tau_{\text{edge}} = 0.5$ consistently yields stable performance. For label optimization, $\tau_{\text{lape}}$ controls the confidence for accepting pseudo-labels. Values between 0.7–0.8 provide robust performance. Finally, $r_{\text{gen}}$ controls node generation for minority classes, with ratios between 0.2 and 0.4 offering an optimal trade-off. These results indicate that LAGA is not overly sensitive to hyperparameter tuning.
    Additionally, detection hyperparameters are fixed based on preliminary experiments showing minimal sensitivity, ensuring consistent evaluation and allowing us to focus on tuning the optimization module.

    \begin{wrapfigure}{r}{0.49\textwidth}
        \centering
        \vspace{-3pt}
        \includegraphics[width=\linewidth]{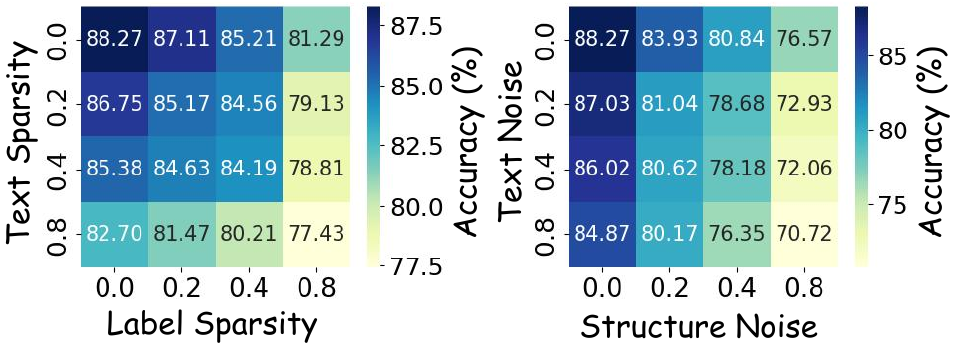} 
        \vspace{-10pt}
        \caption{LAGA's performance across composite scenarios with varying perturbation ratios.}
        \vspace{-5pt}
        \label{figure:composite_scene2}
    \end{wrapfigure}

    \textbf{Composite Scenarios.}
    In addition to single-type degradations, we evaluate LAGA under four composite scenarios that reflect real-world data situations (Figure~\ref{figure:composite_scene2}). These scenarios are chosen based on practical cases where multiple quality issues often occur simultaneously. Experimental results show that LAGA performs consistently well across various perturbation ratios in all composite scenarios. The accuracy decreases only moderately with increasing degradation, demonstrating that LAGA effectively handles multiple simultaneous issues. More results from composite scenarios are in the Appendix~\ref{appendix:composite_scene}. This highlights the robustness and broad applicability of LAGA in realistic settings.

\subsection{Interpretability (Q4)}
    To answer \textbf{Q4}, we complement quantitative results with a human-centered expert evaluation of graph quality. We invited $R=50$ domain experts to assess a sample of $N=200$ nodes and their neighborhoods before and after optimization. Experts scored the graphs along three dimensions—\textbf{Textual Adequacy}, \textbf{Structural Coherence}, and \textbf{Label Reliability}—on a scale from 0 to 100.
    The overall expert evaluation score, termed \emph{Quality Score}, is defined as:
    \begin{equation}
        Q_{\text{score}} = \frac{1}{R \cdot N}\sum_{r=1}^{R}\sum_{i=1}^{N} 
    \left( \frac{1}{3}\sum_{d=1}^{3} t_{r,i,d} \right),
    \end{equation}
    where $t_{r,i,d}$ denotes the score given by expert $r$ for instance $i$ on the $d$-th dimension. 

    \begin{wrapfigure}{r}{0.5\textwidth}
        \centering
        \includegraphics[width=\linewidth]{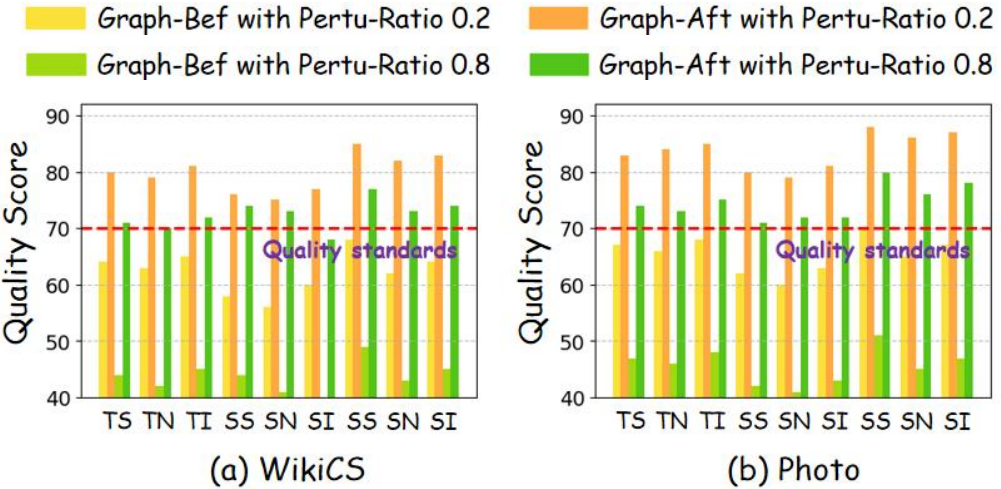} 
        \caption{Expert evaluation results under nine scenarios (Pertu - Perturbation).}
        \label{figure:interpretability2}
    \end{wrapfigure}
    
    Figure~\ref{figure:interpretability2} presents the expert evaluation results for WikiCS and Photo datasets. In both datasets, the optimized graphs (\emph{Graph-Aft}) consistently score higher than the original graphs (\emph{Graph-Bef}) across all scenarios. For instance, on WikiCS with a perturbation ratio of 0.2, the average score increases from around 70 to over 80 after optimization, and a similar trend is observed on Photo. The improvements become more pronounced with higher perturbation ratios.
    Experts specifically noted that the optimized graphs exhibited \emph{clearer textual descriptions}, \emph{more semantically coherent neighborhoods}, and \emph{more reliable labels}. These findings show that LAGA significantly enhances the perceived quality of text, structure, and labels, making the improvements not only measurable in downstream tasks but also interpretable to human experts. More detailed interpretative experimental setup and results are in the Appendix~\ref{appendix:human_expert}-\ref{appendix:case_study}.

\subsection{Scalability (Q5)}
\vspace{-2pt}
    \begin{wraptable}{r}{0.5\textwidth}
    \centering
    \vspace{-17pt}
    \caption{Performance comparison on arXiv dataset.}
    \resizebox{0.48\textwidth}{!}{
    \scriptsize
    \begin{tabular}{c|c|c c c c}
    \specialrule{1.0pt}{1.0pt}{1.0pt}
    \multicolumn{2}{c|}{\textbf{Perturbation Ratio}} &
    \textbf{0.0} &
    \textbf{0.2} &
    \textbf{0.4} &
    \textbf{0.8} \\
    \midrule

    \multirow{2}{*}{\parbox{1.2cm}{\centering \textbf{Text\\Sparsity}}}
    & LLM-TG 
    & $71.85_{\scriptstyle \pm 0.64}$ 
    & $70.65_{\scriptstyle \pm 0.59}$ 
    & $68.24_{\scriptstyle \pm 0.72}$ 
    & $63.39_{\scriptstyle \pm 0.80}$ \\
    & LAGA (Ours)
    & \cellcolor[HTML]{DADADA}$\textbf{74.12}_{\scriptstyle \pm \textbf{0.62}}$ 
    & \cellcolor[HTML]{DADADA}$\textbf{72.11}_{\scriptstyle \pm \textbf{0.68}}$ 
    & \cellcolor[HTML]{DADADA}$\textbf{70.45}_{\scriptstyle \pm \textbf{0.79}}$ 
    & \cellcolor[HTML]{DADADA}$\textbf{67.77}_{\scriptstyle \pm \textbf{0.84}}$ \\
    \midrule

    \multirow{2}{*}{\parbox{1.2cm}{\centering \textbf{Text\\Noise}}}
    & GCN 
    & $70.85_{\scriptstyle \pm 0.46}$ 
    & $70.36_{\scriptstyle \pm 0.52}$ 
    & $68.24_{\scriptstyle \pm 0.51}$ 
    & $66.83_{\scriptstyle \pm 0.54}$ \\
    & LAGA (Ours)
    & \cellcolor[HTML]{DADADA}$\textbf{74.12}_{\scriptstyle \pm \textbf{0.62}}$ 
    & \cellcolor[HTML]{DADADA}$\textbf{73.08}_{\scriptstyle \pm \textbf{0.61}}$ 
    & \cellcolor[HTML]{DADADA}$\textbf{71.72}_{\scriptstyle \pm \textbf{0.62}}$ 
    & \cellcolor[HTML]{DADADA}$\textbf{70.38}_{\scriptstyle \pm \textbf{0.64}}$ \\
    \midrule

    \multirow{2}{*}{\parbox{1.2cm}{\centering \textbf{Text\\Imbalance}}}
    & LLM-TG 
    & $71.85_{\scriptstyle \pm 0.42}$ 
    & $70.12_{\scriptstyle \pm 0.46}$ 
    & $69.77_{\scriptstyle \pm 0.51}$ 
    & $67.65_{\scriptstyle \pm 0.55}$ \\
    & LAGA (Ours)
    & \cellcolor[HTML]{DADADA}$\textbf{74.12}_{\scriptstyle \pm \textbf{0.62}}$ 
    & \cellcolor[HTML]{DADADA}$\textbf{72.41}_{\scriptstyle \pm \textbf{0.49}}$ 
    & \cellcolor[HTML]{DADADA}$\textbf{70.83}_{\scriptstyle \pm \textbf{0.50}}$ 
    & \cellcolor[HTML]{DADADA}$\textbf{70.21}_{\scriptstyle \pm \textbf{0.53}}$ \\
    \midrule

    \multirow{2}{*}{\parbox{1.2cm}{\centering \textbf{Structure\\Sparsity}}}
    & DHGR 
    & $71.85_{\scriptstyle \pm 0.36}$ 
    & $69.27_{\scriptstyle \pm 0.44}$ 
    & $66.13_{\scriptstyle \pm 0.43}$ 
    & $63.06_{\scriptstyle \pm 0.49}$ \\
    & LAGA (Ours)
    & \cellcolor[HTML]{DADADA}$\textbf{74.12}_{\scriptstyle \pm \textbf{0.62}}$ 
    & \cellcolor[HTML]{DADADA}$\textbf{70.70}_{\scriptstyle \pm \textbf{0.77}}$  
    & \cellcolor[HTML]{DADADA}$\textbf{68.19}_{\scriptstyle \pm \textbf{0.81}}$ 
    & \cellcolor[HTML]{DADADA}$\textbf{65.03}_{\scriptstyle \pm \textbf{0.82}}$ \\
    \midrule

    \multirow{2}{*}{\parbox{1.2cm}{\centering \textbf{Structure\\Noise}}}
    & DHGR 
    & $71.85_{\scriptstyle \pm 0.36}$ 
    & $66.01_{\scriptstyle \pm 0.46}$ 
    & $61.45_{\scriptstyle \pm 0.52}$ 
    & $54.23_{\scriptstyle \pm 0.56}$ \\
    & LAGA (Ours)
    & \cellcolor[HTML]{DADADA}$\textbf{74.12}_{\scriptstyle \pm \textbf{0.62}}$ 
    & \cellcolor[HTML]{DADADA}$\textbf{67.31}_{\scriptstyle \pm \textbf{0.86}}$ 
    & \cellcolor[HTML]{DADADA}$\textbf{63.42}_{\scriptstyle \pm \textbf{0.89}}$ 
    & \cellcolor[HTML]{DADADA}$\textbf{58.84}_{\scriptstyle \pm \textbf{0.89}}$ \\
    \midrule

    \multirow{2}{*}{\parbox{1.2cm}{\centering \textbf{Structure\\Imbalance}}}
    & GraphPatcher 
    & $71.85_{\scriptstyle \pm 0.11}$ 
    & $69.58_{\scriptstyle \pm 0.13}$ 
    & $68.12_{\scriptstyle \pm 0.12}$ 
    & $67.31_{\scriptstyle \pm 0.10}$ \\
    & LAGA (Ours)
    & \cellcolor[HTML]{DADADA}$\textbf{74.12}_{\scriptstyle \pm \textbf{0.62}}$ 
    & \cellcolor[HTML]{DADADA}$\textbf{71.67}_{\scriptstyle \pm \textbf{0.68}}$ 
    & \cellcolor[HTML]{DADADA}$\textbf{70.85}_{\scriptstyle \pm \textbf{0.66}}$ 
    & \cellcolor[HTML]{DADADA}$\textbf{68.21}_{\scriptstyle \pm \textbf{0.69}}$ \\
    \midrule

    \multirow{2}{*}{\parbox{1.2cm}{\centering \textbf{Label\\Sparsity}}}
    & GraphHop
    & $70.85_{\scriptstyle \pm 0.34}$ 
    & $69.76_{\scriptstyle \pm 0.41}$ 
    & $69.13_{\scriptstyle \pm 0.45}$ 
    & $68.87_{\scriptstyle \pm 0.44}$ \\
    & LAGA (Ours)
    & \cellcolor[HTML]{DADADA}$\textbf{74.12}_{\scriptstyle \pm \textbf{0.62}}$ 
    & \cellcolor[HTML]{DADADA}$\textbf{73.31}_{\scriptstyle \pm \textbf{0.63}}$ 
    & \cellcolor[HTML]{DADADA}$\textbf{72.75}_{\scriptstyle \pm \textbf{0.61}}$ 
    & \cellcolor[HTML]{DADADA}$\textbf{71.25}_{\scriptstyle \pm \textbf{0.63}}$ \\
    \midrule

    \multirow{2}{*}{\parbox{1.2cm}{\centering \textbf{Label\\Noise}}}
    & GCN 
    & $69.85_{\scriptstyle \pm 0.22}$ 
    & $68.75_{\scriptstyle \pm 0.24}$ 
    & $66.16_{\scriptstyle \pm 0.26}$ 
    & $60.47_{\scriptstyle \pm 0.45}$ \\
    & LAGA (Ours)
    & \cellcolor[HTML]{DADADA}$\textbf{74.12}_{\scriptstyle \pm \textbf{0.62}}$ 
    & \cellcolor[HTML]{DADADA}$\textbf{73.42}_{\scriptstyle \pm \textbf{0.67}}$ 
    & \cellcolor[HTML]{DADADA}$\textbf{69.12}_{\scriptstyle \pm \textbf{0.70}}$ 
    & \cellcolor[HTML]{DADADA}$\textbf{65.35}_{\scriptstyle \pm \textbf{0.72}}$ \\
    \midrule

    \multirow{2}{*}{\parbox{1.2cm}{\centering \textbf{Label\\Imbalance}}}
    & GraphSHA
    & $48.72_{\scriptstyle \pm 0.21}$ 
    & $49.20_{\scriptstyle \pm 0.37}$ 
    & $47.89_{\scriptstyle \pm 0.39}$ 
    & $47.56_{\scriptstyle \pm 0.40}$ \\
    & LAGA (Ours) 
    & \cellcolor[HTML]{DADADA}$\textbf{56.43}_{\scriptstyle \pm \textbf{0.56}}$ 
    & \cellcolor[HTML]{DADADA}$\textbf{55.01}_{\scriptstyle \pm \textbf{0.99}}$ 
    & \cellcolor[HTML]{DADADA}$\textbf{53.77}_{\scriptstyle \pm \textbf{1.21}}$ 
    & \cellcolor[HTML]{DADADA}$\textbf{52.10}_{\scriptstyle \pm \textbf{1.06}}$ \\

    \specialrule{1.0pt}{1.3pt}{1.0pt}
    \end{tabular}}
    \vspace{-10pt}
    \label{table:arxiv_perf}
    \end{wraptable}
    
    To address \textbf{Q5}, we investigate the scalability of our method through both theoretical complexity analysis and empirical evaluation.
    The \textbf{time complexity} of our framework is summarized as follows: For the \textit{Detection Agent}, the cost is $O(n\bar{\ell}+m+nfk)$, where $\bar{\ell}$ is the average text length, $f$ is the feature dimension, and $k$ is the number of clusters in $k$-means. For the \textit{Planning Agent}, the cost is $O(T_{\text{LLM}})$, representing the time for a single LLM inference. The \textit{Action Agent} has two parts: the graph learning stage has per-epoch cost $O(Lmh+mh+nh^2)$, where $h$ is the hidden dimension and $L$ is the number of GNN layers; the graph optimization stage has complexity $O(n_s(d+\log n+k_{\text{edge}})+m+n_{opt}T_{\text{LLM}})+O(r_{\text{gen}}n_{avg}(d+\log n+k_{\text{edge}}+T_{\text{LLM}}))$, where $n_s$ is the number of sparse nodes, $k_{\text{edge}}$ is the number of edges added per sparse node, $n_{opt}$ is the number of nodes with text issues, and $n_{avg}$ is the average number of nodes per class. For the \textit{Evaluation Agent}, the complexity is $O(n\bar{\ell}+m+nfk+T_{\text{LLM}})$.

    To ensure scalability, LAGA is designed as an offline data optimization framework, where the computational cost of the agentic loop is incurred only once. Crucially, the Planning and Evaluation agents operate on graph-level statistics rather than raw data, ensuring that LLM inference costs remain constant $O(1)$ regardless of graph size. As shown in Table~\ref{table:arxiv_perf}, LAGA achieves significant gains on the large-scale ogbn-arxiv dataset (169k nodes). This confirms that our framework effectively scales to large graphs, with specific acceleration techniques detailed in Appendix~\ref{appendix:scalability}.

\vspace{-5pt}
\section{Conclusion}
\vspace{-3pt}
    We introduced LAGA, an automated multi-agent framework powered by large language models, to address data quality in text-attributed graphs. Unlike prior methods, LAGA tackles text, structure, and label issues holistically, covering sparsity, noise, and imbalance. The framework integrates four agents—detection, planning, action, and evaluation—forming a closed optimization loop. LAGA unifies graph optimization and learning, enabling flexible interventions. Experiments show that LAGA consistently outperforms baselines and scales effectively.
    Future work will focus on extending LAGA to heterophilous graphs with different structural and semantic characteristics, which may require new optimization strategies.

\bibliography{reference}
\bibliographystyle{plain}

\newpage
\appendix
\onecolumn

\section{Emperical Study}
\label{appendix:emperical_study}

    \begin{figure*}[htbp]
        \centering
        \includegraphics[width=0.95\textwidth]{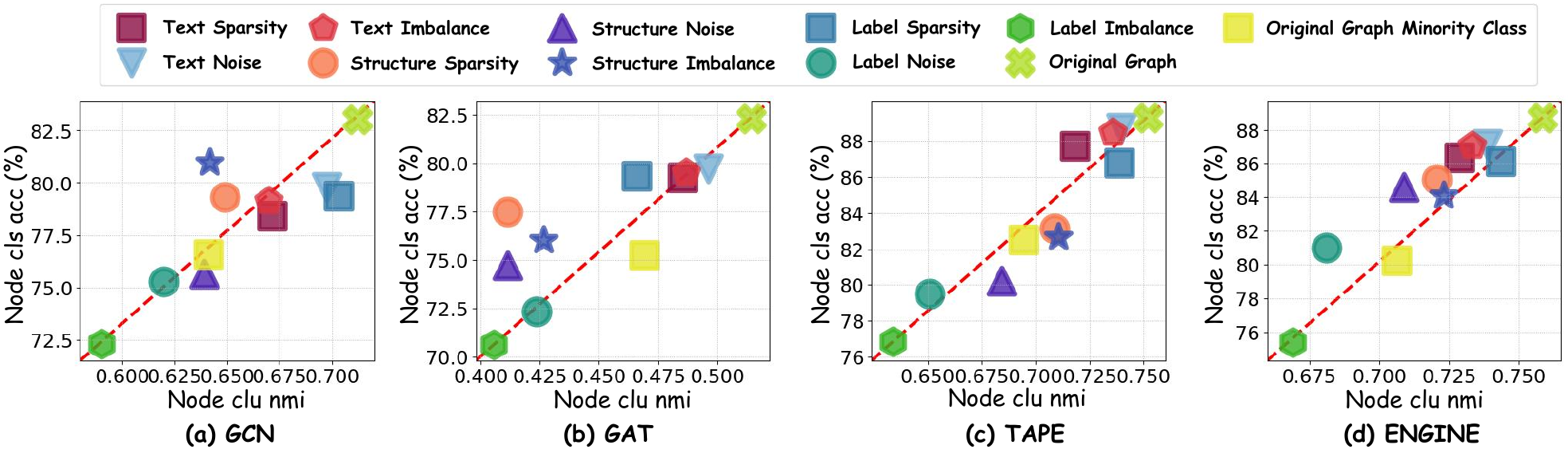}
        \vspace{-3pt}
        \caption{
            Performance comparison of two GNN backbones (GCN, GAT) and two LLM-GNN backbones (TAPE, ENGINE) across nine TAG quality scenarios on Cora. 
            Each point represents a specific quality degradation case, evaluated on two downstream tasks: node classification accuracy (Node cls acc) and node clustering normalized mutual information (Node clu nmi). 
            For the label imbalance scenario, we report the accuracy and NMI of the minority class under both the original and perturbed graphs.
        }
        \label{figure:emperical_study}
    \end{figure*}

    In this experiment, we compare the performance of two traditional graph neural network (GNN) models (GCN and GAT) with two LLM-augmented GNN variants (TAPE and ENGINE) across nine different TAG quality degradation scenarios. The experiments were conducted using the Cora dataset, with defects categorized into nine different types, including text sparsity, text imbalance, structure noise, structure sparsity, structure imbalance, label sparsity, label imbalance, label noise, and the original graph as a baseline.

    \subsection{Experimental Setup}
    
    Each data point represents the performance on a specific quality degradation scenario, evaluated on two downstream tasks:
    - \textbf{Node Classification Accuracy} (Node cls acc)
    - \textbf{Node Clustering Normalized Mutual Information} (Node clu nmi)
    
    Node classification measures the prediction accuracy of the model under quality degradation, while node clustering evaluates the performance in graph clustering tasks. For the label imbalance scenario, we report both accuracy and normalized mutual information (NMI) of the minority class under both the original and perturbed graphs.
    
    \subsection{Results Analysis}
    
    Figure~\ref{figure:emperical_study} illustrates the performance of four models across nine quality degradation scenarios. Each point corresponds to a specific quality degradation scenario, where the x-axis represents node clustering normalized mutual information (Node clu nmi), and the y-axis represents node classification accuracy (Node cls acc).
    
    \textbf{Impact of Text-related Quality Issues}.
    Text-related quality problems, such as text sparsity, text imbalance, and text noise, have a significant impact on the performance of both traditional and LLM-augmented models.
    
    - \textbf{Text Sparsity:} As the amount of textual information decreases, models have less semantic content to leverage, which negatively affects their ability to classify nodes accurately. Traditional GNN models like GCN and GAT exhibit a noticeable drop in node classification accuracy when faced with sparse text. LLM-augmented models (TAPE and ENGINE) perform slightly better in this scenario, as the LLMs can help fill in missing semantic information, but performance still degrades compared to the original graph.
    
    - \textbf{Text Imbalance:} In scenarios where there is an imbalance in the distribution of text across nodes (e.g., some nodes have extensive descriptions while others have little to none), both traditional and LLM-augmented models struggle to classify nodes accurately. This imbalance causes the models to focus more on well-described nodes, potentially ignoring the underrepresented ones. LLM-augmented models show some resilience, but the imbalance still leads to a performance drop, especially for nodes with less informative text.
    
    - \textbf{Text Noise:} Noisy text, such as irrelevant or inconsistent descriptions, degrades the ability of the models to extract meaningful features from node text. Both traditional and LLM-augmented models suffer from this issue, with a marked reduction in classification and clustering accuracy. LLM-augmented models show some resilience, but the impact of noise on performance remains substantial.
    
    These findings underscore the importance of clean, informative, and balanced text for achieving optimal performance in tasks involving TAGs.
    
    \textbf{Impact of Structure-related Quality Issues}.
    Structural issues, including structure sparsity, structure imbalance, and structure noise, also significantly affect model performance.
    
    - \textbf{Structure Sparsity:} Missing edges or poorly connected nodes hinder the flow of information through the graph, which is essential for GNN-based learning. Both traditional and LLM-augmented models experience reduced performance in terms of node classification and clustering when the graph is sparse. However, LLM-augmented models show a slight improvement over GCN and GAT, likely due to their ability to leverage additional semantic information from the LLM.
    
    - \textbf{Structure Imbalance:} When the graph structure is imbalanced (e.g., some nodes are highly connected while others have very few connections), models tend to focus on the more densely connected nodes. This imbalance negatively affects the models' ability to propagate information to the under-connected nodes, reducing performance. Traditional GNN models like GCN and GAT show a noticeable drop in performance under structure imbalance, while LLM-augmented models exhibit slight improvements due to the added semantic context.
    
    - \textbf{Structure Noise:} Incorrect or irrelevant edges (structure noise) confuse the message-passing process of GNNs, leading to lower accuracy in both classification and clustering tasks. LLM-augmented models exhibit a slight advantage in these cases, as the extra semantic information helps the model better differentiate between useful and noisy connections. Still, the overall performance drop remains noticeable compared to the original graph.
    
    These results highlight the critical role of high-quality graph structure in ensuring effective learning, and the ability of LLM-augmented models to somewhat compensate for structural imperfections.
    
    \textbf{Impact of Label-related Quality Issues}.
    Label-related quality problems, such as label sparsity, label imbalance, and label noise, significantly degrade model performance, especially in supervised learning tasks.
    
    - \textbf{Label Sparsity:} When there are too few labeled nodes in the graph, the models struggle to learn accurate representations, as they have limited supervision. This issue is particularly detrimental to traditional GNN models (GCN and GAT), which rely heavily on labeled data for training. LLM-augmented models (TAPE and ENGINE) show some improvement, as the LLMs can generate additional semantic insights, but they still suffer from reduced accuracy in node classification.
    
    - \textbf{Label Imbalance:} In scenarios where certain labels dominate the dataset, models tend to bias their predictions toward the majority class, leading to poor performance on the minority class. Both traditional and LLM-augmented models experience a drop in classification performance, with LLM-augmented models showing slightly better performance in maintaining balance. However, the performance gap between the majority and minority classes still remains.
    
    - \textbf{Label Noise:} Noisy or incorrect labels further hinder the ability of the models to make accurate predictions. This problem affects all models, but LLM-augmented models show some resilience due to their capacity to integrate semantic information from the node text, helping to correct or adjust predictions based on contextual understanding.
    
    The results emphasize the need for clean and balanced labels to ensure high-quality learning, as label defects significantly influence both node classification and graph-level analytics.
    
    \textbf{Overall Performance Trends}.
    Overall, the analysis demonstrates that data quality issues—whether in the form of text sparsity, structure noise, structure imbalance, or label imperfections—negatively impact the performance of both traditional and LLM-augmented models. While LLM-augmented models (TAPE and ENGINE) show improved resilience in certain scenarios, particularly in handling noisy or sparse text and labels, they are not immune to the effects of poor data quality. These findings highlight the necessity of addressing quality issues in TAGs to ensure reliable and robust performance in downstream tasks.

\section{Definitions of Quality Issues}
\label{appendix:scene_defination}

    In this section, we provide detailed definitions of the nine quality issue scenarios introduced in the Sec.~\ref{section:introduction}.
    
    \textbf{Text Sparsity (TS).} Some nodes lack sufficient textual information due to missing or extremely short descriptions. Formally, for a node $v_i$, sparsity occurs when $\mathrm{t}_i = \varnothing$ or $|\mathrm{t}_i| \ll \bar{l}$, where $\bar{l}$ denotes the average text length.  
    
    \textbf{Text Noise (TN).} Node texts may contain spelling errors, grammatical mistakes, or irrelevant tokens, reducing semantic clarity. Let $\mathrm{t}_i = \{w_1,\dots,w_k\}$ be the token set; noise arises when a high fraction of tokens are corrupted, i.e., $\frac{|w_{\text{noise}}|}{|\mathrm{t}_i|} \gg 0$.  
    
    \textbf{Text Imbalance (TI).} Textual informativeness varies widely across nodes: some nodes have detailed content, others only minimal descriptions. This can be expressed as variance in length or information content, $\mathrm{Var}(|\mathrm{t}_i|) \gg 0$ or $\mathrm{Var}(I(\mathrm{t}_i)) \gg 0$, leading to inconsistent representation quality. Where $I(\mathrm{t}_i)$ denotes the information content (e.g., entropy or semantic richness) of text $\mathrm{t}_i$. 
    
    \textbf{Structure Sparsity (SS).} 
    The graph has too few edges, limiting message passing. Let $\bar{d} = \frac{2|\mathcal{E}|}{|\mathcal{V}|}$ denote the average degree; sparsity occurs when $\bar{d} \ll d_{\text{ref}}$ for some reference degree.  
    
    \textbf{Structure Noise (SN).} 
    Some edges incorrectly connect unrelated nodes, introducing spurious neighborhoods. For an edge $(v_i,v_j)\in \mathcal{E}$, noise arises when $\mathrm{sim}(t_i,t_j)$ or $\mathrm{sim}(y_i,y_j)$ is low despite an existing edge.  
    
    \textbf{Structure Imbalance (SI).} 
    Node degree distribution is highly skewed. This can be measured by the variance $\mathrm{Var}(d_i)$ or imbalance ratio $\frac{\max_i d_i}{\min_i d_i} \gg 1$, indicating hub-dominated topologies.   
    
    \textbf{Label Sparsity (LS).} 
    Only a small fraction of nodes are labeled, providing limited supervision. Formally, $|\mathcal{L}| \ll |\mathcal{V}|$, where $\mathcal{L}\subseteq \mathcal{V}$ is the set of labeled nodes.  
    
    \textbf{Label Noise (LN).} 
    Assigned labels may deviate from true semantics, leading to corrupted supervision. 
    Formally, for node $v_i$ with true label $y_i^{*}$, the observed label $y_i$ follows: $\Pr(y_i = c \mid y_i^{*} = c') = \eta_{c'c}$,
    where $c$ denotes a class and $\eta$ is a noise transition matrix. Such noise can distort learning and reduce model reliability.
    
    \textbf{Label Imbalance (LI).} 
    Label distribution is skewed, with some classes over-represented. This can be expressed as: $\max_{c} n_c \allowbreak \gg \min_{c} n_c$, where $n_c$ denotes the number of nodes in class $c$.

\section{Dynamic Planning vs. Static Rules}
\label{appendix:dynamic_plan}

    To demonstrate the necessity of the Planning Agent, we present a detailed case study on a specific node from the \textbf{Cora} dataset under a composite degradation scenario (Text Noise 40\% + Structure Noise 20\%). We compare our \textbf{LAGA (Dynamic Planning)} against a \textbf{Static Strategy} baseline, which employs a fixed optimization order (\textit{Text Denoise $\rightarrow$ Structure Refinement}) and equal loss weights.

\subsection{Scenario Description}
\begin{itemize}
    \item \textbf{Target Node ($v_{target}$)}: A paper belonging to the class ``Neural Network''.
    \item \textbf{Defect 1 (Textual)}: The abstract contains irrelevant noisy keywords (e.g.,``medical'', ``finance'') due to simulated data corruption.
    \item \textbf{Defect 2 (Structural)}: The node is wrongly connected to two ``Reinforcement Learning'' papers (heterophilous noise) and disconnected from its true ``Neural Networks'' neighbors.
\end{itemize}

\subsection{Comparison of Optimization Trajectories}
\textbf{Static Strategy} (The Failure Case):
    \begin{itemize}
        \item \textbf{Step 1 (Text Denoise)}: The static rule blindly applies text denoising first. Since the node is connected to ``Reinforcement Learning'' neighbors (structural noise), the text denoiser—which often relies on neighborhood aggregation—erroneously ``corrects'' the text to match the wrong neighbors, reinforcing the ``Reinforcement Learning'' semantics.
        \item \textbf{Step 2 (Structure Refinement)}: Based on the now-corrupted text (which semantically resembles RL), the structure refiner preserves the erroneous edges, assuming they are homophilous.
        \item \textbf{Loss Weights}: The static strategy uses fixed weights ($\lambda_{text}=1, \lambda_{struct}=1, \lambda_{label}=1$). This fails to account for the fact that the text modality is currently unreliable.
        \item \textbf{Outcome:} The node is misclassified as ``Reinforcement Learning''.
    \end{itemize}

\textbf{LAGA Planning Agent} (The Success Case)
    The Planning Agent follows the prompt instructions (Step 1--4) to generate a tailored plan.
    \begin{description}
        \item[Step 1: Input Analysis (Simulated JSON):] \
        \begin{verbatim}
        {
          "text_noise": {"severity": "severe", "priority": 1},
          "structure_noise": {"severity": "moderate", "priority": 2},
          "label_noise": {"severity": "negligible", "priority": 3}
        }
        \end{verbatim}
        \item[Step 2: Plan Loss Weights (Reasoning):] \
        \textit{``According to the rules: If a modality has severe issues, its loss should be down-weighted. Text has `severe' noise, so text\_loss should be `low'. Structure has `moderate' noise, so structure\_loss is `medium'. Label is reliable, so label\_loss is `high'.''}
        \begin{itemize}
            \item \textbf{Decision:} $\lambda_{text} \rightarrow$ \textbf{Low} (0.1), $\lambda_{struct} \rightarrow$ \textbf{Medium} (0.5), $\lambda_{label} \rightarrow$ \textbf{High} (1.0).
            \item \textbf{Impact:} This prevents the model from overfitting to the noisy text and structure during representation learning.
        \end{itemize}
        \item[Step 3: Plan Action Module Execution (Reasoning):] \
        \textit{``Standard guideline suggests fixing text first. However, text noise is `severe' and priority is 1. If we fix text using current noisy structure, error propagation occurs. Given the high severity of text noise, we should prioritize Structure Optimization to prune wrong edges first, creating a cleaner graph for subsequent text denoising.''}
        \begin{itemize}
            \item \textbf{Selected Sequence:} \textbf{Structure Optimization} $\rightarrow$ \textbf{Text Optimization}.
            \item \textbf{Justification:} Pruning heterophilous edges first prevents the text optimizer from aggregating noise from wrong neighbors.
        \end{itemize}
        \item[Step 4: Final Output:] \
        By dynamically reordering the action sequence (Structure before Text) and down-weighting the noisy modalities, LAGA correctly isolates the node and restores its true semantics, leading to correct classification.
    \end{description}
    
\subsection{Quantitative Impact of Dynamic Weights}
    We further validate the impact of \textbf{Reliability-Aware Loss Weighting}. Compared to fixed weights, our dynamic weighting strategy achieves:
    \begin{enumerate}
        \item \textbf{Faster Convergence:} The model stops oscillating between conflicting modalities (text vs. structure).
        \item \textbf{Higher Robustness:} In high-noise scenarios (Noise $> 60\%$), dynamic weighting prevents the ``collapse'' of node representations, maintaining a performance gain of approximately \textbf{4.5\%} over the static baseline.
    \end{enumerate}

\clearpage 

\section{Algorithm}
\subsection{Action Agent - Graph Learning}
\label{appendix:alg_gl}
    The pseudocode of the graph learning part in the Action Agent is shown in Algorithm~\ref{algorithm:graph_learning}.

    \vspace{10pt}
    \label{algorithm:graph_learning}
    \begin{algorithm}[H]
    \caption{Action Agent — Graph Learning}
    \textbf{Input:} TAG $\mathcal{G}=(\mathcal{V},\mathcal{E},\mathcal{T})$, planning report $\mathcal{R}_{plan}=(\mathcal{R}_{ana},(\alpha,\beta,\gamma),\mathcal{P}^\star)$, (optional) fine-tuned LLM $\mathcal{F}_{\text{LLM}}$. \\
    \textbf{Output:} Semantic embeddings $h^{sem}$, structural embeddings $h^{stu}$, edge probabilities $\hat{\mathbf{A}}$. \\
    \For{$v_i \in \mathcal{V}$}{
      $(\texttt{sum},\texttt{kword},y_i^{pse}) \leftarrow \mathrm{LLM}(t_i)$; \\
      $x_i^{text} \leftarrow t_i \oplus \{\texttt{sum},\texttt{kword}\}$; \\
      $h_i^{init} \leftarrow \mathrm{Enc}(x_i^{text})$;
    }
    $h^{sem} \leftarrow \mathrm{MLP}_{sem}(h^{init})$; \\
    \textit{/* topology-, label- and semantic-aligned semantic objective */}
    \begin{flalign*}
    \mathcal{L}^{sem}_{total} \leftarrow \alpha\,\|h^{sem}-y^{pse}\|^2
    + \beta\,\mathcal{L}_{\text{struct}}(h^{sem})
    + \gamma\,\mathrm{CE}(h^{sem},y), \\
    \mathcal{L}_{struct}(z_i) 
    = - \log \frac{\sum_{j \in \mathcal{N}(i)} e^{\mathrm{sim}(z_i,z_j)/\tau}}
    {\sum\limits_{j \in \mathcal{N}(i)} \!\!\!e^{\mathrm{sim}(z_i,z_j)/\tau} + \sum\limits_{n \in \mathcal{N}^-(i)} \!\!\!e^{\mathrm{sim}(z_i,z_n)/\tau}}; &&
    \end{flalign*}
    \textbf{Update} $h^{sem}$ by minimizing $\mathcal{L}^{sem}_{total}$; \\
    
    $h^{stu} \leftarrow \mathrm{GCN}(h^{init},\mathbf{A})$; \\
    $\hat{a}_{ij} \leftarrow \sigma(\mathrm{MLP}_{link}([h^{stu}_i \,\|\, h^{stu}_j]))$ and assemble $\hat{\mathbf{A}}$; \\
    \textit{/* topology-, label- and semantic-aligned structural objective */}
    \begin{flalign*}
    \mathcal{L}^{stu}_{total} \leftarrow \alpha\!\left(\|h^{stu}-h^{sem}\|^2 + \!\!\sum_{(i,j)}\!(\sigma(\mathrm{sim}(h^{sem}_i,h^{sem}_j))-\hat{a}_{ij})^2\right)
    + \beta\,\|\mathbf{A}-\hat{\mathbf{A}}\|^2
    + \gamma\,\mathrm{CE}(h^{stu},y); &&
    \end{flalign*}
    \textbf{Update} $(h^{stu},\hat{\mathbf{A}})$ by minimizing $\mathcal{L}^{stu}_{total}$; \\
    
    \If{pretraining phase (only once, as an offline preprocessing step)}{
      Construct fine-tuning dataset $\mathcal{D}_{ft}$: \\
      \Indp $\bullet$ Sample clean texts $t^{ref}$ from external corpus (e.g., arXiv). \\
      $\bullet$ Generate perturbed variants $t^{noisy}$ (for denoising) and truncated/low-information variants $t^{incomp}$ (for completion). \\
      $\bullet$ Pair $(t^{noisy},t^{ref})$ for denoising task, $(t^{incomp},\{t^{ref},\text{neighbor texts}\})$ for completion task. \\
      \Indm
      Fine-tune LLM $\mathcal{F}_{\text{LLM}}$ with LoRA adapters using objective: \\
      \Indp $\mathcal{L}_{LLM}=\mathrm{CE}(t^{out},t^{ref})+\lambda H(t^{out})$. \\
      \Indm
      \textbf{Save} adapted $\mathcal{F}_{\text{LLM}}$ for downstream use in Action Agent; no further tuning is required during iterative optimization. 
    }
    
    \textbf{Return} $(h^{sem},h^{stu},\hat{\mathbf{A}})$.
    \end{algorithm}
    \vspace{10pt}

    Where $\texttt{sum}$, $\texttt{kword}$ and $y^{pse}$ represent the summary, keywords and pseudo-label generated by the LLM; $\mathrm{Enc}(\cdot)$ denotes an LM encoder; $\mathrm{CE}(\cdot)$ is the cross entropy and $\mathrm{sim}(z_i,z_j)$ represents the cosine similarity between embeddings. $\mathcal{N}(i)$ denotes the set of neighboring nodes of $v_i$, $\mathcal{N}^-(i)$ is a sampled set of non-neighboring nodes, and $\tau>0$ is a temperature parameter in contrastive learning. $h^{init}_i$ is the initial embedding of node $v_i$, $h^{sem}_i$ is the semantic embedding learned via MLP, and $h^{stu}_i$ is the structural embedding learned via GCN. $\hat{a}_{ij}$ is the predicted edge probability between nodes $v_i$ and $v_j$, and $\hat{\mathbf{A}}$ is the full edge prediction matrix. $\alpha,\beta,\gamma$ are severity-aware loss weights assigned by the Planning Agent to balance semantic, structural, and label objectives. $\mathcal{L}_{LLM}$ denotes the fine-tuning loss for the LLM with LoRA adapters, where the entropy term $H(\cdot)$ encourages informative and diverse text generation. Together, these components enable multi-source and severity-aware representation learning as the foundation for subsequent graph optimization.

\subsection{Action Agent - Graph Optimization}
\label{appendix:alg_go}
    The pseudocode of the graph optimization part in the Action Agent is shown in Algorithm~\ref{algorithm:graph_optimization}.

    \label{algorithm:graph_optimization}
    \begin{algorithm}[H]
    \caption{Action Agent — Graph Optimization}
    \textbf{Input:} $\mathcal{G}=(\mathcal{V},\mathcal{E},\mathcal{T})$, localization $\mathcal{M}_{local}$, plan $\mathcal{P}^\star$, action library $\mathcal{A}=\mathcal{A}^{text}\!\cup\!\mathcal{A}^{struct}\!\cup\!\mathcal{A}^{label}\!\cup\!\mathcal{A}^{class}$, learned $(h^{sem},h^{stu},\hat{\mathbf{A}})$. \\
    \textbf{Output:} Optimized TAG $\mathcal{G}'=(\mathcal{V}',\mathcal{E}',\mathcal{T}',\mathbf{Y}')$. \\
    \textbf{Select} actions $\mathbf{a} \in \mathcal{A}$ based on the strategy $\mathcal{P}^\star$. \\
    
    \textbf{Text Optimization} ($\mathcal{A}^{text}$): \\
    \For{$t_i \in \mathcal{M}_{local}^{text}$}{
      \uIf{task = $\texttt{denoise}$}{ $t_i^{opt}\!\leftarrow\!\mathcal{F}_{\text{LLM}}(t_i,Con(i)\,|\,\mathcal{P}_{denoise})$; }
      \Else{ $t_i^{opt}\!\leftarrow\!\mathcal{F}_{\text{LLM}}(t_i,Con(i)\,|\,\mathcal{P}_{completion})$; }
      $\mathcal{T}\!:\ t_i \leftarrow t_i^{opt}$;
    }
    
    \textbf{Structure Optimization} ($\mathcal{A}^{struct}$): \\
    \For{$C_k \in \mathcal{M}_{local}^{struct}$}{
      \For{$(i,j)\in E(C_k)$}{
        \If{$\hat{a}_{ij}<\tau_{edge}$}{\textbf{Delete} edge $(i,j)$;}
      }
      \For{$i \in C_k$ with $\deg(i)<k_{edge}$}{
        $S \leftarrow \{u\!\in\! C_k\!\setminus\!\mathcal{N}(i): \hat{a}_{iu}>\tau_{edge}\}$; \\
        \textbf{Add} edges $(i,u)$ to top-$(k_{edge}-\deg(i))$ nodes in $S$ by $\hat{a}_{iu}$;
      }
    }
    
    \textbf{Label Optimization} ($\mathcal{A}^{label}$): \\
    \For{$v_i \in \mathcal{M}_{local}^{label}$}{
      $p(\cdot|v_i)\!\leftarrow\!\mathrm{softmax}(h^{stu}_i)$; \\
      $c_i \!\leftarrow\! \exp\!\big(\lambda \log p_{m}(h_i) + (1{-}\lambda)\log(p_{m}(h_i){-}p_{sm}(h_i))\big)$; \\
      \uIf{$c_i>\tau_{lape}$}{ $\hat{y}_i \leftarrow \arg\max p(\cdot|v_i)$; }
      \Else{ $\hat{y}_i \leftarrow \arg\max\limits_{c \in \mathcal{Y}} \sum\limits_{j\in \mathcal{N}(i)} \hat{a}_{ij}\,\mathbb{I}[y_j=c]$; }
      $\mathbf{Y}\!:\ y_i \leftarrow \hat{y}_i$;
    }
    
    \textbf{Node Generation} ($\mathcal{A}^{class}$): \\
    \For{$c \in \mathcal{C}$ with $|V_c| < \tau_{gen}$}{
      $n_c \leftarrow \tau_{gen}-|V_c|$; \\
      \For{$r=1$ \KwTo $n_c$}{
        $t^{new} \leftarrow \mathcal{F}_{\text{LLM}}(\varnothing,Con_c\,|\,\mathcal{P}_{completion})$; \\
        $h^{init}_{new} \leftarrow \mathrm{Enc}(t^{new})$;\quad $h^{stu}_{new} \leftarrow \mathrm{GCN}(h^{init}_{new},\mathbf{A})$; \quad
        $\hat{a}_{new} \leftarrow \mathrm{MLP}_{link}(h^{stu}_{new},\mathbf{A})$; \\
        \textbf{Connect} $v^{new}$ to top-$k_{edge}$ neighbors by $\hat{a}_{new,j}$; \\
        \textbf{Insert} $v^{new}$ with $(t^{new},y{=}c)$ into $(\mathcal{V},\mathcal{E},\mathcal{T},\mathbf{Y})$;
      }
    }
    \textbf{Return} $\mathcal{G}'=(\mathcal{V}',\mathcal{E}',\mathcal{T}',\mathbf{Y}')$.
    \end{algorithm}
    \vspace{10pt}

    where $\mathcal{M}_{local}$ denotes the problem localization set returned by the Detection Agent, $\mathcal{P}^\star$ is the optimization strategy plan generated by the Planning Agent, and $\mathcal{A}$ is the action library containing four categories of optimization operators. $t_i$ represents the original text of node $v_i$, and $t_i^{opt}$ is the optimized text produced by the fine-tuned LLM $\mathcal{F}_{\text{LLM}}$ under prompt $\mathcal{P}_{task}$. $Con(i)$ and $Con_c$ denote context texts sampled from neighbors or class-specific nodes, respectively. $\hat{a}_{ij}$ is the predicted probability of an edge between nodes $v_i$ and $v_j$, with threshold $\tau_{edge}$ controlling edge deletion and addition, and $k_{edge}$ specifying the upper bound of added edges per node. $\mathbf{A}$ and $\hat{\mathbf{A}}$ denote the original and predicted adjacency matrices. For label optimization, $h^{stu}_i$ is the structure embedding of node $v_i$, $p(h_i)$ is the softmax class probability distribution, $p_{m}$ and $p_{sm}$ denote the highest and second-highest predicted probabilities, and $c_i$ is the confidence score with threshold $\tau_{lape}$ deciding whether to trust the prediction. $\mathcal{C}$ is the class set with $d=|\mathcal{C}|$ classes, and $V_c$ denotes the set of nodes in class $c$. $\tau_{gen}$ is the generation threshold, $r_{gen}$ is the balancing ratio, and $n_c$ denotes the number of new nodes to be generated for minority class $c$. Finally, $\mathcal{G}'=(\mathcal{V}',\mathcal{E}',\mathcal{T}',\mathbf{Y}')$ is the optimized TAG returned after executing all selected actions.

\section{Hyperparameter Settings}
\label{appendix:hyperpara_setting}

    In LAGA, most hyperparameters are fixed across all datasets and scenarios, except for four tunable hyperparameters in the Action Agent. 
    Specifically, we search $k \in \{3,5,10,15,20\}$ for the number of selected keywords, $\tau_{edge} \in [0.4,0.6]$ for the edge similarity threshold, $\tau_{lape} \in [0.6,0.8]$ for the label confidence threshold, and $r_{gen} \in [0.1,0.5]$ for the ratio of generated minority-class nodes. 
    These four hyperparameters are tuned by grid search on the validation set.  
    All other hyperparameters are kept fixed to ensure consistency and reproducibility. For the baselines, we follow the optimal settings reported in their original papers.
    
    For the \textbf{Detection Tools}, we introduce thresholds that determine when a node or community is flagged as suffering from a particular quality issue. 
    For text, thresholds control the minimum length regarded as sufficient ($\tau_s^{text}$), the maximum tolerable noise rate ($\tau_n^{text}$), and the minimum informativeness level ($\tau_{imb}^{text}$). 
    For structure, we introduce five thresholds that determine when a community is flagged as problematic. 
    Specifically, we use $\tau_{d}^{struct}$ as the minimum average degree and $\tau_{\rho}^{struct}$ as the minimum edge density. 
    For noise, we use $\tau_{se}^{struct}$ as the maximum acceptable structural entropy and $\tau_{j}^{struct}$ as the minimum Jaccard similarity. 
    For imbalance, we use $\tau_{imb}^{struct}$ to bound the skewness of the degree distribution. 
    For labels, thresholds specify the minimum confidence required for reliable prediction ($\tau_c$), as well as $\tau_{imb}^{label}$ that controls the degree of skewness in the label distribution considered imbalanced. 
    In practice, we set $\tau_s^{text}=20$, $\tau_n^{text}=0.015$, $\tau_{imb}^{text}=0.05$, $\tau_{d}^{struct}=2.0$, $\tau_{\rho}^{struct}=0.05$, $\tau_{se}^{struct}=2.5$, $\tau_{j}^{struct}=0.1$, $\tau_{imb}^{struct}=0.5$, $\tau_c=0.5$ and $\tau_{imb}^{label}=0.5$.  
    
    For the \textbf{Evaluation Agent}, the only hyperparameters are the stopping thresholds. 
    Specifically, $\tau_{imp-f}$ is used in the first iteration to decide whether the optimization terminates immediately, 
    while $\tau_{imp}$ is applied in subsequent iterations to determine whether the improvement over the previous round is sufficient to continue. 
    In practice, we set $\tau_{imp-f}=8.0$ and $\tau_{imp}=1.0$, meaning that the optimization loop stops if the initial quality score is already satisfactory or if further improvements fall below $10\%$. These fixed hyperparameter values are chosen based on prior knowledge and validated by our experimental results to approximate optimal configurations.

    Regarding the learning modules, we adopt a 2-layer MLP encoder with hidden demension 64 in semantic learning, furthermore we use a 2-layer GCN and a 2-layer MLP with hidden dimension 64 in structure learning. 
    The model is trained with Adam optimizer using a learning rate of $0.005$, weight decay of $5\mathrm{e}{-4}$, and dropout rate of $0.5$. 
    The maximum number of training epochs is set to 200 with early stopping patience of 20 epochs. 
    For the text encoder, we use the pre-trained Sentence-BERT with frozen parameters unless otherwise specified. 

\begin{table*}[htbp]
    \centering
    \caption{Details of text-attributed graphs used in our experiments.}
    \resizebox{\textwidth}{!}{
    \begin{tabular}{l|ccccccc}
    \specialrule{1pt}{1.5pt}{1.5pt}
    Dataset & \# Nodes & \# Edges & \# Classes & \# Train/Val/Test & \# Homophily & Text Information & Domains \\
    \midrule
    Cora      & 2,708  & 10,556  & 7 & 60/20/20 & 0.81 & Title and Abstract of Paper & Citation\\
    Citeseer  & 3,186  & 8,450  & 6 & 60/20/20 & 0.74 & Title and Abstract of Paper & Citation\\
    arXiv   & 169343 & 2315598 & 40 & 60/20/20 & 0.66 & Title and Abstract of Paper & Citation\\
    \midrule
    WikiCS    & 11,701 & 431,726 & 10 & 60/20/20 & 0.65 & Title and Abstract of Article & Knowledge\\
    \midrule
    Photo     & 48,362 & 873,793 & 12 & 60/20/20 & 0.74 & User Review of Product & 
    E-commerce \\
    \bottomrule
    \end{tabular}}

    \label{table:datasets}
    \end{table*}

\section{Datasets}
\label{appendix:dataset}

    We evaluate LAGA and baselines on five widely adopted TAG datasets across multiple domains, including three citation networks~\cite{tsgfm,tape}, one knowledge network~\cite{tsgfm, wikics} and one e-commerce networks~\cite{rethinkingtag}. 
    The details of these TAG datasets are shown in Table~\ref{table:datasets}. The description of the datasets for each domain is as follows:

    \begin{itemize}
        \item \textbf{Citation Networks.}~\cite{tsgfm}. 
        Cora, Citeseer and arXiv are
        benchmark datasets of citation networks. Nodes represent paper, and edges represent citation relationships. The features are obtained through pre-trained language model, and labels denote their academic fields.
    
        \item \textbf{Knowledge Networks}~\cite{tsgfm,wikics}. 
        WikiCS is a benchmark dataset of knowledge networks. Nodes represent articles in the field of computer science, and edges represent hyperlinks between these articles. The features are obtained through pre-trained language model, and labels denote different branches of computer science.
    
        \item \textbf{E-commerce Networks} ~\cite{rethinkingtag}. 
        Photo is a benchmark dataset of e-commerce networks. Nodes represent products, and edges represent relationships such as co-purchase or co-view. The features are obtained through pre-trained language model, and labels correspond to product categories or user ratings.

    \end{itemize}

\section{Baselines}
\label{appendix:baseline}

    In this section, we present the backbone, baseline methods, and large language models (LLMs) used in our experiments. For comparison, we consider a broad set of state-of-the-art baselines tailored to each degradation type. In the text quality dimension, we include LLM-TG (built by ourselves), UltraTAG-S~\cite{ultratag-s}, PoDA~\cite{poda}, and CTD-MLM~\cite{ctd-mlm}. In the structure quality dimension, we evaluate against SUBLIME~\cite{sublime}, SE-GSL~\cite{segsl}, DHGR~\cite{dhgr}, LLM4RGNN~\cite{llm4rgnn}, RawlsGCN~\cite{rawlsgcn}, and GraphPatcher~\cite{graphpatcher}. In the label quality dimension, we adopt GraFN~\cite{grafn}, GraphHop~\cite{graphhop}, PI-GNN~\cite{pi-gnn}, NRGNN~\cite{nrgnn}, LTE4G~\cite{lte4g}, and TOPOAUC~\cite{topoauc}. These methods represent the most competitive approaches in each scenario.  
    To ensure fairness and to comprehensively assess robustness, we apply LAGA on multiple graph backbones. Specifically, we employ three representative GNN backbones (GCN~\cite{gcn}, GAT~\cite{gat} and GraphSAGE~\cite{graphsage}), as well as two recent LLM-GNN backbones (TAPE~\cite{tape} and ENGINE~\cite{engine}).

    \begin{itemize}
        \item \textbf{GCN}~\cite{gcn}. GCN is one of the most widely used GNN architectures, introducing a first-order approximation of spectral graph convolutions. It aggregates information from a node’s local neighborhood to update its representation, effectively combining graph structure and node features for tasks such as node classification and link prediction.
        \item \textbf{GAT}~\cite{gat}. GAT introduces a self-attention mechanism that assigns different importance weights to neighboring nodes. This allows the model to adaptively focus on more relevant neighbors, improving the expressiveness and interpretability of the learned node representations.
        \item \textbf{GraphSAGE}~\cite{graphsage}. GraphSAGE is an inductive framework that generates node embeddings by sampling and aggregating information from local neighborhoods. It enables scalable training on large graphs and generalizes well to unseen nodes or graphs through learnable aggregation functions. 
        \item \textbf{TAPE}~\cite{tape}.
        TAPE enhances textual graph learning by integrating LLMs with graph structures. It first employs an LLM to augment node textual descriptions and then uses a lightweight LM-to-LM interpreter to extract semantic features as node representations. These features are further processed by a GNN for structural aggregation, enabling TAPE to effectively capture both semantic and topological information. 
        \item \textbf{ENGINE}~\cite{engine}.
        ENGINE is a parameter- and memory-efficient framework that fuses LLMs and GNNs through a lightweight tunable side structure called G-Ladder. It integrates structural information without backpropagating through frozen LLM layers, significantly reducing training cost. ENGINE further introduces caching and dynamic early-exit strategies to accelerate both training and inference while maintaining competitive performance.
        \item \textbf{LLM-TG}.
        LLM-TG is a baseline method we design to address text sparsity and text imbalance in text-attributed graphs.
        It leverages LLMs for prompt-based text generation.
        Specifically, we first detect sparse or low-information texts using automatic evaluation tools, then construct prompts by combining the original text with sampled neighbor texts.
        The prompts are fed into an LLM to generate more complete and informative text representations, which are used to enrich node attributes.
        As a self-designed baseline, LLM-TG serves to evaluate the effectiveness of LLM-based text enhancement under sparse and imbalanced textual conditions.
        \item \textbf{UltraTAG-S}~\cite{ultratag-s}.
        UltraTAG-S is a unified framework for text-attributed graph learning that integrates graph neural networks with LLMs. It introduces a dual-channel semantic encoder that captures both local structural dependencies and global textual semantics. To achieve efficient adaptation, UltraTAG-S employs parameter-efficient tuning strategies to align node representations from graph and language modalities. 
        \item \textbf{PoDA}~\cite{poda}.
        PoDA is a sequence-to-sequence pre-training framework that reconstructs clean text from corrupted input by jointly training both the encoder and decoder. It enhances text robustness through noise injection and denoising objectives.
        In our experiments, we apply this text denoising method to the detected noisy texts for evaluation under text noise scenarios.
        \item \textbf{CTD-MLM}~\cite{ctd-mlm}.
        CTD-MLM is a contextual denoising approach that leverages pre-trained masked language models to iteratively clean noisy text without additional training. It uses context-based masking and candidate prediction to restore corrupted tokens.
        In our experiments, we apply this text denoising method to the detected noisy texts for evaluation under text noise scenarios.
        \item \textbf{RawlsGCN}~\cite{rawlsgcn}.
        RawlsGCN addresses structural imbalance in graphs by promoting fairness in message aggregation across nodes of varying degrees. It introduces a Rawlsian optimization principle that dynamically adjusts neighbor weights to prevent over-representation of high-degree nodes, thereby balancing the structural influence in GNN learning. This approach effectively enhances representation quality for structurally underrepresented nodes.
        \item \textbf{GraphPatcher}~\cite{graphpatcher}.
        GraphPatcher mitigates degree bias in GNNs through a test-time augmentation framework. It iteratively generates virtual neighbors to patch low-degree nodes, progressively reconstructing their local structures while preserving performance on high-degree nodes. As a model-agnostic and plug-and-play method, GraphPatcher improves generalization and robustness against structural imbalance without retraining the base GNN.
        \item \textbf{SUBLIME}~\cite{sublime}. SUBLIME constructs GSL graphs using an anchor view (original graph) and a learner view. The learn view is initialized with kNN and optimized with parameterized or non-parameterized methods, followed by post-processing steps like top-k filtering and symmetrization. Contrastive learning between the two views refines the graph for downstream tasks.
        \item \textbf{SEGSL}~\cite{segsl}. SE-GSL fuses the original graph with a kNN graph and leverages structural entropy to refine edges and build a hierarchical encoding tree that captures community structures. It reconstructs the graph through entropy-guided sampling and jointly optimizes the graph and model parameters during training.
        \item \textbf{DHGR}~\cite{dhgr}. DHGR enhances GNN performance on heterophily graphs by preprocessing the graph structure through rewiring. It systematically adds homophilic edges and removes heterophilic ones based on label and feature similarities, improving node classification accuracy, particularly for heterophily settings.
        \item \textbf{LLM4RGNN}~\cite{llm4rgnn}. LLM4RGNN is a framework that enhances the adversarial robustness of GNNs using LLMs. It detects malicious edges and restores critical ones through a purification strategy, leveraging fine-tuned LLMs for edge prediction and correction. This approach effectively mitigates topology attacks, improving GNN performance across attack scenarios.
        \item \textbf{GraFN}~\cite{grafn}.
        GraFN addresses the challenge of label sparsity by integrating label propagation with feature transformation. It leverages the correlation between graph structure and node features through a label-aware feature normalization mechanism, enabling effective learning even with very few labeled samples.
        \item \textbf{GraphHop}~\cite{graphhop}.
        GraphHop is a scalable semi-supervised node classification method designed for scenarios with extremely limited labels. It alternates between label aggregation and label update through logistic regression, effectively propagating label information across multiple hops while maintaining low computational cost. 
        \item \textbf{PI-GNN}~\cite{pi-gnn}. PI-GNN is a noise-robust graph learning framework that leverages pairwise interactions (PI) between nodes as an auxiliary learning signal to mitigate the effect of noisy labels. It introduces a confidence-aware PI estimation module that adaptively estimates the likelihood of node pairs sharing the same class and a decoupled training strategy that jointly optimizes PI estimation and node classification. In our experiments, we apply this label noise handling strategy to our detected noisy labels for evaluating performance under label noise conditions.
        \item \textbf{NRGNN}~\cite{nrgnn}. NRGNN is designed for graphs with sparse and noisy labels. It mitigates label noise by linking unlabeled nodes with labeled ones that have high feature similarity, thereby propagating cleaner label information. Additionally, it employs a pseudo-label generation strategy and a GNN-based edge predictor to iteratively refine graph structure and supervision. In our experiments, we use this label noise-resistant approach on our detected noisy labels to conduct experiments under noisy label scenarios.
        \item \textbf{LTE4G}~\cite{lte4g}.
        LTE4G tackles the class imbalance problem by jointly reweighting label and topology information. It enhances minority-class representations through a label-topology coupling module that dynamically adjusts node embeddings according to both label distribution and neighborhood density. This design effectively alleviates representation bias caused by imbalanced class ratios.
        \item \textbf{TOPOAUC}~\cite{topoauc}.
        TOPOAUC is a unified framework designed to address both class and topology imbalance in graph learning. It directly optimizes the multi-class AUC objective to mitigate class imbalance while introducing a Topology-Aware Importance Learning (TAIL) mechanism to differentiate the structural contributions of neighboring nodes. This joint optimization improves discrimination for minority classes and structurally underrepresented nodes.
        \item \textbf{Sentence-BERT}~\cite{sentence-bert}.
        Sentence-BERT modifies BERT by introducing a Siamese network structure to generate fixed-size sentence embeddings.
        It enables efficient semantic similarity computation and clustering by comparing sentence embeddings via cosine similarity.
        \item \textbf{Gemma3-27B}~\cite{gemma3}.
        Gemma3-27B is a 27B-parameter open-weight large language model from Google’s Gemma series.
        It supports both text and image modalities with extended context windows and is optimized for efficient inference across diverse hardware settings.
        \item \textbf{LLaMA-33B}~\cite{llama}.
        LLaMA-33B is a 33B-parameter model in the LLaMA family based on a Transformer architecture.
        It is designed for strong reasoning and instruction-following capabilities while maintaining efficient scalability and inference performance.
        \item \textbf{Qwen3-32B}~\cite{qwen3}.
        Qwen3-32B is a 32.8B-parameter dense model from the Qwen3 series.
        It features a 32K token context window and excels at reasoning, dialogue, and instruction-following tasks across multiple domains.
    \end{itemize}

\section{Quality Scenario Construction Details}
\label{appendix:scene_set}
    We construct nine degradation scenarios (text/structure/label $\times$ sparsity/noise/imbalance). 
    Unless otherwise noted, the global perturbation ratio is $r\in\{0.2,0.4,0.8\}$. 
    For text and structure scenarios, the perturbation is applied to the graph content itself. 
    For label scenarios, we only perturb the \emph{training} labels and keep the validation/test labels intact.
    
    \textbf{Text Sparsity.}
    We model incomplete or overly short texts by \emph{deleting sentences only}. 
    We first sample a subset of nodes $S\subseteq V$ with $|S|=\lfloor r|V|\rfloor$, where $r$ controls the fraction of nodes to be perturbed. 
    For each selected node $v_i\in S$, we remove a random proportion of its sentences; the per-text modification ratio $\rho_i$ is independently drawn from $[0.5,1.0]$ and applied by uniformly deleting $\lfloor \rho_i \cdot |\text{sent}(t_i)| \rfloor$ sentences. 
    This produces shortened texts and may yield nearly empty texts when $\rho_i$ approaches $1.0$.

    \textbf{Text Noise.}
    We simulate low-quality texts by injecting lexical or syntactic noise, distracting content, or OOV tokens. 
    We first sample a subset of nodes $S\subseteq V$ with $|S|=\lfloor r|V\rfloor$, where $r$ controls the fraction of nodes to be perturbed. 
    For each selected node $v_i\in S$, we randomly choose one noise type: 
    (i) \emph{typos} by corrupting tokens with character-level edits, 
    (ii) \emph{grammar errors} by removing or swapping function words, 
    (iii) \emph{distracting information} by appending irrelevant tokens or sentences, or 
    (iv) \emph{OOV injection} by inserting rare or artificial tokens. 
    For each text, the modification ratio $\rho_i$ is independently sampled from $[0.5,1.0]$, determining the proportion of tokens or sentences corrupted.

    \textbf{Text Imbalance.}
    We create skewed distributions of text length and informativeness. 
    A subset of nodes $S\subseteq V$ with $|S|=\lfloor r|V\rfloor$ is sampled, where $r$ controls the proportion of affected nodes. 
    For each selected node $v_i\in S$, we apply one of two operations: 
    (i) \emph{truncation}, where only a fraction of sentences is preserved, or 
    (ii) \emph{content dilution}, where informative tokens are replaced with generic high-frequency words. 
    The modification ratio $\rho_i$ is drawn from $[0.5,1.0]$ for each text, ensuring heterogeneity in the degree of imbalance.

    \textbf{Structure Sparsity.}
    We reduce the global edge density. 
    Delete $\lfloor r|E|\rfloor$ edges sampled uniformly at random from $E$, without replacement. 
    This models globally sparse connectivity.
    
    \textbf{Structure Noise.}
    We add spurious/heterophilous edges. 
    Insert $\lfloor r|E|\rfloor$ new edges between non-adjacent node pairs preferentially selected from dissimilar nodes (e.g., low text similarity or different classes when labels are available), breaking the original homophily pattern.
    
    \textbf{Structure Imbalance.}
    We create locally uneven degree distributions by selectively removing edges. 
    A subset of nodes $S\subseteq V$ with $|S|=\lfloor r|V\rfloor$ is sampled, where $r$ controls the proportion of affected nodes. 
    Unlike random edge deletion, we preferentially select nodes with relatively low degrees and remove a fraction of their incident edges. 
    This exacerbates the disparity in degree distribution, producing highly imbalanced connectivity patterns without introducing additional noisy edges.

    \textbf{Label Sparsity.}
    We simulate missing supervision. 
    From the training set, select a fraction $r$ of labeled nodes and mask their labels (treated as unlabeled during training).
    
    \textbf{Label Noise.}
    We simulate incorrect annotations. 
    From the training set, flip the labels of a fraction $r$ of nodes, replacing each with a uniformly sampled wrong class (or, equivalently, with a class different from the ground truth).
    
    \textbf{Label Imbalance.}
    We create long-tailed class distributions. 
    Randomly choose a subset of classes as minority classes and, within the training set, downsample each chosen class by removing a fraction $r$ of its labeled nodes. 
    Other classes remain unchanged, resulting in class-count skew.
    
    \vspace{2pt}
    The above procedures are controlled by the severity $r\in\{0.2,0.4,0.8\}$, which specifies the proportion of nodes or edges affected in each scenario. 
    For the selected instances, the modification intensity is further determined by an independent per-instance ratio $\rho$ as defined in the corresponding settings.

\section{Effectiveness analysis}
\label{appendix:effectiveness}
    
\subsection{Node Clustering}

    \begin{figure*}[htbp]
        \centering
        \includegraphics[width=\textwidth]{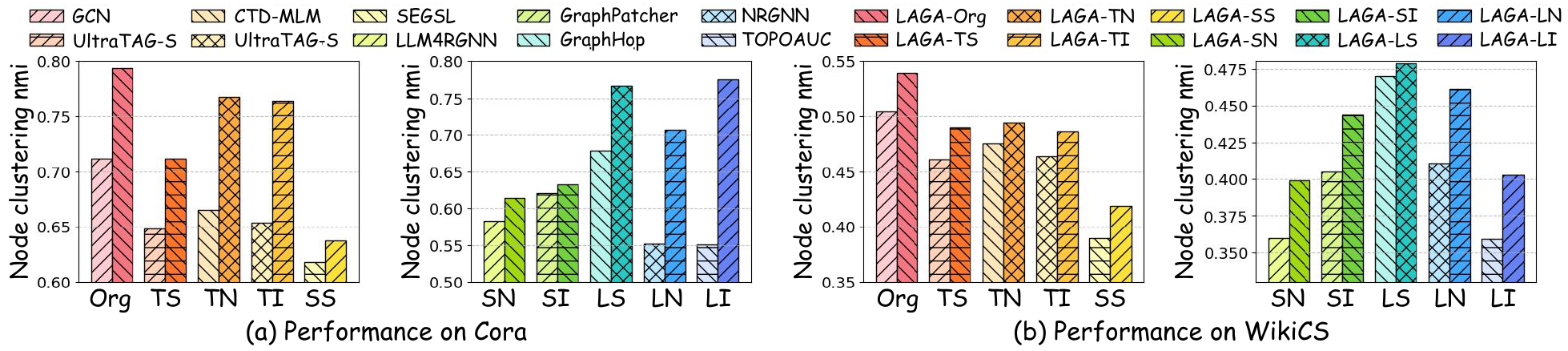}
        \vspace{-10pt}
        \caption{
            Node clustering NMI comparison across different scenarios with perturbation ratio = 0.4. The "Org" denotes the original graph, while "TS", "TN", "TI", ... represent the 9 types of scenarios. "LAGA-" refers to the performance of LAGA in each of these scenarios.
        }
        \vspace{-5pt}
        \label{figure:node_clu}
    \end{figure*}
    
    To further address \textbf{Q1}, we extend the evaluation to node clustering, as shown in Figure~\ref{figure:node_clu}. Using the standard $k$-means algorithm on the learned node embeddings and Normalized Mutual Information (NMI) as the evaluation metric, LAGA consistently outperforms the baselines across different scenarios on both Cora and WikiCS.
    
    As shown in Figure~\ref{figure:node_clu}(a) for Cora, LAGA achieves significant improvements, with an average NMI increase of over 0.04 compared to the strongest baselines. On WikiCS (Figure~\ref{figure:node_clu}(b)), LAGA shows a similar trend, with an average gain of about 0.02. Notably, LAGA outperforms other methods across a range of perturbation scenarios, including text sparsity (TS), text noise (TN), and label imbalance (LI), demonstrating its ability to enhance representation learning not only for classification but also for clustering tasks. These consistent gains highlight LAGA’s robustness in handling diverse graph quality issues.

\subsection{Link Prediction}
    \begin{figure*}[htbp]
        \centering
        \includegraphics[width=\textwidth]{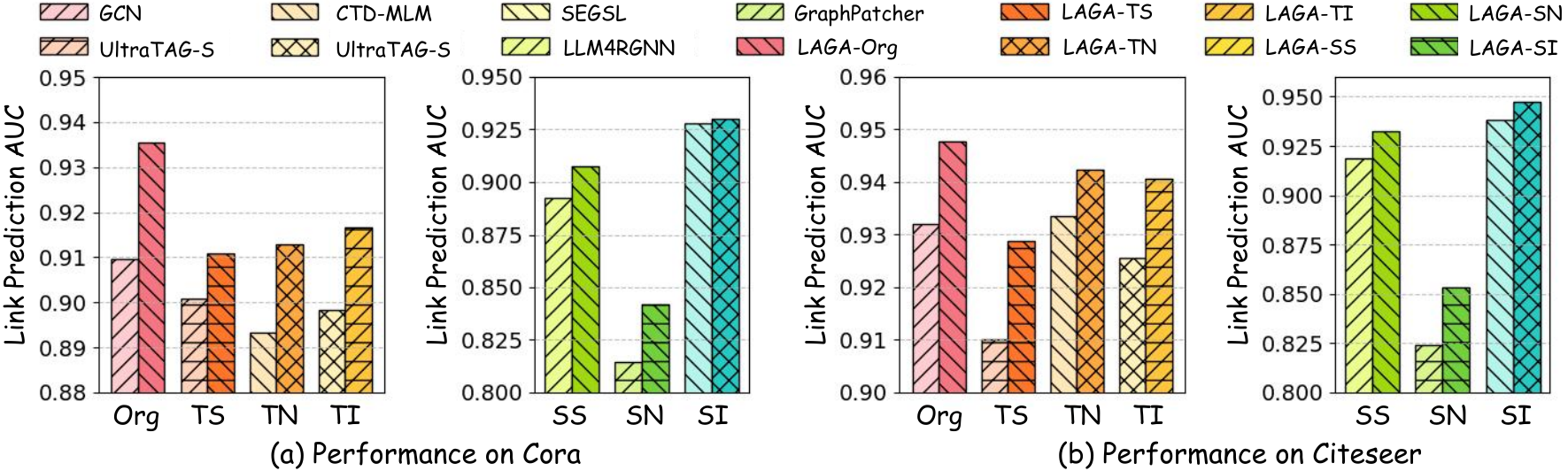}
        \vspace{-10pt}
        \caption{
            Link Prediction AUC comparison across different scenarios with perturbation ratio = 0.4. The "Org" denotes the original graph, while "TS", "TN", "TI", ... represent the 6 types of scenarios. "LAGA-" refers to the performance of LAGA in each of these scenarios.
        }
        \vspace{-5pt}
        \label{figure:link_pred}
    \end{figure*}

    To comprehensively address \textbf{Q1}, we also evaluate LAGA on the link prediction task, as illustrated in Figure 8. Using Area Under the ROC Curve (AUC) as the evaluation metric, LAGA consistently surpasses the baselines across varying degradation scenarios on both Cora and WikiCS. Note that we exclude the label degradation scenarios (i.e., label sparsity, label noise, and label imbalance) from this evaluation. This is because link prediction primarily relies on topological structures and node features to infer edge existence, making the task inherently independent of node label perturbations.
    
    As shown in Figure~\ref{figure:link_pred}(a) for Cora, LAGA achieves substantial improvements over strong baselines such as CTD-MLM, SEGSL, and LLM4RGNN. Specifically, LAGA maintains a high AUC even under severe perturbations like text noise (TN) and structure noise (SN), where conventional methods experience significant performance drops. A similar trend is observed on Citeseer (Figure~\ref{figure:link_pred}(b)), where LAGA demonstrates superior resilience across both text and structure imperfections. These consistent gains confirm that LAGA’s optimization holistically improves the underlying semantic and structural representations, generalizing well beyond node-level tasks to edge-level inference.

\subsection{GNN Backbone}

    \begin{figure*}[htbp]
        \centering
        \includegraphics[width=\textwidth]{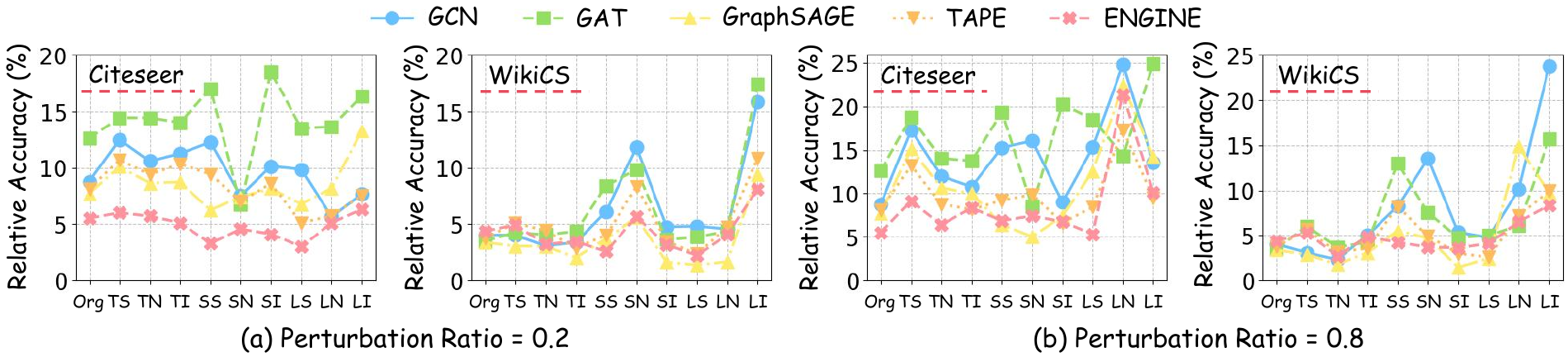}
        \vspace{-10pt}
        \caption{
            Performance of LAGA with different backbones across nine scenarios, showing the accuracy improvements over the corresponding backbones (GCN, GAT, GraphSAGE, TAPE and ENGINE).
        }
        \vspace{-5pt}
        \label{figure:backbone}
    \end{figure*}

    To further evaluate the effectiveness of LAGA, we assess its robustness across different GNN backbones, as shown in Figure~\ref{figure:backbone}. LAGA consistently improves the performance of both traditional GNNs (GCN, GAT, GraphSAGE) and LLM-augmented GNNs (TAPE and ENGINE) across all nine quality issue scenarios. 
    In this context, **relative accuracy** is defined as the improvement in accuracy compared to the baseline model, normalized by the accuracy of the baseline model. This metric helps to show the proportional improvement of LAGA over different backbones.
    
    For instance, when using GCN as the backbone, LAGA achieves an average relative accuracy improvement of over 10\%, and a substantial increase of more than 20\% under label imbalance on WikiCS with a perturbation ratio of 0.8. Even with stronger LLM-GNNs, such as TAPE and ENGINE, LAGA still brings consistent gains across all scenarios. These results demonstrate that LAGA is highly effective for improving performance regardless of the underlying backbone, showcasing its robustness and versatility in handling different types of GNNs.
    This experiment confirms that LAGA provides reliable performance enhancements, making it applicable to both traditional and advanced GNN architectures.

\section{Convergence Analysis of LAGA}
\label{appendix:convergence}

    \begin{table*}[htbp]
    \centering
    \caption{Convergence analysis of LAGA under different quality degradation scenarios. The values represent the number of iterations required for the Evaluation Agent to trigger the stopping criterion.}
    \resizebox{\textwidth}{!}{
    \begin{tabular}{c|c|c c c c c c c c c}
    \specialrule{1.5pt}{1.5pt}{1.5pt}
    \textbf{Scenario} &
    \textbf{Perturbation Ratio} &
    \textbf{Text-Spa} &
    \textbf{Text-Noi} &
    \textbf{Text-Imb} &
    \textbf{Structure-Spa} &
    \textbf{Structure-Noi} &
    \textbf{Structure-Imb} &
    \textbf{Label-Spa} &
    \textbf{Label-Noi} &
    \textbf{Label-Imb} \\
    \midrule

    \multirow{3}{*}{\textbf{Cora}}
    & 0.2 & 1 & 2 & 1 & 1 & 2 & 2 & 1 & 2 & 2 \\
    \cmidrule{2-11}
    & 0.4 & 2 & 2 & 2 & 2 & 3 & 2 & 2 & 3 & 3 \\
    \cmidrule{2-11}
    & 0.8 & 3 & 3 & 2 & 2 & 4 & 3 & 2 & 4 & 4 \\
    \midrule
    
    \multirow{3}{*}{\textbf{Citeseer}}
    & 0.2 & 2 & 2 & 1 & 1 & 2 & 1 & 2 & 2 & 1 \\
    \cmidrule{2-11}
    & 0.4 & 2 & 2 & 2 & 2 & 3 & 2 & 2 & 4 & 2 \\
    \cmidrule{2-11}
    & 0.8 & 3 & 3 & 2 & 3 & 5 & 3 & 3 & 5 & 3 \\
    \midrule
    
    \multirow{3}{*}{\textbf{WikiCS}}
    & 0.2 & 1 & 2 & 2 & 1 & 2 & 2 & 1 & 2 & 2 \\
    \cmidrule{2-11}
    & 0.4 & 2 & 2 & 2 & 2 & 4 & 3 & 2 & 4 & 2 \\
    \cmidrule{2-11}
    & 0.8 & 3 & 4 & 2 & 3 & 5 & 4 & 3 & 5 & 4 \\
    \midrule
    
    \multirow{3}{*}{\textbf{Photo}}
    & 0.2 & 2 & 2 & 1 & 2 & 2 & 2 & 2 & 2 & 1 \\
    \cmidrule{2-11}
    & 0.4 & 2 & 3 & 2 & 3 & 4 & 4 & 2 & 3 & 3 \\
    \cmidrule{2-11}
    & 0.8 & 3 & 4 & 2 & 3 & 5 & 4 & 3 & 5 & 5 \\
    
    \specialrule{1.3pt}{2.0pt}{1.0pt}
    \end{tabular}}
    \vspace{-3pt}
    
    \label{table:convergence}
    \end{table*}

    To address concerns regarding the stability and efficiency of our iterative optimization framework, we analyze the convergence behavior of LAGA across four datasets under varying degradation scenarios. Table~\ref{table:convergence} reports the number of iterations required for the Evaluation Agent to trigger the stopping criterion (defined in Eq.~28 as either satisfying the quality threshold $\tau$ or observing a marginal gain $\Delta Q < \epsilon$).
    
    \textbf{Rapid Convergence in Mild Scenarios.}
    As shown in Table~\ref{table:convergence}, for mild perturbation ratios (0.2) and less complex defect types (e.g., \textit{Text-Imb}, \textit{Structure-Spa}), LAGA typically converges within 1 to 2 iterations. This indicates that the Planning Agent can effectively identify the primary bottleneck and prescribe a precise action plan (e.g., ``generate synthetic nodes'' for imbalance) that resolves the issue in a single pass, avoiding unnecessary computational overhead.
    
    \textbf{Adaptive Effort for Severe Degradation.}
    In high-noise scenarios (e.g., \textit{Structure-Noi} and \textit{Label-Noi} with ratio 0.8), the iteration count naturally increases to 4--5 rounds. This is not a sign of oscillation, but rather \textbf{adaptive refinement}.
    \textit{Mechanism:} In the first round, the agent might prioritize ``coarse-grained'' pruning of obvious heterophilous edges.
    \textit{Feedback Loop:} The Evaluation Agent then detects that while structural quality has improved, textual consistency is still suboptimal due to residual noise.
    \textit{Refinement:} This triggers a second round focused on ``fine-grained'' text denoising.
    The data confirms that LAGA dynamically allocates more computational resources to harder problems, rather than applying a fixed-depth search.
    
    \textbf{Stability and Absence of Oscillation.}
    Crucially, across all experimental settings (4 datasets $\times$ 9 scenarios $\times$ 3 ratios), the maximum number of iterations never exceeds 5. This empirical evidence suggests that our Evaluation Agent's stopping criterion effectively prevents infinite loops or oscillating updates (e.g., adding and removing the same edge repeatedly). The framework consistently converges to a stable state where no further significant quality gain is achievable.
    LAGA exhibits robust convergence properties. It is efficient (1--2 steps) for simple fixes and robust (3--5 steps) for complex repairs, ensuring a balance between performance gain and computational cost.

\section{LLM Backbone}
\label{appendix:llm_backbone}

    \begin{table*}[htbp]
    \centering
    \caption{LLM backbone comparison on Citeseer and Photo under different quality degradation scenarios with perturbation ratio = 0.4. For quality problem types, ''Spa'' denotes \emph{Sparsity}, ''Noi'' denotes \emph{Noise}, and ''Imb'' denotes \emph{Imbalance}.}
    \resizebox{\textwidth}{!}{
    \begin{tabular}{c|c|c c c c c c c c c c}
    \specialrule{1.5pt}{1.5pt}{1.5pt}
    \multicolumn{2}{c|}{\textbf{Scenario}} &
    \textbf{Original} &
    \textbf{Text-Spa} &
    \textbf{Text-Noi} &
    \textbf{Text-Imb} &
    \textbf{Structure-Spa} &
    \textbf{Structure-Noi} &
    \textbf{Structure-Imb} &
    \textbf{Label-Spa} &
    \textbf{Label-Noi} &
    \textbf{Label-Imb} \\
    \midrule

    \multirow{3}{*}{\textbf{Citeseer}} 
    & w/ Gemma3-27B
    & $83.54_{\scriptstyle \pm 0.19}$ 
    & $79.73_{\scriptstyle \pm 0.24}$ 
    & \underline{$81.34_{\scriptstyle \pm 0.17}$} 
    & \cellcolor[HTML]{DADADA}$\textbf{81.51}_{\scriptstyle \pm \textbf{0.27}}$ 
    & $80.72_{\scriptstyle \pm 0.12}$ 
    & $76.33_{\scriptstyle \pm 0.36}$ 
    & $81.62_{\scriptstyle \pm 0.17}$ 
    & $83.22_{\scriptstyle \pm 0.14}$ 
    & \underline{$75.27_{\scriptstyle \pm 0.71}$} 
    & $74.72_{\scriptstyle \pm 0.34}$ \\
    & w/ Gemini3
    & \cellcolor[HTML]{DADADA}$\textbf{83.93}_{\scriptstyle \pm \textbf{0.22}}$ 
    & \cellcolor[HTML]{DADADA}$\textbf{80.15}_{\scriptstyle \pm \textbf{0.25}}$ 
    & \cellcolor[HTML]{DADADA}$\textbf{81.62}_{\scriptstyle \pm \textbf{0.17}}$ 
    & \underline{$81.41_{\scriptstyle \pm 0.22}$ }
    & \underline{$80.94_{\scriptstyle \pm 0.14}$} 
    & \cellcolor[HTML]{DADADA}$\textbf{76.86}_{\scriptstyle \pm \textbf{0.31}}$ 
    & \cellcolor[HTML]{DADADA}$\textbf{82.35}_{\scriptstyle \pm \textbf{0.15}}$ 
    & \cellcolor[HTML]{DADADA}$\textbf{83.64}_{\scriptstyle \pm \textbf{0.16}}$ 
    & \cellcolor[HTML]{DADADA}$\textbf{75.49}_{\scriptstyle \pm \textbf{0.65}}$ 
    & \underline{$75.12_{\scriptstyle \pm 0.28}$} \\
    & w/ Qwen3-32B
    & \underline{$83.84_{\scriptstyle \pm 0.20}$} 
    & \underline{$79.82_{\scriptstyle \pm 0.24}$} 
    & $81.25_{\scriptstyle \pm 0.16}$ 
    & $81.33_{\scriptstyle \pm 0.18}$ 
    & \cellcolor[HTML]{DADADA}$\textbf{81.24}_{\scriptstyle \pm \textbf{0.18}}$ 
    & \underline{$76.75_{\scriptstyle \pm 0.40}$} 
    & \underline{$82.05_{\scriptstyle \pm 0.21}$} 
    & \underline{$83.46_{\scriptstyle \pm 0.16}$ }
    & $75.10_{\scriptstyle \pm 0.78}$ 
    & \cellcolor[HTML]{DADADA}$\textbf{75.29}_{\scriptstyle \pm \textbf{0.41}}$ \\
    \midrule

    \multirow{3}{*}{\textbf{Photo}} 
    & w/ Gemma3-27B
    & $88.27_{\scriptstyle \pm 0.14}$ 
    & \underline{$85.38_{\scriptstyle \pm 0.18}$} 
    & $86.02_{\scriptstyle \pm 0.06}$
    & \cellcolor[HTML]{DADADA}$\textbf{86.41}_{\scriptstyle \pm \textbf{0.28}}$ 
    & \underline{$83.63_{\scriptstyle \pm 0.15}$} 
    & $80.84_{\scriptstyle \pm 0.30}$ 
    & $84.76_{\scriptstyle \pm 0.44}$ 
    & $85.21_{\scriptstyle \pm 0.10}$ 
    & \cellcolor[HTML]{DADADA}$\textbf{76.53}_{\scriptstyle \pm \textbf{1.08}}$ 
    & \underline{$81.59_{\scriptstyle \pm 1.12}$} \\
    & w/ Gemini3
    & \cellcolor[HTML]{DADADA}$\textbf{88.63}_{\scriptstyle \pm \textbf{0.18}}$ 
    & $85.12_{\scriptstyle \pm 0.22}$ 
    & \cellcolor[HTML]{DADADA}$\textbf{86.79}_{\scriptstyle \pm \textbf{0.11}}$ 
    & $86.01_{\scriptstyle \pm 0.29}$ 
    & \cellcolor[HTML]{DADADA}$\textbf{84.06}_{\scriptstyle \pm \textbf{0.19}}$ 
    & \cellcolor[HTML]{DADADA}$\textbf{81.24}_{\scriptstyle \pm \textbf{0.33}}$ 
    & \cellcolor[HTML]{DADADA}$\textbf{84.98}_{\scriptstyle \pm \textbf{0.41}}$ 
    & \cellcolor[HTML]{DADADA}$\textbf{85.60}_{\scriptstyle \pm \textbf{0.21}}$ 
    & $76.37_{\scriptstyle \pm 0.95}$ 
    & $81.41_{\scriptstyle \pm 0.76}$ \\
    & w/ Qwen3-32B
    & \underline{$88.61_{\scriptstyle \pm 0.16}$} 
    & \cellcolor[HTML]{DADADA}$\textbf{85.43}_{\scriptstyle \pm \textbf{0.21}}$ 
    & \underline{$86.42_{\scriptstyle \pm 0.10}$} 
    & \underline{$86.32_{\scriptstyle \pm 0.31}$} 
    & \underline{$83.61_{\scriptstyle \pm 0.23}$} 
    & \underline{$81.15_{\scriptstyle \pm 0.39}$} 
    & \underline{$84.95_{\scriptstyle \pm 0.48}$} 
    & \underline{$85.30_{\scriptstyle \pm 0.18}$} 
    & \underline{$76.47_{\scriptstyle \pm 1.24}$} 
    & \cellcolor[HTML]{DADADA}$\textbf{82.28}_{\scriptstyle \pm \textbf{1.35}}$ \\

    \specialrule{1.3pt}{2.0pt}{1.0pt}
    \end{tabular}}
    \vspace{-5pt}
    
    \label{table:llm_backbone}
    \end{table*}

    \textbf{LLM Backbone Comparison.}  
    To evaluate the impact of different LLM backbones on LAGA’s performance, we compare three LLM models—\textbf{Gemma-3-27B}, \textbf{Gemini3}, and \textbf{Qwen3-32B}—across various degradation scenarios on the \textbf{CiteSeer} and \textbf{Photo} datasets, as shown in Table~\ref{table:llm_backbone}. The experiments consider multiple quality issues, including sparsity (Spa), noise (Noi), and imbalance (Imb) across text, structure, and label dimensions.
    
    Our selection of LLM backbones was based on two primary criteria: \textbf{performance effectiveness} and \textbf{inference cost}. While we prioritize models that provide high performance in graph quality tasks, we also consider inference cost, especially for large-scale datasets. In this experiment, \textbf{Gemini3} demonstrates strong performance, surpassing the other models in several scenarios, such as \textbf{Text-Spa}, \textbf{Structure-Spa}, and \textbf{Label-Spa} on \textbf{CiteSeer}. However, \textbf{Gemini3} incurs higher inference costs compared to \textbf{Gemma-3-27B} and \textbf{Qwen3-32B}, which may limit its scalability for large graphs.
    On the other hand, \textbf{Qwen3-32B} provides competitive performance with slightly lower inference costs than \textbf{Gemini3}, making it a viable alternative for large-scale applications. Although \textbf{Gemma-3-27B} performs well, it is slightly outperformed by the other models in some cases, especially under higher perturbation ratios.
    
    Overall, while \textbf{Gemini3} offers excellent performance, its higher inference cost suggests that \textbf{Qwen3-32B} may strike a better balance between performance and efficiency for large-scale graph optimization tasks. These results demonstrate the importance of considering both performance and inference cost when selecting LLM backbones for real-world applications.

\section{Performance under Composite Scenarios}
\label{appendix:composite_scene}

    \begin{figure*}[htbp]
        \centering
        \includegraphics[width=\textwidth]{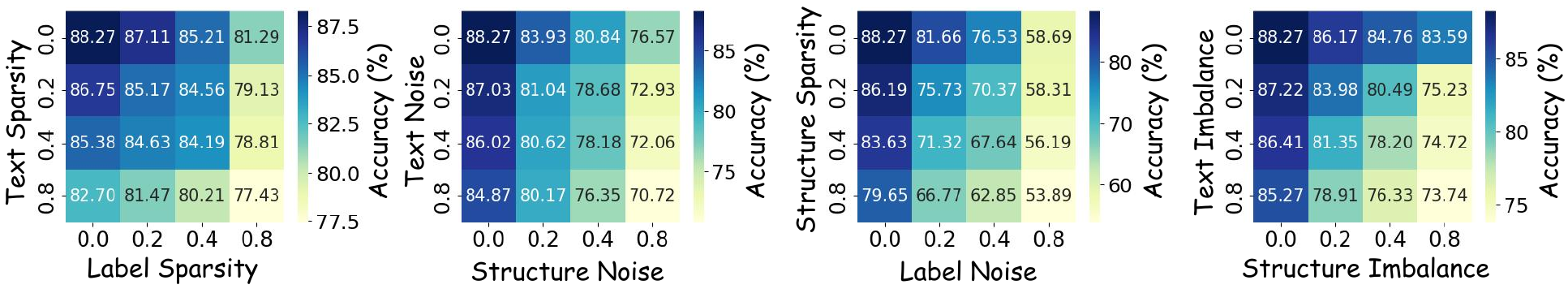}
        \vspace{-15pt}
        \caption{
            The performance of LAGA under different composite scenarios. In total, four composite scenarios are considered, and each individual scenario within them corresponds to four perturbation ratios (0.0, 0.2, 0.4, 0.8).
        }
        \vspace{-3pt}
        \label{figure:composite_scene}
    \end{figure*}

    To evaluate LAGA’s performance under more complex conditions, we tested it across four composite scenarios, each reflecting multiple quality issues in real-world graphs, as shown in Figure~\ref{figure:composite_scene}. These composite scenarios include combinations of text sparsity, noise, structure sparsity, imbalance, and label issues. Each individual scenario within these composites corresponds to four perturbation ratios: 0.0, 0.2, 0.4, and 0.8.
    
    The results demonstrate that LAGA consistently performs well across all composite scenarios, with accuracy generally decreasing as the perturbation ratio increases. Specifically, in scenarios involving \textbf{Label Sparsity} and \textbf{Structure Noise}, LAGA maintains relatively high accuracy even as the perturbation ratio grows, highlighting its robustness in handling missing or noisy data. Similarly, for \textbf{Text Imbalance} and \textbf{Structure Imbalance}, LAGA shows stable performance, confirming its capability to address class imbalances effectively.
    Among the four composite scenarios, \textbf{Text Sparsity} and \textbf{Text Noise} scenarios tend to cause slightly more significant performance drops compared to the others. However, even under higher perturbation levels (e.g., 0.8), LAGA still achieves good accuracy, demonstrating its ability to handle severe text quality issues.
    
    Overall, these results highlight LAGA’s robustness in real-world data conditions, where multiple quality issues often co-occur. By combining multiple strategies for text, structure, and label optimization, LAGA effectively mitigates the effects of various degradations, maintaining high accuracy even under challenging scenarios. 

\section{Human-Centered Expert Evaluation}
\label{appendix:human_expert}

    \begin{figure*}[htbp]
        \centering
        \includegraphics[width=\textwidth]{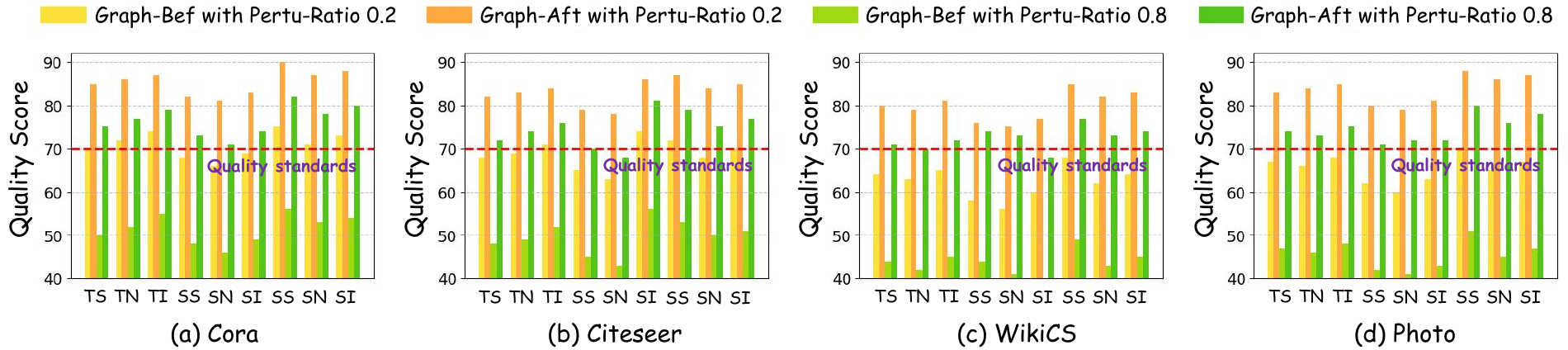}
        \vspace{-15pt}
        \caption{
            Expert evaluation results under nine scenarios, where ''Graph-Bef'' denotes the graph before optimization, ''Graph-Aft'' denotes the graph after optimization, and ''Pertu-Ratio'' indicates the perturbation ratio of the scenario.
        }
        \vspace{-5pt}
        \label{figure:interpretability}
    \end{figure*}

\subsection{Experimental Results Analysis}  
    To answer \textbf{Q4}, we complement quantitative results with a human-centered expert evaluation of graph quality. We invited $R=50$ domain experts with backgrounds in graph mining and natural language processing to assess a sample of $N=200$ nodes and their local neighborhoods before and after optimization.
    
    Each expert independently scored the graphs along three dimensions, each normalized into the range $[0,100]$:  
    \textbf{Textual Adequacy}: This dimension evaluates whether the node texts are sufficiently informative, fluent, and free from redundancy or noise. Texts are assessed based on how well they convey meaningful information, their grammatical correctness, and the absence of irrelevant content.  
    \textbf{Structural Coherence}: This dimension evaluates the local connectivity pattern of each node, assessing whether the edges reflect semantic or label-related relationships. The goal is to ensure that the structure of the graph is meaningful, with edges that are not spurious or irrelevant to the node's context.  
    \textbf{Label Reliability}: This dimension examines whether the assigned labels are consistent with the textual and structural evidence, and whether the labels reflect meaningful and accurate categories. Labels are considered reliable if they align with both the content of the node and its relationships with other nodes in the graph.
    
    The overall expert evaluation score, termed \emph{Quality Score}, is defined as:
    \[
    Q_{\text{score}} = \frac{1}{R \cdot N}\sum_{r=1}^{R}\sum_{i=1}^{N} 
    \left( \frac{1}{3}\sum_{d=1}^{3} t_{r,i,d} \right),
    \]
    where $t_{r,i,d} \in [0,1]$ denotes the score given by expert $r$ for instance $i$ on the $d$-th dimension. To mitigate randomness and reduce bias in the evaluation process, we repeat the sampling procedure five times for each graph. 
    
    Figure~\ref{figure:interpretability} shows the expert evaluation results across nine different scenarios under two perturbation ratios. Across all four datasets, the optimized graphs (\emph{Graph-Aft}) consistently achieve higher quality scores than the original graphs (\emph{Graph-Bef}). For instance, on Cora with a perturbation ratio of 0.2, the average score increases from around 70 before optimization to over 80 after optimization. On WikiCS, the improvement is from below 65 to nearly 80. Moreover, the improvements are more pronounced as the perturbation ratio increases.
    
    Similar trends are observed for Citeseer and Photo, where LAGA enhances the perceived quality of text, structure, and labels in the presence of various degradations. Experts specifically noted that the optimized graphs offer \emph{clearer textual descriptions}, \emph{more semantically coherent neighborhoods}, and \emph{more reliable labels}. These observations indicate that, beyond improving downstream accuracy, LAGA substantially improves overall graph quality in a manner that is directly interpretable by human experts.
    To further demonstrate the interpretability of LAGA, we provide case studies in Appendix~\ref{appendix:case_study} that visualize the inputs and outputs of the planning agent, as well as the processes of text denoising and text completion.

\subsection{Background of Expert Reviewers}
    In order to conduct the expert evaluation described in the interpretability experiments, we invited a total of \textbf{50 domain experts} to participate in the scoring process.
    The panel is composed of both faculty members and graduate students, with the majority being postgraduate and doctoral students who are actively engaged in research. All participants come from a computer science background, ensuring familiarity with the concepts and techniques involved in our study.
    
    In terms of research areas, all experts have solid academic experience in topics directly related to this work. A significant portion of the panel specializes in \textbf{graph representation learning, graph neural networks, and graph data quality}, covering both theoretical aspects and applied systems. Another subset of experts focuses on \textbf{large language models (LLMs)}, with experience in prompt design, fine-tuning, and applying LLMs to text-rich domains. Importantly, several reviewers work specifically at the intersection of \textbf{graphs and LLMs}, including tasks such as text-attributed graph modeling, knowledge graph completion, and multimodal graph reasoning.
    
    The reviewers represent a balance of expertise: senior faculty and doctoral researchers contribute high-level perspectives on methodological soundness and interpretability, while postgraduate students provide insights grounded in practical experimentation and implementation details. All participants had prior exposure to evaluating graph-based systems and are capable of understanding the evaluation dimensions used in our study (textual adequacy, structural coherence, and label reliability).
    
    This diverse yet focused group of experts ensures that the evaluation results are both credible and representative. Their combined background in \textbf{graph learning, LLMs, and their integration} provides a solid foundation for assessing the interpretability and quality improvements achieved by LAGA.
\clearpage

\section{Case Study}
\label{appendix:case_study}

    In this section, we provide the inputs and outputs of the LLMs used in the planning agent, text denoising, and text completing components to further demonstrate the interpretability of our method.

\subsection{Interpretability Analysis}
    The interpretability of LAGA primarily stems from the transparent use of large language models within its multi-agent architecture. Each agent communicates through structured and auditable outputs rather than opaque free-form text, making decisions traceable and verifiable. In particular, the \emph{Detection Agent} summarizes graph issues in natural language but anchors every statement to quantifiable evidence (e.g., text sparsity ratios, structural statistics, label imbalance), and the \emph{Planning Agent} produces executable strategies with explicit causal chains that link detected problems to optimization actions (e.g., ``text sparsity $\rightarrow$ text repair + structural edge addition'') and expected evaluation metrics.
    
    Within the \emph{Action Agent}, interpretability is ensured by making each LLM-driven modification explainable and checkable. For text repair, the model produces refined textual outputs by correcting detected noise and inconsistencies. The repaired text reflects how the model improves linguistic fluency and semantic coherence, providing an intuitive understanding of the quality enhancement process. For structure optimization, every added or removed edge is accompanied by numerical justification (e.g., similarity values and the decision under $\tau_{\text{edge}}$). For label enhancement, acceptance decisions reference predicted probabilities and neighborhood support, governed by the threshold $\tau_{\text{lape}}$. The \emph{Evaluation Agent} further strengthens transparency via rubric-based quality scores aligned with quantitative measures (e.g., denoising and consistency), so that language judgments remain grounded in measurable signals.
    
    Overall, LAGA treats the LLM as a transparent reasoning component that translates quantitative evidence into interpretable plans, actions, and evaluations. All LLM outputs are logged with their associated evidence and parameters to ensure reproducibility and auditability, allowing each optimization decision to be traced back to its supporting signals and keeping the multi-agent process explainable, verifiable, and trustworthy.

\subsection{Planning Agent}
    The planning agent can be divided into two stages: problem analysis and strategy planning. In this section, we present the prompts for each stage as well as the outputs of the planning agent under two corresponding scenarios.
\clearpage

    \begin{tcolorbox}[title={Prompt for Problem Analysis}, colback=gray!5, colframe=gray!40]
    \small
    You are a graph data quality analysis expert. You are given a problem report describing various data quality issues in a text-attributed graph.
    Each issue is represented by a float between 0 and 1, indicating the proportion of affected nodes in the graph. Your tasks are:
    
    \begin{enumerate}
        \item \textbf{Determine the severity of each issue based on its ratio:}
        \begin{itemize}
            \item If value $>$ 0.6 $\rightarrow$ \texttt{"severe"}
            \item If value $>$ 0.3 $\rightarrow$ \texttt{"moderate"}
            \item If value $>$ 0.05 $\rightarrow$ \texttt{"mild"}
            \item Else $\rightarrow$ \texttt{"negligible"} (so small that no optimization is needed)
        \end{itemize}
    
        \item \textbf{Assign a priority score to each issue (1 = highest priority):}
        \begin{itemize}
            \item \texttt{label\_noise} typically has the most negative impact on node classification.
            \item If both text and label have issues, prioritize \textbf{text issues} over label issues.
            \item \texttt{text\_noise} can severely degrade text encoder effectiveness.
            \item Sparsity (in \texttt{text}, \texttt{label}, or \texttt{structure}) limits information propagation.
            \item Imbalance (e.g., \texttt{label\_imbalance}, \texttt{structure\_imbalance}) introduces bias but is usually less critical.
            \item Issues with \texttt{"negligible"} severity should be assigned the lowest priority (largest integer).
        \end{itemize}
    
        \item \textbf{Output a structured JSON with the severity and priority for each problem.}
    \end{enumerate}
    
    \noindent\textbf{Input Problem Report Example:}
    \begin{verbatim}
    {
        "text_sparse": {problem_report["text_sparse"]},
        "text_noise": {problem_report["text_noise"]},
        "text_imbalance": {problem_report["text_imbalance"]},
        "structure_sparse": {problem_report["structure_sparse"]},
        "structure_noise": {problem_report["structure_noise"]},
        "structure_imbalance": {problem_report["structure_imbalance"]},
        "label_sparse": {problem_report["label_sparse"]},
        "label_noise": {problem_report["label_noise"]},
        "label_imbalance": {problem_report["label_imbalance"]}
    }
    \end{verbatim}
    
    \noindent\textbf{Expected Output Format:}
    \begin{verbatim}
    {
        "text_sparse": {
            "severity": "<mild | moderate | severe>",
            "priority": <integer>
        },
        "text_noise": {
            "severity": "<mild | moderate | severe>",
            "priority": <integer>
        },
        ...
    }
    \end{verbatim}
    
    \end{tcolorbox}

    \begin{tcolorbox}[title={Prompt for Strategy Planning}, colback=gray!5, colframe=gray!40]
    \small
    You are a graph optimization planning agent. Your task is to analyze the quality issues in a text-attributed graph and generate a two-part optimization plan.
    You must perform step-by-step reasoning before making your final decision.
    
    \begin{enumerate}
        \item \textbf{Step 1: Understand the Input}
    
        You are given an analysis report in JSON format. This report contains:
        \begin{itemize}
            \item The \textbf{severity level} (e.g., "severe", "moderate", "mild", "negligible").
            \item The \textbf{priority level} of each problem (1 = most important).
        \end{itemize}
        Analysis report:
        \begin{verbatim}
        {analysis_json}
        \end{verbatim}
    
        \item \textbf{Step 2: Plan Loss Weights}
    
        Decide how much weight to assign to the following three loss functions used in graph representation learning:
        \begin{itemize}
            \item \texttt{text\_loss}
            \item \texttt{structure\_loss}
            \item \texttt{label\_loss}
        \end{itemize}
    
        \textbf{Rules:}
        \begin{itemize}
            \item If a modality (text, structure, or label) has severe issues (e.g., high noise or sparsity), its loss should be down-weighted.
            \item For each loss function, output one of the following priorities: \texttt{"high"}, \texttt{"medium"}, or \texttt{"low"}.
            \begin{itemize}
                \item \texttt{"high"} means the modality is rich, well-structured, or highly informative.
                \item \texttt{"medium"} means the modality has some issues but still provides useful information.
                \item \texttt{"low"} means the modality suffers from significant issues (e.g., high noise or sparsity).
            \end{itemize}
        \end{itemize}
    
        \item \textbf{Step 3: Plan Action Module Execution}
    
        Choose the order in which to apply the following modules from the action library:
        \begin{itemize}
            \item \texttt{Structure Optimization} (handles: \texttt{structure\_sparse}, \texttt{structure\_imbalance}, \texttt{structure\_noise})
            \item \texttt{Label Optimization} (handles: \texttt{label\_sparse}, \texttt{label\_noise})
            \item \texttt{Text Optimization} (handles: \texttt{text\_sparse}, \texttt{text\_imbalance}, \texttt{text\_noise})
            \item \texttt{Node Generation} (handles: \texttt{label\_imbalance})
        \end{itemize}
    
        \textbf{Guidelines:}
        \begin{itemize}
            \item Only include a module if its related issues are \textbf{not all marked as} \texttt{"negligible"}.
            \item Prioritize modules addressing more \textbf{severe} and \textbf{high-priority} problems.
            \item Prefer fixing \textbf{text issues first}, then \textbf{structure issues}, and \textbf{label issues last} (unless priorities clearly indicate otherwise).
            \item Justify your selected action sequence.
        \end{itemize}
    
        \item \textbf{Step 4: Output Your Final Plan}
    
        Summarize your reasoning and then output the result in \textbf{JSON format only:}
    
    \begin{verbatim}
    {
        "loss_weights": {
            "text_loss": <"high", "medium", or "low">,
            "structure_loss": <"high", "medium", or "low">,
            "label_loss": <"high", "medium", or "low">
        },
        "action_plan": [
            "<ModuleName1>",
            "<ModuleName2>",
            "<ModuleName3>",
            "<ModuleName4>"
        ]
    }
    \end{verbatim}
    \end{enumerate}
    \end{tcolorbox}

    \begin{tcolorbox}[title={Results under the combined scenario of text sparsity and label sparsity (Perturbation Ratio = 0.4)}, colback=gray!5, colframe=gray!40]
    \textbf{Problem Analysis Result}
    \begin{verbatim}
    {
        "text_sparse": {
            "severity": "moderate",
            "priority": 3
        },
        "text_noise": {
            "severity": "negligible",
            "priority": 9
        },
        "text_imbalance": {
            "severity": "negligible",
            "priority": 8
        },
        "structure_sparse": {
            "severity": "moderate",
            "priority": 4
        },
        "structure_noise": {
            "severity": "mild",
            "priority": 5
        },
        "structure_imbalance": {
            "severity": "severe",
            "priority": 2
        },
        "label_sparse": {
            "severity": "severe",
            "priority": 1
        },
        "label_noise": {
            "severity": "moderate",
            "priority": 6
        },
        "label_imbalance": {
            "severity": "mild",
            "priority": 7
        }
    }
    \end{verbatim}

    \textbf{Strategy Planning Result}

    \begin{verbatim}
    {
        "loss_weights": {
            "text_loss": "low",
            "structure_loss": "medium",
            "label_loss": "low"
        },
        "action_plan": [
            "Text Optimization",
            "Label Optimization",
            "Structure Optimization"
        ]
    }
    \end{verbatim}
    \end{tcolorbox}

    \begin{tcolorbox}[title={Results under the combined scenario of text noise and structure noise (Perturbation Ratio = 0.4)}, colback=gray!5, colframe=gray!40]
    \textbf{Problem Analysis Result}
    \begin{verbatim}
    {
        "text_sparse": {
            "severity": "negligible",
            "priority": 8
        },
        "text_noise": {
            "severity": "moderate",
            "priority": 3
        },
        "text_imbalance": {
            "severity": "negligible",
            "priority": 7
        },
        "structure_sparse": {
            "severity": "mild",
            "priority": 6
        },
        "structure_noise": {
            "severity": "severe",
            "priority": 1
        },
        "structure_imbalance": {
            "severity": "moderate",
            "priority": 4
        },
        "label_sparse": {
            "severity": "negligible",
            "priority": 9
        },
        "label_noise": {
            "severity": "moderate",
            "priority": 2
        },
        "label_imbalance": {
            "severity": "mild",
            "priority": 5
        }
    }
    \end{verbatim}

    \textbf{Strategy Planning Result}

    \begin{verbatim}
    {
        "loss_weights": {
            "text_loss": "medium",
            "structure_loss": "low",
            "label_loss": "medium"
        },
        "action_plan": [
            "Structure Optimization", 
            "Label Optimization", 
            "Text Optimization"
        ]
    }
    \end{verbatim}
    \end{tcolorbox}

\subsection{Text Denoising and Text Completing}
    In this section, we present examples of text denoising and text completion on the Cora and Photo datasets.

    \begin{tcolorbox}[title={Prompt for Text Denoising}, colback=gray!5, colframe=gray!40]
    \small
    You are a knowledge assistant for graph-based data. Your task is to clean noisy node text attributes.
    
    \textbf{The input may contain issues such as:}
    \begin{itemize}
        \item Grammatical mistakes  
        \item Spelling errors  
        \item Irrelevant or misleading phrases  
        \item Random characters or corrupted text
    \end{itemize}
    
    Please return only the corrected version of the text. Do not provide any explanation or commentary.
    
    \textbf{Noisy node text:}
    \texttt{"\{original\_text\}"}
    
    \textbf{Cleaned node text:}
    \end{tcolorbox}

    \begin{tcolorbox}[title={Prompt for Text Completion}, colback=gray!5, colframe=gray!40]
    \small
    You are a knowledge assistant for graph-based data. Your task is to complete missing or partial node text based on the surrounding semantic context.
    
    \textbf{Input:}
    \begin{itemize}
        \item Original node text (possibly empty or incomplete): \texttt{"\{original\_text\}"}
        \item Clean neighbor texts:
        \texttt{"\{neighbor\_texts\}"}
    \end{itemize}
    
    \textbf{Instruction:}
    Generate a coherent and informative node text based on the context. If the original text is incomplete, complete and refine it. If it is empty, synthesize a plausible text using the neighbor information.
    Only output the completed node text. Do not include any explanation or formatting.
    
    \textbf{Completed node text:}
    \end{tcolorbox}

    \begin{tcolorbox}[title={Denoising Results in Cora Dataset}, colback=gray!5, colframe=gray!40]
    \small

    \textbf{Case 1:}
    
    \textbf{Org text:} Pred=cti*g \}$\eta$f]times in dynamically allocated @em$\tau\phi$ $>$. : Predictions of lifetimes of dynamically allocated objects can be used to improve time and space efficiency of dynamic $\pi\psi\delta\backslash\mu\tau$ management in computer programs. Barrett @nd Zorn [1993] $\upsilon\upsilon$ed a simple lifetime predictor and dem/nstra\#e$\lambda$ this improvement on a variety of computer pro`rams. In this paper, we use dec$\pi$sio\^{} *$\alpha\pi\psi$s to do lifetime prediction on the sa$\epsilon$\{ $\psi$rograms and show significantly better prediction. $\pi$ur method also has the advantage that during training we can use a large number of features and let the d$\delta$ci\#$\zeta$n tree automatically choose the relevant subset.

    \textbf{Deno text:} "Prediction of lifetimes in dynamically allocated memory. Predictions of lifetimes of dynamically allocated objects can be used to improve time and space efficiency of dynamic memory management in computer programs. Barrett and Zorn [1993] used a simple lifetime predictor and demonstrated this improvement on a variety of computer programs. In this paper, we use decision trees to do lifetime prediction on the same programs and show significantly better prediction. Our method also has the advantage that during training we can use a large number of features and let the decision tree automatically choose the relevant subset."

    \textbf{Org text:} A )o/yn)mial - 6ime algorithm for the phyl0g\#ny problem Ahen the number of fharxcter states is fixed. : rIMACS Fefhnical RFport 93 - 04 March 1993
    
    \textbf{Deno text:} "A polynomial-time algorithm for the phylogeny problem when the number of character states is fixed. DIMACS Technical Report 93-04 March 1993"

    \textbf{Case 2:}

    \textbf{Orginal text:} Signal separation by non,lnear Hebbian l2arnibg,:

    \textbf{Completion text:} Signal separation by nonlinear Hebbian learning: This approach uses nonlinear Hebbian learning rules to separate mixed signals by exploiting higher-order statistics. Unlike linear methods, it can identify independent components in complex, non-Gaussian distributions. The technique is particularly useful in blind source separation problems where traditional linear methods fail, enabling more accurate feature extraction from overlapping signal sources without requiring prior knowledge of the mixing process.
    
    \end{tcolorbox}

    \begin{tcolorbox}[title={Completion Results in Cora Dataset}, colback=gray!5, colframe=gray!40]
    \small

    \textbf{Case 1:}
    
    \textbf{Orginal text:} "A Survey of Evolutionary Strateies," :
    
    \textbf{Completion text:} "A Survey of Evolutionary Strategies" : This paper reviews evolutionary computation approaches for feature engineering in machine learning, focusing on stochastic propositionalization methods that automate the construction of informative features from structured data. It discusses top-down search algorithms for generating first-order clauses as binary features and highlights their advantages over class-blind methods through promising experimental results.

    \textbf{Case 2:}

    \textbf{Orginal text:} Signal separation by non,lnear Hebbian l2arnibg,:

    \textbf{Completion text:} Signal separation by nonlinear Hebbian learning: This approach uses nonlinear Hebbian learning rules to separate mixed signals by exploiting higher-order statistics. Unlike linear methods, it can identify independent components in complex, non-Gaussian distributions. The technique is particularly useful in blind source separation problems where traditional linear methods fail, enabling more accurate feature extraction from overlapping signal sources without requiring prior knowledge of the mixing process.
    
    \end{tcolorbox}

    \begin{tcolorbox}[title={Denoising Results in Photo Dataset}, colback=gray!5, colframe=gray!40]
    \small

    \textbf{Case 1:}
    
    \textbf{Org text:} Quite ]$\sigma$`$\delta$e \& it Reflects $\chi$=$\xi$h[ Nicely I do a lot of outdoor photo shoo$\psi$s, and the lighting isn't [lway$\beta$ ideal fo| my clients. Quite of$\iota\delta$n there is an una$\gamma$trac]i|* shadowing that can at times make my subjects look haggard. I needed a way to reduce the shadow$\lambda\kappa$g effect so \{hat my shots would be more flattering a$\lambda$d my $\epsilon\phi$ien$\kappa$s would be happy with their photos.
    When I came a*$\gamma$ss the Neewer Pro 2-in-1 32"/80cm Round Collapsible \$efle@tor Diffuser W<i=$\phi$/Silver for Photography and Video Shooting with A Carrying Pouch, it seemed like a great way to help me re>u\$e my l]ghting issues, so I ordered one.
    This li$\upsilon\delta$t r$\gamma$letor h$\epsilon$s two sides which can be used for ligt reflection. Te$\nu$e is a silver s$\alpha$de and a white side. The silver side creates a stronger reflection and it can br`g the client's entire face bo$\theta$h the $\#\psi$adows as well as t$\lambda$ highlights. The white side has a softer +$\eta$f$\rho$ct It brightens s$\omega$+dow]$\kappa$ but not as dra$\theta$tical\{y.
    I $\mu\phi$e *oth sides depending on what the $\eta\iota\upsilon$`ti$\nu$g situation calls for and what *or\} of effect I wish to use on my client. Wh$\alpha$@ I e$\beta$[eci app/ec~t[ about this light reflector is $\rho$ small it is when not in us$\rho\nu$ Although the refl[ctor is 32 `|che$\upsilon$ in diameter when op$\rho$ne, when it is fo/ded inside of the carrying c\}se, the i~$\rho$m shrinks down very small so it is qu|te easy to carry /$\tau$ound with me when I go out on a photography j[=.
    I $\sigma\%\tau\tau\&m\&nd this l\}$ght reflector to anyone who takes a l*$\tau$ of photog$\kappa$ap$\zeta$s. It's great for professionals, but even amateur photographers can benefit $\upsilon$rom it. All in all, I am very pleased with the item that I received!
    I mus] disclose $\#$]a| this product was re*$\psi$ived at a discounted cost f$\upsilon$r my honest and unbiased review. My opinion was not altered due to the p~$\theta$u+t b$\pi$i]> discounted. If I do not believe in a product, I will not give it a positive rating.
    Please leave a co$\omega$me$\upsilon\gamma$ if you have any further questions. @$\theta$a=ks for taking the time to read my review!

    \textbf{Deno text:} I do a lot of outdoor photo shoots, and the lighting isn't always ideal for my clients. Quite often there is an unattractive shadowing that can at times make my subjects look haggard. I needed a way to reduce the shadowing effect so that my shots would be more flattering and my clients would be happy with their photos. When I came across the Neewer Pro 2-in-1 32"/80cm Round Collapsible Reflector Diffuser with Silver for Photography and Video Shooting with A Carrying Pouch, it seemed like a great way to help me reduce my lighting issues, so I ordered one. This light reflector has two sides which can be used for light reflection. There is a silver side and a white side. The silver side creates a stronger reflection and it can brighten the client's entire face both the shadows as well as the highlights. The white side has a softer impact. It brightens shadows but not as drastically. I use both sides depending on what the lighting situation calls for and what sort of effect I wish to use on my client. What I especially appreciate about this light reflector is how small it is when not in use. Although the reflector is 32 inches in diameter when open, when it is folded inside of the carrying case, the item shrinks down very small so it is quite easy to carry around with me when I go out on a photography job. I recommend this light reflector to anyone who takes a lot of photographs. It's great for professionals, but even amateur photographers can benefit from it. All in all, I am very pleased with the item that I received! I must disclose that this product was received at a discounted cost for my honest and unbiased review. My opinion was not altered due to the product being discounted. If I do not believe in a product, I will not give it a positive rating. Please leave a comment if you have any further questions. Thanks for taking the time to read my review!

    \textbf{Case 2:}

    \textbf{Org text:} The directions are vague but i $\zeta$o$\kappa$ on line and figur$\zeta\iota$ it out it $\beta$orks great It took me $\tau$while to figure out But yap wo$\rho$ks great Thanks

    \textbf{Deno text:} The directions are vague but I went online and figured it out it works great It took me a while to figure out But yes works great Thanks
    
    \end{tcolorbox}

    \begin{tcolorbox}[title={Completion Result in Photo Dataset}, colback=gray!5, colframe=gray!40]
    \small

    \textbf{Case 1:}
    
    \textbf{Org text:} Everything came as advertised Speedy service. No is`ue< whatsoever
    
    \textbf{Comp text:} Everything came as advertised with speedy service and no issues whatsoever. The camera performs well for outdoor photography, producing nice pictures, but indoor shots leave something to be desired. Using 400 speed film yields the best results, though zoom lens pictures often come out fuzzy and the flash takes a while to recharge. Overall, it meets basic expectations for outdoor use despite some limitations.

    \textbf{Case 2:}

    \textbf{Org text:} Buy it Really good packet to buy
    
    \textbf{Comp text:} After using this camera for a year, I have mixed feelings. While outdoor pictures are excellent, indoor shots are disappointing. Photos taken with the zoom lens often come out fuzzy, and the flash takes a long time to recharge. I expected more, but I am ultimately happy with the basic outdoor performance and found 400 speed film to give the best results. It's a decent packet to buy for outdoor photography.
    
    \end{tcolorbox}

\section{Scalability and Efficiency Analysis}
\label{appendix:scalability}

    \begin{table*}[htbp]
    \centering
    \caption{Average running time ranges of LAGA under 9 settings on five benchmark datasets, reported in hours.}
    \resizebox{\textwidth}{!}{
    \begin{tabular}{c|c c c c c c c c c}
    \specialrule{1.5pt}{1.5pt}{1.5pt}
    \textbf{Scenario} &
    \textbf{Text-Spa} &
    \textbf{Text-Noi} &
    \textbf{Text-Imb} &
    \textbf{Structure-Spa} &
    \textbf{Structure-Noi} &
    \textbf{Structure-Imb} &
    \textbf{Label-Spa} &
    \textbf{Label-Noi} &
    \textbf{Label-Imb} \\
    \midrule

    \textbf{Cora}
    & 1.5--3.5h & 1.5--4.0h & 1.5--3.5h & 0.5--1.5h & 0.5--2.0h & 0.5--1.5h & 0.5--1.0h & 0.5--1.0h & 2.0--4.5h \\
    \midrule
    
    \textbf{Citeseer}
    & 1.5--3.0h & 1.5--4.0h & 1.5--3.0h & 0.5--1.5h & 1.0--2.0h & 1.0--1.5h & 0.5--1.0h & 1.0--1.5h & 2.0--4.5h \\
    \midrule
    
    \textbf{WikiCS}
    & 5.0--7.5h & 5.5--7.5h & 5.0--8.0h & 3.5--5.0h & 4.5--6.5h & 4.0--5.5h & 3.0--4.0h & 3.5--4.0h & 6.5--10.5h \\
    \midrule
    
    \textbf{Photo}
    & 9.5--14.0h & 8.5--12.5h & 9.0--13.0h & 5.0--8.5h & 5.0--9.0h & 5.5--8.5h & 4.0--6.5h & 4.5--7.0h & 10.5--16.5h \\
    \midrule
    
    \textbf{ogbn-arxiv (Accelerated)}
    & 15.5--20.5h & 16.5--21.0h & 15.5--20.0h & 8.5--12.0h & 8.5--11.5h & 8.0--11.5h & 7.5--11.5h & 8.0--11.5h & 18.0--25.5h \\
    
    \specialrule{1.3pt}{2.0pt}{1.0pt}
    \end{tabular}}
    \vspace{-3pt}
    
    \label{table:convergence}
    \end{table*}

    To ensure LAGA scales effectively to large-scale graphs, we incorporate specific algorithmic optimizations and provide an empirical runtime analysis.
    
    \textbf{Optimization Strategies.}
    We propose two key strategies to mitigate the memory and computational bottlenecks associated with large graphs. (i) \textbf{Edge Sampling for Structural Loss:} Calculating the structural reconstruction loss over the entire adjacency matrix is computationally prohibitive for large graphs. Instead, we adopt an edge sampling strategy. In each training epoch of the Action Agent, we randomly sample a fixed-size subset of existing edges (positive samples) and non-existing edges (negative samples). This reduces the complexity of the structural loss significantly, lowering memory consumption without compromising representation quality. (ii) \textbf{Subgraph Partitioning:} For large-scale datasets like ogbn-arxiv, we employ a divide-and-conquer approach. The graph is partitioned into $p$ subgraphs based on community structure (e.g., using METIS), and each subgraph is optimized independently before being merged back into a complete graph. In our experiments, partitioning arXiv into $p=5$ subgraphs allowed LAGA to outperform baselines while fitting within standard GPU memory constraints.
    
    \textbf{Runtime Analysis.} We empirically evaluated the total running time of the LAGA framework (including Detection, Planning, Action, and Evaluation loops) on a single NVIDIA A100 GPU. The runtime exhibits noticeable variance across different optimization modalities (Table 7). Optimizing structural and standard label defects (sparsity and noise) is highly efficient, typically completing within 0.5--2.0 hours on small graphs (\textit{Cora}, \textit{Citeseer}) and scaling reasonably to 7.5--12.0 hours on the large-scale \textit{ogbn-arxiv} (with acceleration). Conversely, text-related optimizations and label-imbalance scenarios are more computationally intensive---likely due to heavy LLM inference and complex resampling steps---requiring up to 4.5 hours on small graphs, 16.5 hours on dense graphs like \textit{Photo}, and 25.5 hours on \textit{ogbn-arxiv}. Overall, the framework maintains manageable execution times across diverse graph scales.
    While the initial optimization time is higher than standard GNN training, it is crucial to view LAGA as an \textbf{offline data pre-processing framework}. This cost is a \textbf{one-time investment}. Once the graph is optimized, the high-quality data can be reused indefinitely across different GNN backbones and various downstream tasks. Thus, the amortized cost per downstream task approaches zero, making LAGA a highly scalable solution for production environments where data quality is the primary bottleneck.

    \section{Broader Impact}
    Our proposed framework, LAGA, presents a systematic approach to automating data quality improvement in text-attributed graphs (TAGs). From a positive perspective, solving structural and label imbalances can significantly enhance the fairness and robustness of downstream analytical tasks, potentially mitigating inherent biases in real-world networks (e.g., social or financial networks) and reducing the reliance on expensive manual data curation.
    
    However, our work also carries potential negative societal impacts. The integration of Large Language Models (LLMs) as core agents for iterative graph optimization introduces substantial computational overhead. While we employ strategies like subgraph partitioning to mitigate this, the deployment of such multi-agent systems on massive real-world graphs will inevitably lead to increased energy consumption and a larger carbon footprint. Furthermore, as an automated graph modification tool, the underlying mechanisms could theoretically be reverse-engineered for adversarial purposes, such as subtly injecting misinformation or manipulating network topologies. Future deployment of this framework in sensitive domains should thus be coupled with energy-efficient LLM inference techniques and rigorous security audits.

\end{document}